\documentclass{article}

\PassOptionsToPackage{numbers, square}{natbib}  

\usepackage[final]{neurips_2025}

\usepackage[utf8]{inputenc} 
\usepackage[T1]{fontenc}    

\usepackage[
  backref=page,         
  colorlinks,
  linkcolor=blue,
  citecolor=blue,
  urlcolor=blue
]{hyperref}
\usepackage{url}
\usepackage{booktabs}       
\usepackage{amsfonts}       
\usepackage{nicefrac}       
\usepackage{microtype}      
\usepackage{xcolor}         

\usepackage{graphicx}       
\usepackage{multirow}
\usepackage{makecell}
\usepackage{amsmath}
\usepackage{amssymb}
\usepackage{mathtools}
\usepackage{amsthm}
\usepackage{algorithmic}
\usepackage{algorithm}
\usepackage[capitalise]{cleveref}
\usepackage{wrapfig}
\usepackage{subcaption}  
\usepackage{tcolorbox}    
\usepackage{footmisc}     
\newtheorem{theorem}{Theorem}[section]
\newtheorem{proposition}[theorem]{Proposition}

\newtheorem{corollary}[theorem]{Corollary}

\title{Angular Constraint Embedding via SpherePair Loss for Constrained Clustering}

\author{
 \hspace*{-6mm} Shaojie Zhang ~~~~~~~~ Ke Chen\\
  Department of Computer Science,
  The University of Manchester, Manchester M13 9PL, U.K. \\
  \texttt{\{shaojie.zhang,ke.chen\}@manchester.ac.uk} \\
}

\begin{document}

\maketitle

\begin{abstract}
Constrained clustering integrates domain knowledge through pairwise constraints.
However, existing deep constrained clustering (DCC) methods are either limited by anchors inherent in end-to-end modeling or struggle with learning discriminative Euclidean embedding, restricting their scalability and real-world applicability.
To avoid their respective pitfalls, we propose a novel angular constraint embedding approach for DCC, termed SpherePair.
Using the SpherePair loss with a geometric formulation,
our method faithfully encodes pairwise constraints and leads to embeddings that are clustering-friendly in angular space, effectively separating representation learning from clustering. 
SpherePair preserves pairwise relations without conflict, 
removes the need to specify the exact number of clusters, 
generalizes to unseen data, 
enables rapid inference of the number of clusters, 
and is supported by rigorous theoretical guarantees.
Comparative evaluations with state-of-the-art DCC methods on diverse benchmarks, along with empirical validation of theoretical insights, confirm its superior performance, scalability, and overall real-world effectiveness.
Code is available at \href{https://github.com/spherepaircc/SpherePairCC/tree/main}{our repository}.
\end{abstract}


\section{Introduction}
\label{introduction}

Clustering is pivotal in machine learning and data mining. Unsupervised clustering methods, being fundamentally ill-posed, often partition data based solely on instance similarities or connections, which may misalign with domain knowledge \cite{CCsurvey2007}. To address this issue, integrating domain knowledge through weakly supervised methods like Constrained Clustering \cite{CC1st,copkmeans,CCsurvey2023} has gained attention. These methods enforce both positive and negative instance-level pairwise constraints, significantly boosting clustering accuracy. Moreover, they offer a cost-effective solution in scenarios where obtaining pairwise relations is easier than acquiring class labels \cite{SDCD}.

Most early pairwise constrained clustering (CC) methods adapt traditional unsupervised clustering by introducing constraints through modified similarity metrics or penalty functions \cite{copkmeans,GMMCC,MPCKMeans,PCKMeans,KernelCC}. 
Recent advances in deep clustering have led to the emergence of deep constrained clustering (DCC) paradigms, which outperform traditional methods, particularly on high-dimensional and complex data across various data types, and generalize well to unseen instances. 
Broadly, based on whether constraints are enforced at the level of cluster assignments or instance embeddings, we propose categorizing DCC methods into two paradigms: \textit{end-to-end DCC} and \textit{deep constraint embedding}.

As the dominant paradigm, end-to-end DCC methods (e.g., \cite{PCC,SDEC,CIDEC1,DCGMM,VolMaxDCC}) reformulate clustering as a pseudo-classification task by introducing anchors to represent classes, learning representations and cluster assignments jointly.
However, the absence of global supervision hinders the proper alignment of anchors with cluster centers, resulting in a mismatch between local instance-level similarities and global cluster-level decisions. 
Moreover, these methods require prior knowledge of the number of clusters in the data to formulate the pseudo-classification task. 
These weaknesses, along with other technical issues reviewed in the next section, limit their practical usability in real applications.
In contrast, deep constraint embedding methods \cite{CPAC,AutoEmbedder} transform CC into traditional clustering through learned representations that encode constraint information using deep learning models. 
Nevertheless, these methods still struggle to maintain appropriate distances between positive and negative pairs in Euclidean space during representation learning.

\begin{wrapfigure}{r}{0.39\textwidth}
    \centering
    \vskip -0.2in
    \includegraphics[width=0.39\textwidth]{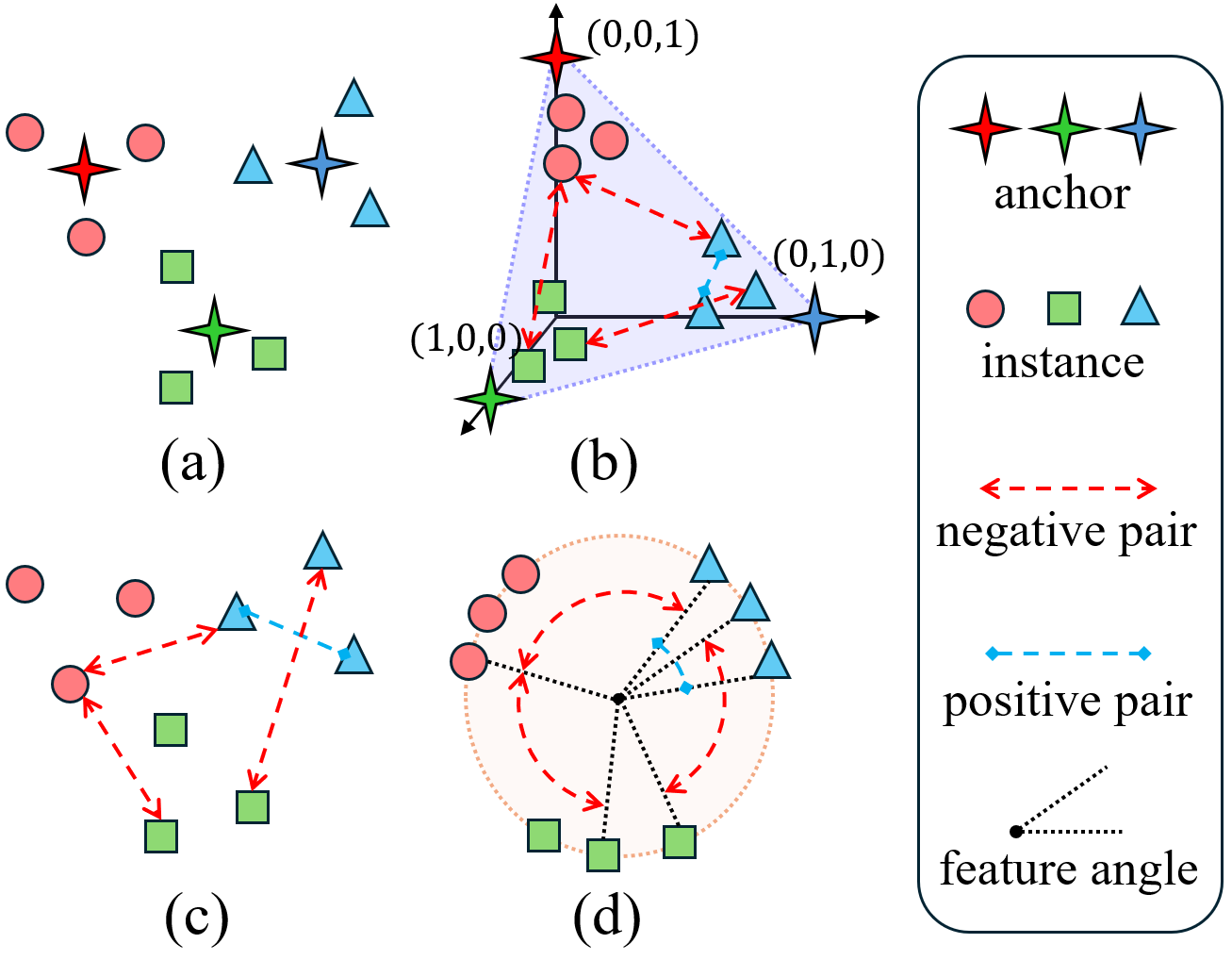}
    \vskip -0.0in
    \caption{Different pairwise learning approaches.
    End-to-end DCC introduces anchors to transform features in (a) into soft cluster assignments in (b) for pairwise losses.
    Deep constraint embedding in (c) focuses on the Euclidean distances between features, while ours in (d) operates in angular space.}
    \label{fig:z_p_space}
    \vskip -0.2in
\end{wrapfigure}

In this paper, we propose the novel \textit{SpherePair} loss function within a deep constraint embedding approach, addressing the limitations of existing DCC methods. 
As illustrated in  Fig.~\ref{fig:z_p_space}, unlike existing DCC methods that either require anchors or rely on pairwise loss based on Euclidean distance, our SpherePair loss employs cosine similarity to learn a latent representation in angular space without relying on  any anchors. 
This effectively balances pairwise relationships, resulting in a representation that accurately encodes constraint information towards minimizing intra-cluster distances and maximizing inter-cluster distances. 
Furthermore, our approach is supported by a theoretical foundation that ensures optimal performance under certain conditions, and does not require knowing the exact number of clusters, thereby reducing the effort required for hyperparameter tuning.
These strengths make our approach more scalable and practically applicable. 
As demonstrated, our approach outperforms state-of-the-art DCC baselines on various benchmark datasets, even when a simple K-means algorithm is applied to its learned representations.

Our main contributions are summarized as follows:
(i) We propose an angular constraint embedding method using the \textit{SpherePair} loss, offering a scalable and practical solution for real-world CC tasks.
(ii) We establish a rigorous theoretical foundation for our approach.
(iii) We demonstrate that our approach can handle unknown number of clusters, rapidly infer the number of clusters and generalize effectively to unseen instances.
(iv) We conduct extensive evaluations, demonstrating that our approach outperforms state-of-the-art DCC methods across diverse benchmark datasets.

\section{Related work}
\label{RelatedWorks}

\paragraph{End-to-end DCC.} 
End-to-end DCC methods utilize a predefined number of anchors to connect representations (see Fig.~\ref{fig:z_p_space}(a)) and clustering assignments (see Fig.~\ref{fig:z_p_space}(b)), serving as key elements in structuring clustering 
(See \cref{sec:formulation} for a preliminary with the formal formulation of end-to-end DCC).
Key differences arise from anchor configurations and pairwise losses: 
(i) \textit{Anchors} include class weights from neural network classification layers (e.g., \cite{VolMaxDCC}), centroids of temporary clusters in embedding-based models (e.g., \cite{SDEC,CIDEC2,CIDEC1}), and components of Gaussian mixture distributions in generative models (e.g., \cite{DCGMM}). 
(ii) \textit{Pairwise losses} constrain clustering assignments, 
using measures such as Kullback–Leibler (KL) divergence \cite{KCL,ClusterNet,CDEC} or inner products \cite{PCC,MCL,DCGMM}. The Meta Classification Likelihood (MCL) loss \cite{MCL} is theoretically validated and has inspired numerous extensions \cite{S3C2,CIDEC1,scDCC,ConstraintMatch,VolMaxDCC}. 
Despite their utility, anchors have notable limitations:
anchors often struggle with data misaligned to explicit clustering centers \cite{MV-Effi} and miss local relationships due to their emphasis on a global perspective \cite{MV-Tens, MV-Rob}.
In end-to-end DCC, this can further leads to potential mismatches, as global assignments can only be inferred indirectly from local pairwise relationships, especially when inappropriate anchors propagate errors during iterations.
These issues are exacerbated by imbalanced constraints (i.e., many constraints arising from a small number of clusters), which fail to capture nuanced data structures and ultimately degrade clustering performance. 
Finally, these methods require the exact number of clusters, limiting practical usability and scalability.

\paragraph{Deep constraint embedding.} 
Deep constraint embedding methods \cite{CPAC,AutoEmbedder} address CC problems by learning latent representations with deep learning models 
(see a preliminary in \cref{sec:formulation} for the formal formulation),
typically using anchor-free, pairwise Euclidean distance-based losses (see Fig.~\ref{fig:z_p_space}(c)).
However, AutoEmbedder \cite{AutoEmbedder} requires manually setting a margin due to the unbounded range of Euclidean distances $[0, +\infty)$. 
While CPAC \cite{CPAC} avoids margin tuning, its excessive expansion of negative pairs leads to non-convex, poorly distinguishable clusters, and its reliance on connectivity graphs limits generalization to unseen instances.
In contrast, our approach leverages angular distances between feature vectors (see Fig.~\ref{fig:z_p_space}(d)) to overcome the challenges of anchor-based and Euclidean distance-based methods. 
The angular space ensures equal, definitive inter-cluster distances without anchors, satisfying both positive and negative constraints while enabling inference of the true cluster number from this geometric configuration.
Moreover, our approach is supported by a theoretical framework that guarantees optimal performance under specific conditions, eliminates the need for manual tuning, and determines the embedding space dimension without trial and error.

\paragraph{Deep angular learning.} 
Deep learning methods in angular space have been extensively studied for supervised classification tasks \cite{L-softmax,SphereFace,DHL,CosFace,ArcFace} like face recognition, where each instance has a label.
In contrast, CC involves defining pairwise relationships with sparse constraints applied only to subsets of data, presenting a unique challenge of learning representations while satisfying incomplete constraints without explicit labels.
Some supervised prototype-based methods \cite{proto-NIPS,proto-CVPR2,proto-ICML,proto-CVPR1,proto-TNNLS,proto-ACML} have explored angular output spaces by guiding instances to converge around equidistant prototypes, benefiting aspects of supervised tasks such as boundary discriminability for imbalanced classes and the alignment of Euclidean and cosine metrics \cite{proto-TNNLS}.
However, these approaches rely on distances to class centers, making them unsuitable for instance-level pairwise learning in CC. 
Moreover, enforcing class margins does not help resolve complex pairwise constraints that cause conflicts during embedding learning.
In contrast, our proposed method is the first to apply deep angular learning to CC.
By focusing exclusively on the angles between feature vectors (see Fig.~\ref{fig:z_p_space}(d)), we establish equal inter-cluster distances without using anchors, effectively satisfying both positive and negative constraints while also generalizing well to unseen instances. 
Our approach leverages the closed nature of angular space to prevent constraint conflicts and is supported by a theoretical foundation.

\section{SpherePair constraint clustering}
\label{sec:spherepair}

CC aims to partition a dataset $\mathcal{X} = \{\boldsymbol{x}_j\}_{j=1}^{|\mathcal{X}|}$ into $K$ clusters $\mathcal{S} = \{\mathcal{S}_k\}_{k=1}^K$ while satisfying pairwise constraints $\mathcal{C} = \{(a_i, b_i, y_i)\}_{i=1}^{|\mathcal{C}|}$, where each constraint $(a_i, b_i, y_i)$ requires that instances $\boldsymbol{x}_{a_i}$ and $\boldsymbol{x}_{b_i}$ be in the same cluster if $y_i = 1$, or in different clusters if $y_i = 0$.
To avoid the reliance on anchors inherent to end-to-end DCC, we learn a constrained yet clustering-friendly representation $\mathcal{Z} \subset \mathbb{R}^D$ to determine $\mathcal{S}$, constituting deep constraint embedding.
Distinct from existing approaches in Euclidean space, we propose angular embedding to effectively preserve pairwise distances and eliminate the need for complex hyperparameter tuning.
To facilitate the learning of $\mathcal{Z}$, we adopt a deep autoencoder with encoder $f_{\boldsymbol{\phi}}: \mathcal{X} \to \mathcal{Z}$ and decoder $g_{\boldsymbol{\phi}'}: \mathcal{Z} \to \mathcal{X}$, parameterized by $\boldsymbol{\phi}$ and $\boldsymbol{\phi}'$, respectively.

\paragraph{SpherePair loss.}
We formulate an anchor-free pairwise loss based on angular distance, optimizing the encoder $f_{\boldsymbol{\phi}}$ to generate latent representations $\mathcal{Z}$ aligned with constraints $\mathcal{C}$ in angular space. 
For each constrained pair $\boldsymbol{z}_{a_i}, \boldsymbol{z}_{b_i} \in \mathcal{Z}$, the angle $\theta_{z_{a_i}, z_{b_i}} \in [0, \pi]$ is normalized to a similarity score $\text{Sim}(a_i, b_i) \in [0,1]$ for constraint embedding using logistic loss, promoting angular similarity and separation for positive and negative pairs. The resulting SpherePair loss, $\mathcal{L}_{\text{ang}}$, is defined as:
\begin{equation}
\begin{split}
    \mathcal{L}_{\text{ang}} = -\frac{1}{|\mathcal{C}|} \sum_{i=1}^{|\mathcal{C}|} \Big( & y_i \log \text{Sim}(a_i, b_i) + (1 - y_i) \log (1 - \text{Sim}(a_i, b_i)) \Big)
\end{split}
\label{eq:angular-pairs},
\end{equation}

\[
    \text{Sim}(a_i, b_i) = \frac{1}{2} 
    \begin{cases}
    \cos(\theta_{\boldsymbol{z}_{a_i}, \boldsymbol{z}_{b_i}}) + 1, &\text{if } y_i = 1, \\
    \cos(\min(\omega \theta_{\boldsymbol{z}_{a_i}, \boldsymbol{z}_{b_i}}, \pi)) + 1, &\text{if } y_i = 0.
    \end{cases}
\]

Here, \(\omega\) is an angular factor that ensures sufficient separation between clusters in the embedding space. Our proposed loss promotes constrained embedding learning in the angular space by enforcing the following: 
(i) for \((a_i, b_i, 1) \in \mathcal{C}\), smaller angles \(\theta_{\boldsymbol{z}_{a_i}, \boldsymbol{z}_{b_i}}\) are favored to emphasize similarity; and 
(ii) for \((a_i, b_i, 0) \in \mathcal{C}\), a \textit{negative zone} of angular size \(\frac{\pi}{\omega}\) regulates the spacing of dissimilar pairs. 
The optimal negative-zone factor $\omega$, theoretically determined in \cref{sec:theoretical}, mitigates conflicts among negative pairs in the embedding while ensuring sufficient separation.
Notably, our SpherePair loss ensures a bounded angular distance \([0, \pi]\), providing stable similarity mapping and avoiding normalization issues associated with unbounded Euclidean distances \([0, +\infty)\) \cite{SDEC,CPAC}.

\paragraph{Regularization and learning.}
While the SpherePair loss $\mathcal{L}_{\text{ang}}$ aligns $\mathcal{Z}$ with constraints $\mathcal{C}$, minimizing it directly may lead to degenerate representations that fail to capture intrinsic cluster structures. Inspired by deep clustering methods \cite{deepclustering-kmeans, IDEC}, we incorporate a reconstruction loss:
\begin{equation}
    \mathcal{L}_{\text{recon}} = \frac{1}{|\mathcal{X}|} \sum_{j=1}^{|\mathcal{X}|} \|\boldsymbol{x}_j - \hat{\boldsymbol{x}}_j\|_2^2.
    \label{eq:L_recon}
\end{equation}
Unlike DCC methods that directly construct latent representations \cite{CIDEC1,CDC,scDCC}, our autoencoder enforces instance reconstruction from normalized angular latent embeddings. 
To preserve angular properties during regularization, latent embeddings are normalized prior to decoding:
\begin{equation}
    \hat{\boldsymbol{x}}_j = g_{\boldsymbol{\phi}'}(\text{Norm}(f_{\boldsymbol{\phi}}(\boldsymbol{x}_j))),
    \label{eq:x_j_hat}
\end{equation}
where $\text{Norm}(\cdot)$ ensures unit-length embeddings, preserving angular information. Thus,
our overall objective for deep constraint embedding is:
\begin{equation}
    \mathcal{L} = \mathcal{L}_{\text{ang}} + \lambda \mathcal{L}_{\text{recon}},
    \label{eq:L}
\end{equation}
where trade-off factor $\lambda$ balances the angular loss in Eq.~\ref{eq:angular-pairs}  and reconstruction loss in Eq.~\ref{eq:L_recon}.

\begin{wrapfigure}{r}{0.53\textwidth}
    \centering
    \vskip -0.2in
    \includegraphics[width=0.48\textwidth]{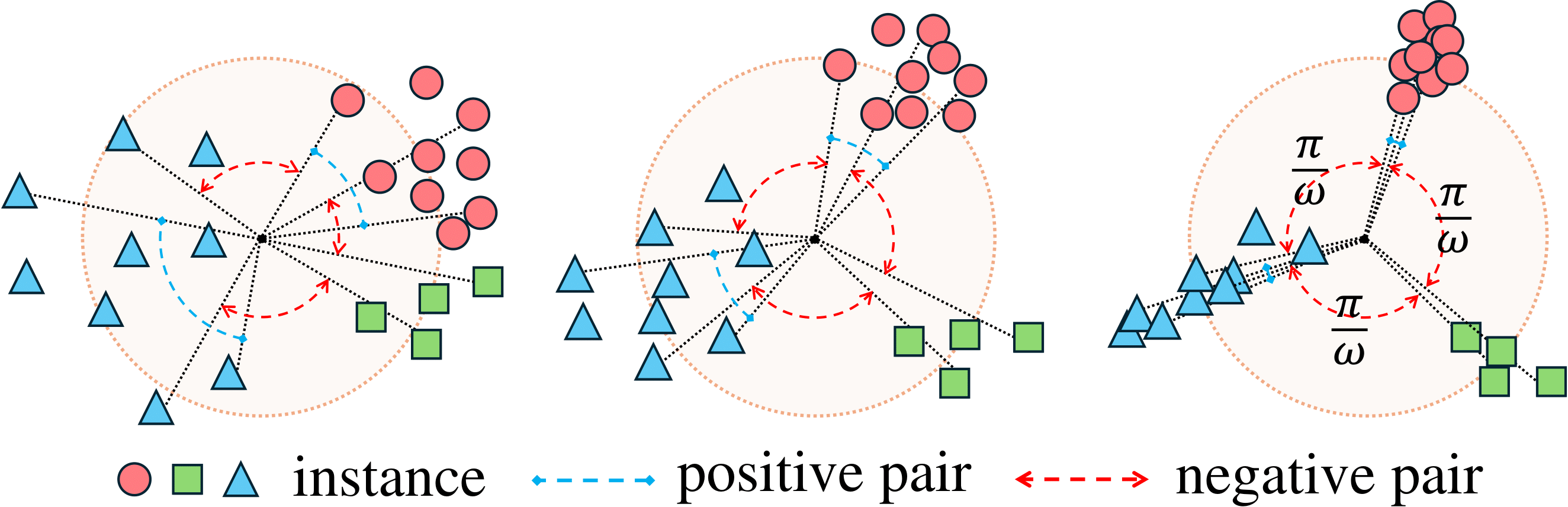}
    \vskip -0.05in
    \caption{$\mathcal{Z}$ change in the SpherePair embedding learning (from left to right): the angular distances of positive pairs decrease, while those of negative pairs gradually adhere to the negative zone $\frac{\pi}{\omega}$.}
    \label{fig:SpherePairs}
\end{wrapfigure}

Minimizing the overall loss in Eq.~\ref{eq:L} yields the optimal angular embeddings, $\mathcal{Z}^*$. As illustrated in Fig.~\ref{fig:SpherePairs}, our angular constraint embedding learning compacts instances within the same cluster while separating different clusters using the negative zone in a (hyper)sphere. This structure simplifies subsequent clustering. Consequently, we name our deep constraint embedding framework \textit{SpherePair}, and the resulting embeddings are spherical representations:
\begin{equation}
    \mathcal{Z}_{\text{sphere}} = \{\text{Norm}(\boldsymbol{z}_j^*) \mid \boldsymbol{z}_j^* \in \mathcal{Z}^*\},
    \qquad
    \boldsymbol{z}_j^* = f_{\boldsymbol{\phi}^*}(\boldsymbol{x}_j).
    \label{eq:Z_sphere}
\end{equation}

\paragraph{Deployment.}
Applying an unsupervised clustering method to these spherical representations completes the CC process and enables generalization to unseen data. 
Note that for normalized features in $\mathcal{Z}_{\text{sphere}}$, the cosine and Euclidean distances can be shown to be equivalent \cite{barber2012bayesian}, so either metric can be used for clustering.
The SpherePair CC algorithm is outlined in \cref{alg:SpherePairs} in \cref{sec:alg}.

\paragraph{Inferring cluster numbers.}
If the number of clusters is unknown, it can be inferred from the pre-learned $\mathcal{Z}_{\text{sphere}}$ via its geometric properties, thereby avoiding both retraining of the latent embedding as required by end-to-end methods and cumbersome post-clustering validation \cite{everitt2011cluster}. 
By applying principal component analysis (PCA) \cite{PCA} to the training data involved in the negative constraints in \(\mathcal{Z}_{\text{sphere}}\), we obtain all top-\(d\) subspace projections. 
In each subspace, we compute inter-cluster angles, then select the smallest \(\rho\)-fraction \((0<\rho\ll1)\) of these angles and compute their tail average. With the theoretical justification presented in \cref{sec:theoretical}, we infer the number of clusters by identifying the onset of the plateau in the sequence of mean inter-cluster angles across subspaces. The cluster number inference algorithm is presented as \cref{alg:inferK} in \cref{sec:alg}.


\section{Theoretical foundation}
\label{sec:theoretical}

We establish a theoretical foundation for SpherePair to rigorously determine the negative-zone factor $\omega$ and embedding dimension $D$, which are indispensable for the unique properties of our angular constraint embedding,
and further enable our inference of the unknown cluster number $K$. 

To fix the negative-zone factor $\omega$ and prevent conflicts in negative pair embedding, we first establish the \emph{conflict-free condition} for an optimal angular representation:

\begin{proposition}[Conflict-free]
\label{prop:z^*}
Let $\mathcal{S}^* = \{\mathcal{S}_k^*\}_{k=1}^K$ be the ground-truth partition of $\mathcal{X} = \{\boldsymbol{x}_j\}_{j=1}^{|\mathcal{X}|}$. An optimal angular representation $\mathcal{Z}^* = \{\boldsymbol{z}_j^*\}_{j=1}^{|\mathcal{X}|} \subset \mathbb{R}^D$ achieves $\mathcal{L}_{\mathrm{ang}}=0$.
To ensure no conflicts in negative pairs, for any $\boldsymbol{x}_j \in \mathcal{S}_k^*$ and $\boldsymbol{x}_{j'} \in \mathcal{S}_{k'}^*$, $\theta_{\boldsymbol{z}_j^*,\boldsymbol{z}_{j'}^*}=0$ if $k=k'$, and $\theta_{\boldsymbol{z}_j^*,\boldsymbol{z}_{j'}^*}\ge \frac{\pi}{\omega}$ if $k\neq k'$.
\end{proposition}

Proof of \cref{prop:z^*} is given in \cref{sec:proof1}, based on which we further constrain $\omega$ as follows:

\begin{proposition}[Equidistance]
\label{prop:requirement}
Given $\mathcal{X} = \{\boldsymbol{x}_j\}_{j=1}^{|\mathcal{X}|}$ with ground-truth partition
$\mathcal{S}^* = \{\mathcal{S}_k^*\}_{k=1}^K$, and a factor $\omega$ satisfying the conflict-free
condition of $\mathcal{Z}^* = \{\boldsymbol{z}_j^*\}_{j=1}^{|\mathcal{X}|} \subset \mathbb{R}^D$, 
then for each constraint $(a_i,b_i,y_i)$ from $\mathcal{S}^*$, the angle
$\theta_{\boldsymbol{z}_{a_i}^*, \boldsymbol{z}_{b_i}^*}$ is uniquely determined by
$(a_i,b_i,y_i)$ if and only if: 
(i) $\mathcal{Z}^*$ is equidistant among clusters, i.e., all cross-cluster angles
  $\{\theta_{\boldsymbol{z}^*_j, \boldsymbol{z}^*_{j'}}\}$ are the same for any
  $\boldsymbol{x}_j\in \mathcal{S}_k^*$ and $\boldsymbol{x}_{j'}\in \mathcal{S}_{k'\neq k}^*$;
  (ii) $\omega$ matches this unique angular separation in (i), i.e.\ $\omega=\omega^*$,
  where $\theta_{\boldsymbol{z}^*_j, \boldsymbol{z}^*_{j'}} = \frac{\pi}{\omega^*}$.
\end{proposition}

The proof is given in \cref{sec:proof2}. 
An $\omega$ satisfying \cref{prop:requirement} also meets the conflict-free condition of \cref{prop:z^*}, ensuring each local pairwise constraint. 
Moreover, it enforces equidistant cluster embeddings, reflecting the fairness of all negative relationships $\{(a_i,b_i,0)\}\subseteq\mathcal{C}$ and thus balancing pairwise relations. 
While \cref{prop:z^*,prop:requirement} describe the ideal geometric configuration under $\mathcal{L}_{\text{ang}}=0$, the following corollary complements them via perturbation analysis:

\begin{corollary}[Geometric Deviations under Near-zero Residual Loss]
\label{corollary:perturbation}
Given a set of $|\mathcal{C}|$ constraints $\mathcal{C} = \{(a_i, b_i, y_i)\}_{i=1}^{|\mathcal{C}|}$, and average angular loss $\mathcal{L}_{\mathrm{ang}}\le \varepsilon$ for some $0<\varepsilon\ll 1$, then:
(i) For each positive constraint $(a_i,b_i,1)\in\mathcal{C}$ with $\theta_{z_{a_i},z_{b_i}}$, we have $0 \le \theta_{z_{a_i},z_{b_i}} \le \Delta^+(\varepsilon) := \arccos\!\big(2e^{-|\mathcal{C}|\varepsilon}-1\big) \approx 2\sqrt{|\mathcal{C}|\varepsilon}$;
(ii) For each negative constraint $(a_i,b_i,0)\in\mathcal{C}$ with $\theta_{z_{a_i},z_{b_i}} \le \frac{\pi}{\omega}$, we have $0 \le \tfrac{\pi}{\omega} - \theta_{z_{a_i},z_{b_i}} \le \Delta^-(\varepsilon) := \frac{1}{\omega} \arccos\!\big(1-2e^{-|\mathcal{C}|\varepsilon}\big) \approx 2\sqrt{|\mathcal{C}|\varepsilon} \big/ \omega.$
\end{corollary}

The proof of \cref{corollary:perturbation} is given in \cref{sec:proof3}. \cref{corollary:perturbation} shows that the geometric deviations degrade gracefully as $\mathcal{O}(\sqrt{\varepsilon})$ in the limit $\varepsilon \to 0$, approximately preserving the ideal configuration of Propositions \ref{prop:z^*} and \ref{prop:requirement} under small residual loss.

Despite only such tiny deviations from the ideal embedding, a valid $\omega$ required by \cref{prop:requirement} is not always guaranteed, since an equidistant cluster arrangement $\mathcal{Z}^*$ may fail in arbitrary dimension $D$.
The feasibility and bounds of such $\omega$ depend on $D$ and the number of clusters $K$. 
The following theorem provides conditions for the existence of a valid $\omega$ meeting
\cref{prop:requirement}:

\begin{theorem}[Existence of Valid $\omega$]
\label{thm:bound}
Given $\mathcal{X}$ with the ground truth partition $\mathcal{S}^* = \{\mathcal{S}_k^*\}_{k=1}^K$ containing $K$ clusters. Suppose we seek an $\omega$ that matches a $\mathcal{Z}^* \subset \mathbb{R}^D$ with equidistant clusters, as formalized in \cref{prop:requirement}. Then we have:
(i) When $D < K-1$, such a valid $\omega$ does not exist;
(ii) When $D = K -1$, the unique valid $\omega$ is $\pi \big/ \arccos(-\frac{1}{K-1})$;
(iii) When $D \geq K$, the range of valid $\omega$ values is relaxed to $\omega \geq \pi \big/ \arccos(-\frac{1}{K-1})$.
\end{theorem}

Proofs of \cref{thm:bound} are in \cref{sec:proof4}. For each $D$, \cref{thm:bound} establishes the existence conditions and bounds of $\omega$ that satisfy \cref{prop:requirement}. With such $\omega$, our SpherePair embedding learning drives $\mathcal{Z}$ to converge to $\mathcal{Z}_{\text{sphere}}$, where $K$ clusters
form a regular simplex in a $(K-1)$-dimensional subspace when $D\ge K-1$
(\cref{sec:3D_visualization_appendix} offers a 3D visualization of this convergence based on a real-world dataset).
While any $\omega$ within the range specified by \cref{thm:bound} is admissible, we further restrict the optimal setting through the following corollary:

\begin{corollary}[Minimal Admissible $\omega$]
\label{corollary:optimal_omega}
Given $\mathcal{X}$ with the ground truth partition $\mathcal{S}^* = \{\mathcal{S}_k^*\}_{k=1}^K$ containing $K$ clusters, and an embedding space $\mathbb{R}^D$ with sufficiently large $D \ge K$, the minimal admissible $\omega$ satisfying the validity condition of \cref{thm:bound}, denoted by $\omega^{*}_{\mathrm{min}}(K)$, is bounded as $1 \le \omega^{*}_{\mathrm{min}}(K) < 2$ for all $K > 1$, and is monotone increasing in $K$ with  $\lim_{K \to \infty} \omega^{*}_{\mathrm{min}}(K) = 2$.
\end{corollary}

The proof of \cref{corollary:optimal_omega} is provided in \cref{sec:proof5}. In pursuit of a larger negative-zone (given by $\frac{\pi}{\omega}$) for enhanced inter-cluster separability via a smaller $\omega$, \cref{corollary:optimal_omega} prescribes $\omega = 2$ as the optimal setting for sufficiently large $D$, universally valid for any $K$. This theoretically fixes the choice of $\omega$, leaving the attainment of conflict-free embeddings governed solely by $D$, with the lenient bound $D \geq K$ readily satisfied even when $K$ is not precisely known. Therefore, our theoretical guarantees for $\omega$ and $D$ in SpherePair CC eliminate the need for manual Euclidean margin tuning and strict embedding-dimension calibration \cite{AutoEmbedder}, ensuring consistent scaling.

If the cluster number $K$ is unknown, given the optimal $\mathcal{Z}^*$ (or $\mathcal{Z}_{\mathrm{sphere}}$) learned by \cref{alg:SpherePairs} with valid $\omega$ and $D$ (cf.\ \cref{thm:bound}), $K$ can be inferred from its theoretically supported geometry: $\mathcal{Z}_{\mathrm{sphere}}$ converges to a $K$-vertex regular simplex in a $(K{-}1)$-dimensional subspace, and its projection onto this subspace preserves the same configuration. 
This implies that the intrinsic embedding dimension $d^\ast=K{-}1$ is directly linked to the true cluster number. To exploit this link, we employ PCA and have the following theorem:

\begin{theorem}[Pairwise-angle Invariance]
\label{thm:pca-angle-invariance}
Given $\mathcal{X}=\{\boldsymbol{x}_j\}_{j=1}^{|\mathcal{X}|}$ with ground-truth partition $\mathcal{S}^*=\{\mathcal{S}_k^*\}_{k=1}^K$ and an optimal $\mathcal{Z}^*=\{\boldsymbol{z}_j^*\}_{j=1}^{|\mathcal{X}|}\subset\mathbb{R}^D$ with equidistant clusters satisfying \cref{prop:z^*,prop:requirement}. For $d\in\{1,\dots,D\}$, let $\mathcal{Z}_{\mathrm{pca}}^{(d)}=\{\tilde{\boldsymbol{z}}_j^{(d)}\}_{j=1}^{|\mathcal{X}|}\subset\mathbb{R}^d$ be the $d$-dimensional PCA projection of $\mathcal{Z}_{\mathrm{sphere}}=\{\operatorname{Norm}(\boldsymbol{z}_j^*)\}_{j=1}^{|\mathcal{X}|}$, and denote by $\theta_{\tilde{\boldsymbol{z}}_{j}^{(d)}\!,\tilde{\boldsymbol{z}}_{j'}^{(d)}}$ the pairwise angle between $\tilde{\boldsymbol{z}}_j^{(d)}\!\neq\!\mathbf{0}$ and $\tilde{\boldsymbol{z}}_{j'}^{(d)}\!\neq\!\mathbf{0}$.
Then:
(i) For every $d,d' \ge K-1$ and all such pairs $(j,j')$, 
$\theta_{\tilde{\boldsymbol{z}}_{j}^{(d)},\tilde{\boldsymbol{z}}_{j'}^{(d)}}
= \theta_{\tilde{\boldsymbol{z}}_{j}^{(d')},\tilde{\boldsymbol{z}}_{j'}^{(d')}}$;
(ii) For any $d,d'<K-1$, the cross-dimensional invariance in (i) cannot hold for arbitrary pairs $(j,j')$.
\end{theorem}

The proof is in \cref{sec:proof6}. \Cref{thm:pca-angle-invariance} shows that the cross-$d$ invariance of pairwise angles $\theta_{\tilde{\boldsymbol{z}}_{j}^{(d)}\!,\tilde{\boldsymbol{z}}_{j'}^{(d)}}$ determines $d^\ast$. To monitor this invariance, we consider the \textit{minimal inter-cluster angle},
\[
\delta_d = \min\{\theta_{\tilde{\boldsymbol{z}}_{j}^{(d)}\!,\tilde{\boldsymbol{z}}_{j'}^{(d)}} \mid \boldsymbol{x}_j\in\mathcal{S}_k^*,\ \boldsymbol{x}_{j'}\in\mathcal{S}_{k'\neq k}^*\},
\]
for which we have the following corollary:

\begin{corollary}[$\delta_d$ Invariance]
\label{corollary:min-neg-angle}
Under the same conditions as in \cref{thm:pca-angle-invariance}, and define the cluster frequencies
$p_k=\frac{|\mathcal{S}_k^*|}{{|\mathcal{X}|}}>0$ with $\sum_{k=1}^K p_k=1$. Then:
(i) If $K=2$, the minimal inter-cluster angle $\delta_d = \pi,~ \forall d \ge 1$;
(ii) If $K>2$, then $\delta_1=0$, and there exists a constant $\delta_\star \in (\frac{\pi}{3}, \arccos(-\frac{1}{K-1})]$ such that $\delta_d=\delta_\star$ always holds when $d\ge K-1$.
The upper bound $\arccos(-\frac{1}{K-1})$ of $\delta_\star$ is attained when $p_1=p_2=\!\cdots\!=p_K$, while the lower bound $\frac{\pi}{3}$ is approached when some $p_k\to 1$.
\end{corollary}

Its proof is in \cref{sec:proof7}. \Cref{corollary:min-neg-angle} offers a practical reflection of the invariance in \cref{thm:pca-angle-invariance} through $\delta_d$, which helps determine $d^\ast$, and hence $K$. Concretely, for $d=1,\dots,K{-}1$, $\delta_d$ increases from $0$ (when $K>2$) and reaches some $\delta_\star > \tfrac{\pi}{3}$ at $d=K{-}1$; for $d\ge K{-}1$, $\delta_d$ stabilizes around this $\delta_\star$. Thus, the onset of the plateau of sequence $\{\delta_d\}_{d=1}^D$ identifies $d^\ast=K{-}1$.

In practice, under the assumption that the training negative constraints $\mathcal{C}^-$ cover all true clusters, $\delta_d$ can be readily computed from $\mathcal{C}^-$, 
and due to small deviations (cf.\ \cref{corollary:perturbation}), we replace $\delta_d$ by a tail-averaged variant $\overline{\delta}_d$ for stability. 
The sequence $\{\overline{\delta}_d\}_{d=1}^D$ is then used to locate the plateau entry $d^\ast$ and estimate $\widehat{K}=d^\ast+1$, 
which provides a theoretical foundation for \cref{alg:inferK}.


\section{Experiments}
\label{sec:experiments}

\subsection{Experimental settings}
\label{subsect:ex-setting}
Our experimental settings aim to address the following questions:  
(i) How does our SpherePair perform compared to state-of-the-art DCC methods?  
(ii) How well do our SpherePair and baseline methods capture consistent instance relations under imbalanced constraint distributions? 
(iii) How effectively does our approach handle unknown cluster numbers?
(iv) How are our theoretical insights empirically supported, and how sensitive is our approach to the introduced hyperparameters?

\paragraph{Datasets.}
We adopt eight benchmarks with diverse class counts and class balance: CIFAR-100-20 and CIFAR-10 \cite{CIFAR}, FashionMNIST \cite{FMNIST}, ImageNet-10 \cite{Imagenet10}, MNIST \cite{MNIST}, STL-10 \cite{STL10}, together with two imbalanced text datasets, Reuters subset \cite{DEC} and RCV1-10 (see \cref{sec:datasets_appendix} for details).

\paragraph{Baselines.} We evaluate our SpherePair against three categories of DCC methods: 
(i) state-of-the-art end-to-end approaches, 
including VanillaDCC \cite{MCL}, VolMaxDCC \cite{VolMaxDCC}, DCGMM \cite{DCGMM}, and CIDEC \cite{CIDEC1}; 
(ii) SDEC \cite{SDEC}, which integrates Euclidean constraint embedding loss into end-to-end deep clustering;
and (iii) AutoEmbedder \cite{AutoEmbedder}, a fully anchor-free Eulidean constraint embedding method.
For AutoEmbedder and SpherePair, K-means is applied to their learned representations for clustering unless otherwise specified.   
These baselines encompass the key advancements in DCC research.

\paragraph{Protocol.} For FashionMNIST, MNIST, and the Reuters subset, we use the original pre-split training and test data settings. For the remaining benchmarks, we randomly split the data into 80\% training and 20\% test sets. 
Consistent with \cite{VolMaxDCC}, we reserve a validation set of 1,000 instances from the training data to optimise the hyperparameters for baselines requiring such tuning. 
Constraints are generated based on the ground-truth labels of pairs sampled within the training sets.
For a comparative study, performance with different constraint set sizes (1k/5k/10k) is evaluated on training and test sets over five trials using three standard clustering metrics: \textit{Accuracy} (ACC), \textit{Normalised Mutual Information} (NMI), and \textit{Adjusted Rand Index} (ARI).
To assess performance under imbalanced constraints, we create a balanced set, {\tt IMB0}, via uniform sampling and gradually introduce additional constraints linked to fewer clusters to form two imbalanced sets, {\tt IMB1} and {\tt IMB2}, where {\tt IMB0} $\subset$ {\tt IMB1} $\subset$ {\tt IMB2} 
(see \cref{sec:protocol_appendix} for detailed constraint generation procedure).
To examine the effectiveness of cluster number inference, we repeat the procedure using embeddings pre-learned from five random initializations.
To empirically validate our theoretical insights and assess the robustness of our approach, we explore a wide range of $D$, $\lambda$, and $\rho$ settings.

\paragraph{Implementation.} 
We strictly follow the same fully connected architectures from baseline papers \cite{SDEC,CIDEC1,DCGMM,VolMaxDCC} for fair comparison and compatibility with both image and non-image datasets.
For DCGMM, CIDEC, SDEC, AutoEmbedder, and SpherePair, we use a fully connected encoder with hidden layers of size 500--500--2000 (and a symmetric decoder when required), and an embedding layer of $D=20$ for CIFAR-100-20 and $D=10$ for the remaining datasets, unless stated otherwise. 
For VanillaDCC and VolMaxDCC, we use a fully connected network with two hidden layers of size 512--512 and a classification layer matching the number of clusters, $K$, as recommended in \cite{VolMaxDCC}.
ReLU activations are used across all networks. 
Pretrained autoencoders are employed for model initialisation, except for VanillaDCC and VolMaxDCC. Specifically, a variational autoencoder is pretrained for DCGMM, while stacked denoising autoencoders are pretrained layer-wise for other models. Pretraining is performed unsupervised on the entire training set.  
In SpherePair and cluster number inference, $\omega$ is theoretically fixed at $2$ as per Sect.~\ref{sec:theoretical}, while $\lambda = 0.02$ and $\rho = 0.05$ are used by default unless varied for hyperparameter robustness evaluation.
For baselines, we adopt reported optimal hyperparameters (VanillaDCC, DCGMM, CIDEC, SDEC) or follow the search procedures in VolMaxDCC and AutoEmbedder.
SpherePair and all baselines (except DCGMM, where we use the authors' source code) are implemented in PyTorch 1.5.1. Training is conducted using the Adam optimizer, except for SDEC and VolMaxDCC, which employs SGD as suggested by their authors. 

More details of our experimental settings are provided in \cref{sec:setting_appendix} to ensure full replicability.

\subsection{Experimental results}
\label{subsect:ex-res}

\begin{table}[!h]
    \caption{Comparative performance (\%) (ACC, NMI, ARI) across datasets for models with 1k/5k/10k constraints. \textcolor{blue}{Blue} and black represent \textcolor{blue}{training} and test results, respectively. Best results are in \textbf{bold}, second-best are \underline{underlined}, and \(^{\dagger}\) indicates models without pretraining.}
    \label{table:SpherePair_VS_baselines}
    \begin{center}
    \begin{scriptsize}
    \renewcommand{\arraystretch}{0.0} 
    \setlength{\tabcolsep}{3.1pt}       
    \setlength{\aboverulesep}{0.6pt}
    \setlength{\belowrulesep}{0.6pt}
    \setlength{\extrarowheight}{0pt}
    \begin{tabular}{lllccccccccc}
        \toprule
         &  &  & \makecell{Vanilla-\\DCC} & \makecell{VolMax-\\DCC} & CIDEC\(^{\dagger}\) & CIDEC & DCGMM & SDEC & \makecell{Auto-\\Embedder} & \makecell{SpherePair\(^{\dagger}\)\\(Ours)} & \makecell{SpherePair\\(Ours)} \\   
        \midrule
        \multirow{9}{*}{\raisebox{-3ex}{CIFAR100-20}}   
        & \multirow{3}{*}{1k}  
          & ACC 
            & \textcolor{blue}{34.2}, 34.3 
            & \textcolor{blue}{20.1}, 20.3
            & \textcolor{blue}{32.8}, 33.0
            & \textcolor{blue}{\underline{46.6}}, \underline{46.2}
            & \textcolor{blue}{44.5}, 44.2
            & \textcolor{blue}{45.7}, 45.4
            & \textcolor{blue}{21.5}, 21.6
            & \textcolor{blue}{45.1}, 45.1
            & \textcolor{blue}{\textbf{48.3}}, \textbf{48.2} \\
          & 
          & NMI 
            & \textcolor{blue}{36.0}, 36.3
            & \textcolor{blue}{21.4}, 21.6
            & \textcolor{blue}{35.1}, 35.6
            & \textcolor{blue}{\underline{47.3}}, \underline{47.9}
            & \textcolor{blue}{44.9}, 45.4
            & \textcolor{blue}{47.0}, 47.5
            & \textcolor{blue}{23.1}, 23.4
            & \textcolor{blue}{44.9}, 45.4
            & \textcolor{blue}{\textbf{47.7}}, \textbf{48.0} \\
          & 
          & ARI 
            & \textcolor{blue}{19.3}, 19.3
            & \textcolor{blue}{7.1}, 7.2
            & \textcolor{blue}{19.7}, 19.6
            & \textcolor{blue}{\underline{30.0}}, \underline{29.9}
            & \textcolor{blue}{28.7}, 28.7
            & \textcolor{blue}{29.0}, 29.2
            & \textcolor{blue}{7.1}, 7.1
            & \textcolor{blue}{29.4}, 29.5
            & \textcolor{blue}{\textbf{32.2}}, \textbf{32.4} \\
        \cmidrule(lr){2-12}
        & \multirow{3}{*}{5k}  
          & ACC 
            & \textcolor{blue}{47.4}, 47.4
            & \textcolor{blue}{42.8}, 42.8
            & \textcolor{blue}{42.3}, 42.1
            & \textcolor{blue}{46.7}, 46.1
            & \textcolor{blue}{48.1}, 47.9
            & \textcolor{blue}{45.6}, 45.1
            & \textcolor{blue}{13.8}, 14.2
            & \textcolor{blue}{\underline{55.4}}, \underline{55.7}
            & \textcolor{blue}{\textbf{59.0}}, \textbf{58.8} \\
          & 
          & NMI 
            & \textcolor{blue}{46.7}, 47.1
            & \textcolor{blue}{41.9}, 42.1
            & \textcolor{blue}{42.3}, 42.5
            & \textcolor{blue}{45.4}, 45.7
            & \textcolor{blue}{46.7}, 47.1
            & \textcolor{blue}{47.0}, 47.5
            & \textcolor{blue}{13.5}, 13.8
            & \textcolor{blue}{\underline{51.1}}, \underline{51.7}
            & \textcolor{blue}{\textbf{52.6}}, \textbf{53.0} \\
          & 
          & ARI 
            & \textcolor{blue}{32.2}, 32.2
            & \textcolor{blue}{22.8}, 22.8
            & \textcolor{blue}{27.1}, 26.8
            & \textcolor{blue}{30.3}, 29.6
            & \textcolor{blue}{32.2}, 32.2
            & \textcolor{blue}{29.2}, 29.3
            & \textcolor{blue}{4.7}, 4.7
            & \textcolor{blue}{\underline{39.1}}, \underline{39.4}
            & \textcolor{blue}{\textbf{41.0}}, \textbf{40.9} \\
        \cmidrule(lr){2-12}
        & \multirow{3}{*}{10k} 
          & ACC 
            & \textcolor{blue}{54.6}, 54.5
            & \textcolor{blue}{51.2}, 51.0
            & \textcolor{blue}{49.8}, 49.8
            & \textcolor{blue}{50.9}, 50.1
            & \textcolor{blue}{52.3}, 52.1
            & \textcolor{blue}{45.7}, 45.2
            & \textcolor{blue}{31.3}, 31.3
            & \textcolor{blue}{\underline{60.5}}, \underline{60.4}
            & \textcolor{blue}{\textbf{62.8}}, \textbf{62.6} \\
          & 
          & NMI 
            & \textcolor{blue}{50.2}, 50.3
            & \textcolor{blue}{48.5}, 48.7
            & \textcolor{blue}{47.4}, 47.6
            & \textcolor{blue}{48.5}, 48.
            & \textcolor{blue}{49.2}, 49.6
            & \textcolor{blue}{47.1}, 47.7
            & \textcolor{blue}{36.6}, 36.9
            & \textcolor{blue}{\underline{53.9}}, \underline{54.3}
            & \textcolor{blue}{\textbf{55.1}}, \textbf{55.5} \\
          & 
          & ARI 
            & \textcolor{blue}{37.9}, 37.6
            & \textcolor{blue}{33.4}, 33.3
            & \textcolor{blue}{33.4}, 33.2
            & \textcolor{blue}{34.0}, 33.0
            & \textcolor{blue}{36.7}, 36.7
            & \textcolor{blue}{29.3}, 29.5
            & \textcolor{blue}{20.6}, 20.4
            & \textcolor{blue}{\underline{43.4}}, \underline{43.4}
            & \textcolor{blue}{\textbf{45.3}}, \textbf{45.2} \\
        \midrule
        \multirow{9}{*}{\raisebox{-3ex}{CIFAR10}}   
        & \multirow{3}{*}{1k}  
          & ACC 
            & \textcolor{blue}{70.2}, 70.1
            & \textcolor{blue}{65.2}, 64.9
            & \textcolor{blue}{64.9}, 65.1
            & \textcolor{blue}{\textbf{86.5}}, \textbf{86.5}
            & \textcolor{blue}{82.1}, 82.1
            & \textcolor{blue}{84.0}, 84.1
            & \textcolor{blue}{58.2}, 58.5
            & \textcolor{blue}{84.3}, 84.2
            & \textcolor{blue}{\underline{85.7}}, \underline{85.6} \\
          & 
          & NMI 
            & \textcolor{blue}{67.0}, 66.9
            & \textcolor{blue}{62.5}, 62.4
            & \textcolor{blue}{60.2}, 60.3
            & \textcolor{blue}{\textbf{78.8}}, \textbf{78.9}
            & \textcolor{blue}{75.0}, 74.9
            & \textcolor{blue}{76.5}, 76.5
            & \textcolor{blue}{57.8}, 58.1
            & \textcolor{blue}{75.6}, 75.4
            & \textcolor{blue}{\underline{77.3}}, \underline{77.1} \\
          & 
          & ARI 
            & \textcolor{blue}{57.8}, 57.6
            & \textcolor{blue}{48.6}, 48.3
            & \textcolor{blue}{50.1}, 50.3
            & \textcolor{blue}{\textbf{75.2}}, \textbf{75.1}
            & \textcolor{blue}{69.7}, 69.6
            & \textcolor{blue}{70.5}, 70.6
            & \textcolor{blue}{43.1}, 43.3
            & \textcolor{blue}{71.7}, 71.5
            & \textcolor{blue}{\underline{74.0}}, \underline{73.7} \\
        \cmidrule(lr){2-12}
        & \multirow{3}{*}{5k}  
          & ACC 
            & \textcolor{blue}{87.6}, 87.3
            & \textcolor{blue}{84.9}, 84.6
            & \textcolor{blue}{86.4}, 86.2
            & \textcolor{blue}{\underline{88.9}}, \underline{88.7}
            & \textcolor{blue}{88.3}, 88.0
            & \textcolor{blue}{85.4}, 85.5
            & \textcolor{blue}{85.9}, 85.8
            & \textcolor{blue}{\underline{88.9}}, \underline{88.7}
            & \textcolor{blue}{\textbf{89.2}}, \textbf{88.9} \\
          & 
          & NMI 
            & \textcolor{blue}{79.3}, 79.0
            & \textcolor{blue}{79.0}, 78.6
            & \textcolor{blue}{78.3}, 78.0
            & \textcolor{blue}{\underline{80.9}}, \underline{80.8}
            & \textcolor{blue}{80.2}, 79.8
            & \textcolor{blue}{78.1}, 78.2
            & \textcolor{blue}{79.2}, 79.3
            & \textcolor{blue}{80.7}, 80.3
            & \textcolor{blue}{\textbf{81.2}}, \textbf{80.9} \\
          & 
          & ARI 
            & \textcolor{blue}{76.9}, 76.4
            & \textcolor{blue}{75.2}, 74.5
            & \textcolor{blue}{75.0}, 74.5
            & \textcolor{blue}{79.0}, 78.5
            & \textcolor{blue}{78.0}, 77.4
            & \textcolor{blue}{73.4}, 73.4
            & \textcolor{blue}{75.7}, 75.6
            & \textcolor{blue}{\underline{79.1}}, \underline{78.6}
            & \textcolor{blue}{\textbf{79.6}}, \textbf{79.1} \\
        \cmidrule(lr){2-12}
        & \multirow{3}{*}{10k} 
          & ACC 
            & \textcolor{blue}{90.0}, 89.5
            & \textcolor{blue}{90.0}, 89.5
            & \textcolor{blue}{88.8}, 88.4
            & \textcolor{blue}{90.1}, 89.8
            & \textcolor{blue}{89.9}, 89.7
            & \textcolor{blue}{85.6}, 85.6
            & \textcolor{blue}{87.7}, 87.4
            & \textcolor{blue}{\textbf{90.5}}, \textbf{90.0}
            & \textcolor{blue}{\textbf{90.5}}, \underline{89.9} \\
          & 
          & NMI 
            & \textcolor{blue}{81.6}, 80.9
            & \textcolor{blue}{81.3}, 80.6
            & \textcolor{blue}{81.3}, 80.6
            & \textcolor{blue}{82.0}, \textbf{81.8}
            & \textcolor{blue}{81.9}, \underline{81.6}
            & \textcolor{blue}{78.2}, 78.3
            & \textcolor{blue}{80.7}, 80.4
            & \textcolor{blue}{\textbf{82.3}}, \underline{81.6}
            & \textcolor{blue}{\textbf{82.3}}, \underline{81.6} \\
          & 
          & ARI 
            & \textcolor{blue}{80.4}, 79.5
            & \textcolor{blue}{80.3}, 79.5
            & \textcolor{blue}{79.4}, 78.5
            & \textcolor{blue}{80.7}, \underline{80.1}
            & \textcolor{blue}{80.4}, 80.0
            & \textcolor{blue}{73.8}, 73.8
            & \textcolor{blue}{78.2}, 77.7
            & \textcolor{blue}{\underline{81.3}}, \textbf{80.4}
            & \textcolor{blue}{\textbf{81.4}}, \textbf{80.4} \\
        \midrule
        \multirow{9}{*}{\raisebox{-3ex}{FMNIST}}
        & \multirow{3}{*}{1k}  
          & ACC 
            & \textcolor{blue}{56.6}, 56.3
            & \textcolor{blue}{50.9}, 50.6
            & \textcolor{blue}{52.7}, 52.0
            & \textcolor{blue}{58.0}, 57.6
            & \textcolor{blue}{\underline{64.7}}, \underline{63.5}
            & \textcolor{blue}{56.7}, 56.6
            & \textcolor{blue}{39.6}, 39.4
            & \textcolor{blue}{62.8}, 62.1
            & \textcolor{blue}{\textbf{70.3}}, \textbf{69.8} \\
          & 
          & NMI 
            & \textcolor{blue}{56.4}, 56.1
            & \textcolor{blue}{49.5}, 49.1
            & \textcolor{blue}{53.4}, 52.9
            & \textcolor{blue}{61.1}, 60.3
            & \textcolor{blue}{\underline{62.0}}, \underline{61.1}
            & \textcolor{blue}{62.0}, 61.2
            & \textcolor{blue}{41.1}, 41.0
            & \textcolor{blue}{60.1}, 59.5
            & \textcolor{blue}{\textbf{62.3}}, \textbf{61.7} \\
          & 
          & ARI 
            & \textcolor{blue}{43.9}, 43.3
            & \textcolor{blue}{33.3}, 32.7
            & \textcolor{blue}{38.2}, 37.5
            & \textcolor{blue}{44.9}, 43.9
            & \textcolor{blue}{49.6}, 48.4
            & \textcolor{blue}{44.7}, 43.5
            & \textcolor{blue}{25.0}, 24.7
            & \textcolor{blue}{\underline{49.8}}, \underline{49.1}
            & \textcolor{blue}{\textbf{52.7}}, \textbf{51.9} \\
        \cmidrule(lr){2-12}
        & \multirow{3}{*}{5k}  
          & ACC 
            & \textcolor{blue}{76.2}, 75.2
            & \textcolor{blue}{76.0}, 75.3
            & \textcolor{blue}{71.7}, 71.2
            & \textcolor{blue}{64.6}, 63.8
            & \textcolor{blue}{78.5}, 77.3
            & \textcolor{blue}{57.3}, 57.3
            & \textcolor{blue}{59.0}, 58.6
            & \textcolor{blue}{\underline{80.1}}, \underline{79.0}
            & \textcolor{blue}{\textbf{81.0}}, \textbf{79.9} \\
          & 
          & NMI 
            & \textcolor{blue}{67.4}, 66.5
            & \textcolor{blue}{67.3}, 66.6
            & \textcolor{blue}{65.7}, 65.1
            & \textcolor{blue}{61.2}, 60.4
            & \textcolor{blue}{70.7}, \underline{69.8}
            & \textcolor{blue}{62.8}, 62.0
            & \textcolor{blue}{57.9}, 57.5
            & \textcolor{blue}{\underline{70.8}}, 69.7
            & \textcolor{blue}{\textbf{72.0}}, \textbf{70.9} \\
          & 
          & ARI 
            & \textcolor{blue}{61.4}, 60.0
            & \textcolor{blue}{60.8}, 59.8
            & \textcolor{blue}{57.1}, 56.3
            & \textcolor{blue}{48.7}, 47.5
            & \textcolor{blue}{64.5}, 63.1
            & \textcolor{blue}{45.6}, 44.5
            & \textcolor{blue}{45.6}, 45.0
            & \textcolor{blue}{\underline{65.2}}, \underline{63.7}
            & \textcolor{blue}{\textbf{66.8}}, \textbf{65.3} \\
        \cmidrule(lr){2-12}
        & \multirow{3}{*}{10k} 
          & ACC 
            & \textcolor{blue}{80.3}, 79.0
            & \textcolor{blue}{80.2}, 78.7
            & \textcolor{blue}{77.7}, 76.7
            & \textcolor{blue}{74.7}, 74.0
            & \textcolor{blue}{81.5}, 80.3
            & \textcolor{blue}{58.1}, 58.2
            & \textcolor{blue}{68.1}, 67.5
            & \textcolor{blue}{\underline{83.6}}, \underline{82.3}
            & \textcolor{blue}{\textbf{84.8}}, \textbf{83.6} \\
          & 
          & NMI 
            & \textcolor{blue}{71.3}, 70.1  
            & \textcolor{blue}{71.2}, 69.6  
            & \textcolor{blue}{71.3}, 70.3
            & \textcolor{blue}{68.9}, 68.2  
            & \textcolor{blue}{\underline{73.8}}, 72.3  
            & \textcolor{blue}{63.1}, 62.4  
            & \textcolor{blue}{65.0}, 64.3  
            & \textcolor{blue}{\underline{73.8}}, \underline{72.4}
            & \textcolor{blue}{\textbf{75.6}}, \textbf{74.2} \\  
          & 
          & ARI 
            & \textcolor{blue}{66.4}, 64.6  
            & \textcolor{blue}{66.2}, 63.9  
            & \textcolor{blue}{65.0}, 63.7
            & \textcolor{blue}{61.4}, 60.2  
            & \textcolor{blue}{68.6}, 66.8  
            & \textcolor{blue}{46.1}, 45.2  
            & \textcolor{blue}{55.0}, 53.9
            & \textcolor{blue}{\underline{69.8}}, \underline{67.9}
            & \textcolor{blue}{\textbf{72.0}}, \textbf{70.1} \\
        \midrule
        \multirow{9}{*}{\raisebox{-3ex}{ImageNet10}}
        & \multirow{3}{*}{1k}  
          & ACC 
            & \textcolor{blue}{83.4}, 83.6
            & \textcolor{blue}{84.0}, 83.9
            & \textcolor{blue}{83.9}, 84.1
            & \textcolor{blue}{92.2}, 92.7
            & \textcolor{blue}{94.3}, 94.4
            & \textcolor{blue}{89.0}, 88.9
            & \textcolor{blue}{61.2}, 60.7
            & \textcolor{blue}{\textbf{95.9}}, \underline{95.6}
            & \textcolor{blue}{\textbf{95.9}}, \textbf{95.9}
            \\
          & 
          & NMI 
            & \textcolor{blue}{83.1}, 83.7
            & \textcolor{blue}{81.7}, 82.9
            & \textcolor{blue}{81.0}, 82.2
            & \textcolor{blue}{88.3}, 88.8
            & \textcolor{blue}{89.1}, 89.4
            & \textcolor{blue}{84.8}, 84.4
            & \textcolor{blue}{55.7}, 55.4
            & \textcolor{blue}{\underline{90.6}}, \textbf{91.1}
            & \textcolor{blue}{\textbf{90.7}}, \textbf{91.1}
            \\
          & 
          & ARI 
            & \textcolor{blue}{77.0}, 76.9
            & \textcolor{blue}{75.9}, 76.1
            & \textcolor{blue}{75.2}, 74.9
            & \textcolor{blue}{85.8}, 86.3
            & \textcolor{blue}{88.9}, 88.8
            & \textcolor{blue}{80.7}, 80.1
            & \textcolor{blue}{39.5}, 38.3
            & \textcolor{blue}{\textbf{91.2}}, \underline{91.1}
            & \textcolor{blue}{\textbf{91.2}}, \textbf{91.2} \\
        \cmidrule(lr){2-12}
        & \multirow{3}{*}{5k}  
          & ACC 
            & \textcolor{blue}{\underline{96.8}}, 96.3
            & \textcolor{blue}{\underline{96.8}}, 96.4
            & \textcolor{blue}{96.3}, 96.2
            & \textcolor{blue}{\underline{96.8}}, \underline{96.5}
            & \textcolor{blue}{96.6}, 96.3
            & \textcolor{blue}{89.5}, 89.5
            & \textcolor{blue}{96.4}, 96.3
            & \textcolor{blue}{\underline{96.8}}, 96.4
            & \textcolor{blue}{\textbf{96.9}}, \textbf{96.6} \\
          & 
          & NMI 
            & \textcolor{blue}{\underline{92.5}}, 92.3
            & \textcolor{blue}{\textbf{92.6}}, \textbf{92.7}
            & \textcolor{blue}{91.6}, 92.1
            & \textcolor{blue}{92.4}, 92.2
            & \textcolor{blue}{91.8}, 91.5
            & \textcolor{blue}{86.1}, 86.0
            & \textcolor{blue}{91.6}, 91.6
            & \textcolor{blue}{92.2}, \underline{92.5}
            & \textcolor{blue}{92.4}, 92.2 \\
          & 
          & ARI 
            & \textcolor{blue}{\textbf{93.4}}, 92.1
            & \textcolor{blue}{\underline{93.2}}, 92.3
            & \textcolor{blue}{92.2}, 91.8
            & \textcolor{blue}{93.1}, \underline{92.5}
            & \textcolor{blue}{92.6}, 91.9
            & \textcolor{blue}{82.2}, 82.0
            & \textcolor{blue}{92.3}, 92.0
            & \textcolor{blue}{\underline{93.2}}, 92.3
            & \textcolor{blue}{\underline{93.2}}, \textbf{92.7} \\
        \cmidrule(lr){2-12}
        & \multirow{3}{*}{10k} 
          & ACC 
            & \textcolor{blue}{97.0}, 96.2
            & \textcolor{blue}{96.9}, 96.2
            & \textcolor{blue}{97.1}, 96.5
            & \textcolor{blue}{\underline{97.2}}, \underline{96.6}
            & \textcolor{blue}{97.0}, 96.5
            & \textcolor{blue}{89.6}, 89.6
            & \textcolor{blue}{96.8}, 96.6
            & \textcolor{blue}{97.0}, 96.2
            & \textcolor{blue}{\textbf{97.3}}, \textbf{96.7} \\
          & 
          & NMI 
            & \textcolor{blue}{\textbf{93.2}}, 92.2
            & \textcolor{blue}{92.8}, 92.4
            & \textcolor{blue}{92.9}, 92.3
            & \textcolor{blue}{93.0}, \underline{92.4}
            & \textcolor{blue}{92.7}, 92.0
            & \textcolor{blue}{86.3}, 86.4
            & \textcolor{blue}{92.2}, 92.2
            & \textcolor{blue}{\textbf{93.2}}, \textbf{92.6}
            & \textcolor{blue}{\textbf{93.2}}, \underline{92.5} \\
          & 
          & ARI 
            & \textcolor{blue}{\textbf{94.1}}, 91.9
            & \textcolor{blue}{93.7}, 91.9
            & \textcolor{blue}{93.7}, 92.4
            & \textcolor{blue}{93.9}, \underline{92.7}
            & \textcolor{blue}{93.6}, 92.5
            & \textcolor{blue}{82.4}, 82.4
            & \textcolor{blue}{93.0}, \underline{92.7}
            & \textcolor{blue}{94.0}, 92.4
            & \textcolor{blue}{\textbf{94.1}}, \textbf{93.0} \\
        \midrule
        \multirow{9}{*}{\raisebox{-3ex}{MNIST}}   
        & \multirow{3}{*}{1k}  
          & ACC
            & \textcolor{blue}{54.4}, 54.9
            & \textcolor{blue}{57.4}, 57.3
            & \textcolor{blue}{65.1}, 65.8
            & \textcolor{blue}{\underline{88.6}}, \underline{88.0}
            & \textcolor{blue}{84.4}, 84.6
            & \textcolor{blue}{84.4}, 84.7
            & \textcolor{blue}{43.2}, 43.4
            & \textcolor{blue}{69.8}, 71.0
            & \textcolor{blue}{\textbf{91.6}}, \textbf{91.7} \\
          & 
          & NMI 
            & \textcolor{blue}{48.1}, 49.5
            & \textcolor{blue}{50.8}, 51.6
            & \textcolor{blue}{62.7}, 63.0
            & \textcolor{blue}{\textbf{86.8}}, \textbf{86.0}
            & \textcolor{blue}{80.4}, 80.8
            & \textcolor{blue}{79.6}, 80.5
            & \textcolor{blue}{35.5}, 36.6
            & \textcolor{blue}{59.8}, 61.5
            & \textcolor{blue}{\underline{82.5}}, \underline{82.8} \\
          & 
          & ARI 
            & \textcolor{blue}{39.0}, 40.0
            & \textcolor{blue}{40.8}, 41.2
            & \textcolor{blue}{51.9}, 52.1
            & \textcolor{blue}{\textbf{83.5}}, \underline{82.3}
            & \textcolor{blue}{75.4}, 75.7
            & \textcolor{blue}{75.9}, 76.5
            & \textcolor{blue}{22.9}, 23.4
            & \textcolor{blue}{54.2}, 55.8
            & \textcolor{blue}{\underline{82.6}}, \textbf{82.8} \\
        \cmidrule(lr){2-12}
        & \multirow{3}{*}{5k}  
          & ACC 
            & \textcolor{blue}{82.2}, 82.6
            & \textcolor{blue}{76.5}, 77.0
            & \textcolor{blue}{90.2}, 89.5
            & \textcolor{blue}{\underline{95.8}}, \underline{95.4}
            & \textcolor{blue}{94.0}, 94.1
            & \textcolor{blue}{86.2}, 86.1
            & \textcolor{blue}{58.5}, 58.9
            & \textcolor{blue}{93.2}, 93.3
            & \textcolor{blue}{\textbf{96.1}}, \textbf{95.9} \\
          & 
          & NMI 
            & \textcolor{blue}{75.1}, 76.2
            & \textcolor{blue}{68.0}, 69.1
            & \textcolor{blue}{85.5}, 84.3
            & \textcolor{blue}{\textbf{91.6}}, \textbf{90.7}
            & \textcolor{blue}{88.7}, 88.7
            & \textcolor{blue}{82.6}, 83.0
            & \textcolor{blue}{57.5}, 58.8
            & \textcolor{blue}{84.5}, 84.9
            & \textcolor{blue}{\underline{90.0}}, \underline{89.9} \\
          & 
          & ARI 
            & \textcolor{blue}{72.4}, 73.2
            & \textcolor{blue}{64.2}, 65.1
            & \textcolor{blue}{83.9}, 82.4
            & \textcolor{blue}{\textbf{92.0}}, \underline{91.0}
            & \textcolor{blue}{88.6}, 88.6
            & \textcolor{blue}{79.4}, 79.4
            & \textcolor{blue}{46.9}, 47.5
            & \textcolor{blue}{85.7}, 86.0
            & \textcolor{blue}{\underline{91.5}}, \textbf{91.2} \\
        \cmidrule(lr){2-12}
        & \multirow{3}{*}{10k} 
          & ACC 
            & \textcolor{blue}{92.6}, 92.6
            & \textcolor{blue}{91.4}, 91.3
            & \textcolor{blue}{96.5}, 95.6
            & \textcolor{blue}{\textbf{97.5}}, \textbf{97.0}
            & \textcolor{blue}{95.6}, 95.3
            & \textcolor{blue}{86.3}, 86.1
            & \textcolor{blue}{81.5}, 81.8
            & \textcolor{blue}{95.6}, 95.6
            & \textcolor{blue}{\underline{97.1}}, \textbf{97.0} \\
          & 
          & NMI 
            & \textcolor{blue}{84.1}, 84.3
            & \textcolor{blue}{82.9}, 83.2
            & \textcolor{blue}{91.3}, 89.7
            & \textcolor{blue}{\textbf{93.5}}, \textbf{92.6}
            & \textcolor{blue}{91.2}, 90.8
            & \textcolor{blue}{83.0}, 83.3
            & \textcolor{blue}{77.7}, 78.7
            & \textcolor{blue}{89.0}, 89.1
            & \textcolor{blue}{\underline{92.1}}, \underline{92.1} \\
          & 
          & ARI 
            & \textcolor{blue}{85.3}, 85.3
            & \textcolor{blue}{83.6}, 83.5
            & \textcolor{blue}{92.5}, 90.7
            & \textcolor{blue}{\textbf{94.7}}, \textbf{93.7}
            & \textcolor{blue}{91.7}, 91.2
            & \textcolor{blue}{79.6}, 79.5
            & \textcolor{blue}{74.1}, 74.9
            & \textcolor{blue}{90.7}, 90.6
            & \textcolor{blue}{\underline{93.7}}, \textbf{93.7} \\
        \midrule
        \multirow{9}{*}{\raisebox{-3ex}{REUTERS}}
        & \multirow{3}{*}{1k}  
          & ACC 
            & \textcolor{blue}{71.1}, 70.8
            & \textcolor{blue}{66.7}, 66.7
            & \textcolor{blue}{70.7}, 71.9
            & \textcolor{blue}{77.0}, 76.9
            & \textcolor{blue}{85.1}, \underline{87.1}
            & \textcolor{blue}{72.6}, 71.2
            & \textcolor{blue}{49.5}, 48.8
            & \textcolor{blue}{\underline{87.1}}, 86.0
            & \textcolor{blue}{\textbf{91.2}}, \textbf{91.4} \\
          & 
          & NMI 
            & \textcolor{blue}{42.3}, 43.0
            & \textcolor{blue}{38.8}, 39.8
            & \textcolor{blue}{48.2}, 51.1
            & \textcolor{blue}{59.3}, 60.5
            & \textcolor{blue}{\underline{68.8}}, \underline{68.2}
            & \textcolor{blue}{52.4}, 51.5
            & \textcolor{blue}{15.2}, 14.7
            & \textcolor{blue}{64.1}, 62.5
            & \textcolor{blue}{\textbf{72.9}}, \textbf{74.1} \\
          & 
          & ARI 
            & \textcolor{blue}{49.1}, 49.0
            & \textcolor{blue}{45.7}, 45.4
            & \textcolor{blue}{51.3}, 54.6
            & \textcolor{blue}{66.1}, 66.9
            & \textcolor{blue}{\underline{77.7}}, \underline{77.0}
            & \textcolor{blue}{57.3}, 55.8
            & \textcolor{blue}{14.6}, 13.7
            & \textcolor{blue}{74.1}, 71.9
            & \textcolor{blue}{\textbf{81.0}}, \textbf{81.5} \\
        \cmidrule(lr){2-12}
        & \multirow{3}{*}{5k}  
          & ACC 
            & \textcolor{blue}{\underline{96.1}}, \textbf{94.8}
            & \textcolor{blue}{95.5}, 94.3
            & \textcolor{blue}{82.0}, 82.3
            & \textcolor{blue}{93.0}, 92.7
            & \textcolor{blue}{95.7}, 94.4
            & \textcolor{blue}{72.1}, 71.6
            & \textcolor{blue}{85.6}, 84.9
            & \textcolor{blue}{95.6}, 93.9
            & \textcolor{blue}{\textbf{96.2}}, \textbf{94.8} \\
          & 
          & NMI 
            & \textcolor{blue}{\underline{84.6}}, \textbf{81.5}
            & \textcolor{blue}{83.0}, 79.6
            & \textcolor{blue}{60.5}, 61.4
            & \textcolor{blue}{79.4}, 78.3
            & \textcolor{blue}{83.6}, 80.2
            & \textcolor{blue}{54.6}, 54.6
            & \textcolor{blue}{63.7}, 62.9
            & \textcolor{blue}{83.1}, 78.2
            & \textcolor{blue}{\textbf{84.9}}, \underline{80.7} \\
          & 
          & ARI 
            & \textcolor{blue}{\underline{90.9}}, \textbf{88.1}
            & \textcolor{blue}{89.7}, 86.6
            & \textcolor{blue}{68.0}, 68.2
            & \textcolor{blue}{86.6}, 85.3
            & \textcolor{blue}{90.3}, 87.1
            & \textcolor{blue}{57.1}, 57.7
            & \textcolor{blue}{73.7}, 71.8
            & \textcolor{blue}{89.7}, 85.3
            & \textcolor{blue}{\textbf{91.0}}, \underline{87.2} \\
        \cmidrule(lr){2-12}
        & \multirow{3}{*}{10k} 
          & ACC 
            & \textcolor{blue}{97.4}, 94.9
            & \textcolor{blue}{97.1}, 94.4
            & \textcolor{blue}{88.1}, 86.3
            & \textcolor{blue}{97.6}, \textbf{95.9}
            & \textcolor{blue}{97.5}, 95.6
            & \textcolor{blue}{71.6}, 71.2
            & \textcolor{blue}{92.8}, 91.0
            & \textcolor{blue}{\textbf{97.8}}, 95.2
            & \textcolor{blue}{\textbf{97.8}}, \underline{95.8} \\
          & 
          & NMI 
            & \textcolor{blue}{89.1}, 80.9
            & \textcolor{blue}{88.2}, 79.6
            & \textcolor{blue}{70.8}, 68.7
            & \textcolor{blue}{89.8}, \textbf{84.5}
            & \textcolor{blue}{89.3}, 83.5
            & \textcolor{blue}{54.0}, 53.9
            & \textcolor{blue}{78.6}, 75.4
            & \textcolor{blue}{\textbf{90.3}}, 81.9
            & \textcolor{blue}{\underline{90.2}}, \underline{83.9} \\
          & 
          & ARI 
            & \textcolor{blue}{94.2}, 87.5
            & \textcolor{blue}{93.6}, 86.3
            & \textcolor{blue}{80.3}, 76.0
            & \textcolor{blue}{94.5}, \textbf{90.1}
            & \textcolor{blue}{94.3}, 89.3
            & \textcolor{blue}{56.1}, 56.8
            & \textcolor{blue}{86.6}, 82.7
            & \textcolor{blue}{\textbf{95.0}}, 87.9
            & \textcolor{blue}{\textbf{95.0}}, \underline{89.8} \\
        \midrule
        \multirow{9}{*}{\raisebox{-3ex}{STL10}}
        & \multirow{3}{*}{1k}  
          & ACC 
            & \textcolor{blue}{69.9}, 70.0
            & \textcolor{blue}{74.4}, 74.5
            & \textcolor{blue}{69.8}, 71.1
            & \textcolor{blue}{\underline{88.0}}, \underline{87.8}
            & \textcolor{blue}{77.6}, 77.2
            & \textcolor{blue}{85.5}, 85.6
            & \textcolor{blue}{50.6}, 51.0
            & \textcolor{blue}{86.4}, 87.2
            & \textcolor{blue}{\textbf{88.1}}, \textbf{88.1} \\
          & 
          & NMI 
            & \textcolor{blue}{64.9}, 65.6
            & \textcolor{blue}{69.4}, 69.5
            & \textcolor{blue}{65.4}, 67.3
            & \textcolor{blue}{\textbf{79.5}}, \textbf{79.2}
            & \textcolor{blue}{70.0}, 70.1
            & \textcolor{blue}{77.2}, 77.1
            & \textcolor{blue}{49.9}, 50.5
            & \textcolor{blue}{76.6}, 78.4
            & \textcolor{blue}{\underline{78.7}}, \underline{79.1} \\
          & 
          & ARI 
            & \textcolor{blue}{54.5}, 54.7
            & \textcolor{blue}{61.9}, 61.8
            & \textcolor{blue}{57.0}, 58.9
            & \textcolor{blue}{\underline{76.5}}, \underline{76.1}
            & \textcolor{blue}{63.7}, 63.2
            & \textcolor{blue}{71.8}, 72.1
            & \textcolor{blue}{33.6}, 33.9
            & \textcolor{blue}{74.2}, 75.7
            & \textcolor{blue}{\textbf{76.7}}, \textbf{76.8} \\
        \cmidrule(lr){2-12}
        & \multirow{3}{*}{5k}  
          & ACC 
            & \textcolor{blue}{88.1}, 87.1
            & \textcolor{blue}{88.9}, 87.9
            & \textcolor{blue}{87.2}, 86.9
            & \textcolor{blue}{90.3}, 89.4
            & \textcolor{blue}{88.0}, 87.1
            & \textcolor{blue}{87.3}, 87.6
            & \textcolor{blue}{84.0}, 83.7
            & \textcolor{blue}{\underline{91.2}}, \underline{90.1}
            & \textcolor{blue}{\textbf{91.4}}, \textbf{90.3} \\
          & 
          & NMI 
            & \textcolor{blue}{80.6}, 79.5
            & \textcolor{blue}{80.2}, 79.4
            & \textcolor{blue}{78.6}, 79.1
            & \textcolor{blue}{81.9}, 80.9
            & \textcolor{blue}{78.7}, 78.1
            & \textcolor{blue}{79.1}, 79.3
            & \textcolor{blue}{77.8}, 77.4
            & \textcolor{blue}{\underline{83.0}}, \underline{81.7}
            & \textcolor{blue}{\textbf{83.1}}, \textbf{81.9} \\
          & 
          & ARI 
            & \textcolor{blue}{78.3}, 76.5
            & \textcolor{blue}{78.0}, 76.5
            & \textcolor{blue}{76.0}, 76.0
            & \textcolor{blue}{80.4}, 78.9
            & \textcolor{blue}{77.0}, 75.7
            & \textcolor{blue}{75.0}, 75.6
            & \textcolor{blue}{73.7}, 72.8
            & \textcolor{blue}{\underline{82.2}}, \underline{80.3}
            & \textcolor{blue}{\textbf{82.5}}, \textbf{80.5} \\
        \cmidrule(lr){2-12}
        & \multirow{3}{*}{10k} 
          & ACC 
            & \textcolor{blue}{92.6}, 89.9
            & \textcolor{blue}{92.4}, 90.1
            & \textcolor{blue}{90.7}, 89.7
            & \textcolor{blue}{91.3}, 89.9
            & \textcolor{blue}{91.4}, 89.5
            & \textcolor{blue}{87.8}, 88.0
            & \textcolor{blue}{90.9}, 89.6
            & \textcolor{blue}{\textbf{93.0}}, \textbf{91.0}
            & \textcolor{blue}{\textbf{93.0}}, \textbf{91.0} \\
          & 
          & NMI 
            & \textcolor{blue}{84.9}, 81.8
            & \textcolor{blue}{84.2}, 81.8
            & \textcolor{blue}{82.3}, 81.7
            & \textcolor{blue}{83.2}, 81.5
            & \textcolor{blue}{82.7}, 80.7
            & \textcolor{blue}{79.4}, 79.6
            & \textcolor{blue}{82.5}, 81.0
            & \textcolor{blue}{\textbf{85.4}}, \textbf{83.2}
            & \textcolor{blue}{\textbf{85.4}}, \textbf{83.2} \\
          & 
          & ARI 
            & \textcolor{blue}{84.7}, 80.0
            & \textcolor{blue}{83.9}, 80.0
            & \textcolor{blue}{81.3}, 79.7
            & \textcolor{blue}{82.4}, 79.8
            & \textcolor{blue}{82.3}, 79.1
            & \textcolor{blue}{75.9}, 76.3
            & \textcolor{blue}{81.5}, 79.2
            & \textcolor{blue}{\underline{85.4}}, \textbf{81.9}
            & \textcolor{blue}{\textbf{85.5}}, \underline{81.7} \\
        \midrule
        \multirow{9}{*}{\raisebox{-3ex}{RCV1-10}}   
        & \multirow{3}{*}{1k}  
          & ACC 
            & \textcolor{blue}{38.1}, 38.1
            & \textcolor{blue}{50.0}, 49.9
            & \textcolor{blue}{41.7}, 42.3
            & \textcolor{blue}{39.8}, 39.2
            & \textcolor{blue}{34.7}, 34.6
            & \textcolor{blue}{39.5}, 39.4
            & \textcolor{blue}{33.4}, 33.4
            & \textcolor{blue}{\underline{52.1}}, \underline{52.2}
            & \textcolor{blue}{\textbf{65.8}}, \textbf{65.8} \\
          & 
          & NMI 
            & \textcolor{blue}{12.3}, 12.3
            & \textcolor{blue}{40.1}, 40.3
            & \textcolor{blue}{32.6}, 33.6
            & \textcolor{blue}{45.1}, 46.2
            & \textcolor{blue}{17.8}, 17.9
            & \textcolor{blue}{\underline{49.6}}, \underline{49.9}
            & \textcolor{blue}{10.4}, 10.4
            & \textcolor{blue}{47.6}, 47.8
            & \textcolor{blue}{\textbf{60.8}}, \textbf{61.1} \\
          & 
          & ARI 
            & \textcolor{blue}{12.9}, 12.8
            & \textcolor{blue}{39.5}, 39.4
            & \textcolor{blue}{27.2}, 28.4
            & \textcolor{blue}{31.0}, 31.1
            & \textcolor{blue}{12.3}, 12.2
            & \textcolor{blue}{32.5}, 32.4
            & \textcolor{blue}{7.8}, 7.8
            & \textcolor{blue}{\underline{42.5}}, \underline{42.5}
            & \textcolor{blue}{\textbf{55.8}}, \textbf{55.9} \\
        \cmidrule(lr){2-12}
        & \multirow{3}{*}{5k}  
          & ACC 
            & \textcolor{blue}{78.1}, 78.1
            & \textcolor{blue}{80.7}, 80.6
            & \textcolor{blue}{57.6}, 55.1
            & \textcolor{blue}{61.4}, 61.7
            & \textcolor{blue}{54.1}, 53.5
            & \textcolor{blue}{39.5}, 39.4
            & \textcolor{blue}{49.4}, 49.4
            & \textcolor{blue}{\underline{86.7}}, \underline{86.5}
            & \textcolor{blue}{\textbf{89.8}}, \textbf{89.6} \\
          & 
          & NMI 
            & \textcolor{blue}{61.1}, 60.9
            & \textcolor{blue}{61.9}, 61.8
            & \textcolor{blue}{44.8}, 47.5
            & \textcolor{blue}{53.7}, 54.4
            & \textcolor{blue}{38.2}, 38.0
            & \textcolor{blue}{50.9}, 51.2
            & \textcolor{blue}{31.6}, 31.5
            & \textcolor{blue}{\underline{69.2}}, \underline{68.9}
            & \textcolor{blue}{\textbf{74.2}}, \textbf{73.9} \\
          & 
          & ARI 
            & \textcolor{blue}{70.7}, 70.5
            & \textcolor{blue}{75.3}, 75.1
            & \textcolor{blue}{46.4}, 46.2
            & \textcolor{blue}{53.4}, 54.0
            & \textcolor{blue}{40.2}, 39.3
            & \textcolor{blue}{33.0}, 33.0
            & \textcolor{blue}{35.9}, 35.8
            & \textcolor{blue}{\underline{81.3}}, \underline{80.9}
            & \textcolor{blue}{\textbf{83.8}}, \textbf{83.4} \\
        \cmidrule(lr){2-12}
        & \multirow{3}{*}{10k} 
          & ACC 
            & \textcolor{blue}{84.3}, 84.1
            & \textcolor{blue}{87.8}, 87.5
            & \textcolor{blue}{58.2}, 54.5
            & \textcolor{blue}{80.0}, 79.4
            & \textcolor{blue}{81.0}, 80.0
            & \textcolor{blue}{38.8}, 38.8
            & \textcolor{blue}{77.3}, 77.3
            & \textcolor{blue}{\underline{89.9}}, \underline{89.8}
            & \textcolor{blue}{\textbf{91.8}}, \textbf{91.5} \\
          & 
          & NMI 
            & \textcolor{blue}{70.0}, 69.6
            & \textcolor{blue}{70.8}, 70.3
            & \textcolor{blue}{47.3}, 51.0
            & \textcolor{blue}{66.9}, 67.7
            & \textcolor{blue}{63.4}, 62.7
            & \textcolor{blue}{50.5}, 50.8
            & \textcolor{blue}{58.4}, 58.2
            & \textcolor{blue}{\underline{74.9}}, \underline{74.6}
            & \textcolor{blue}{\textbf{78.0}}, \textbf{77.4} \\
          & 
          & ARI 
            & \textcolor{blue}{81.6}, 81.2
            & \textcolor{blue}{83.0}, 82.4
            & \textcolor{blue}{47.9}, 47.1
            & \textcolor{blue}{74.5}, 73.9
            & \textcolor{blue}{75.4}, 74.4
            & \textcolor{blue}{32.8}, 32.7
            & \textcolor{blue}{68.9}, 68.6
            & \textcolor{blue}{\underline{85.5}}, \underline{85.2}
            & \textcolor{blue}{\textbf{87.0}}, \textbf{86.4} \\
        \bottomrule
    \end{tabular}
    \end{scriptsize}
    \end{center}
    \vskip -0.2in
\end{table}

Using the experimental setup outlined in Sect.~\ref{subsect:ex-setting}, we now present and analyze our results to address the four motivating questions that guided our study.

\subsubsection{Comparative performance}
\label{sec:comparative}

\paragraph{Overall comparison.}
\cref{table:SpherePair_VS_baselines} reports SpherePair’s results against six baselines on all eight datasets under three constraint levels (1k/5k/10k).
Out of 72 total comparisons (8 datasets $\times$ 3 constraint levels $\times$ 3 metrics), SpherePair ranks first in over 60 cases and second in nearly all others.
It is notably dominant on CIFAR-100-20, FMNIST, REUTERS, and RCV1-10, achieving the top result in every metric at every constraint level (9/9 each), surpassing the second-best method by up to $4$-$16\%$ in absolute ACC.
Even when second-best, the performance gap is within $1$-$2\%$.
Additional comparisons with AutoEmbedder using hierarchical clustering (replacing K-means) can be found in \cref{sec:hierarchical}.
These findings showcase SpherePair’s state-of-the-art DCC performance across both image and text datasets, along with the robustness of its geometric formulation under varying supervision levels and data domains.

\paragraph{Comparison without pretraining.}
We further compared two strong models  without pretraining (random initialization), CIDEC\(^{\dagger}\) and SpherePair\(^{\dagger}\).
As shown in \cref{table:SpherePair_VS_baselines}, SpherePair\(^{\dagger}\) exceeds CIDEC\(^{\dagger}\) by $10$-$20\%$ on nearly all datasets except MNIST at 1k constraints; at 5k/10k it remains superior in most cases. 
Except for the highly class-imbalanced RCV1-10, pretrained SpherePair consistently yields a modest improvement (around $1$\%) over its unpretrained version. 
In contrast, CIDEC benefits substantially from pretraining only at 1k constraints, but shows inconsistent gains at higher constraint levels and even a drop on FMNIST.
Hence, end-to-end DCC rely heavily on sufficiently good initial clustering to bootstrap reliable iterations, but our anchor-free SpherePair loss demonstrates robustness to initialization, particularly when pretraining is unavailable or leads to degraded performance.

\subsubsection{Imbalanced constraints}
\label{sec:IMB}

\begin{figure}[t]
    \begin{center}
    \centerline{\includegraphics[width=\textwidth]{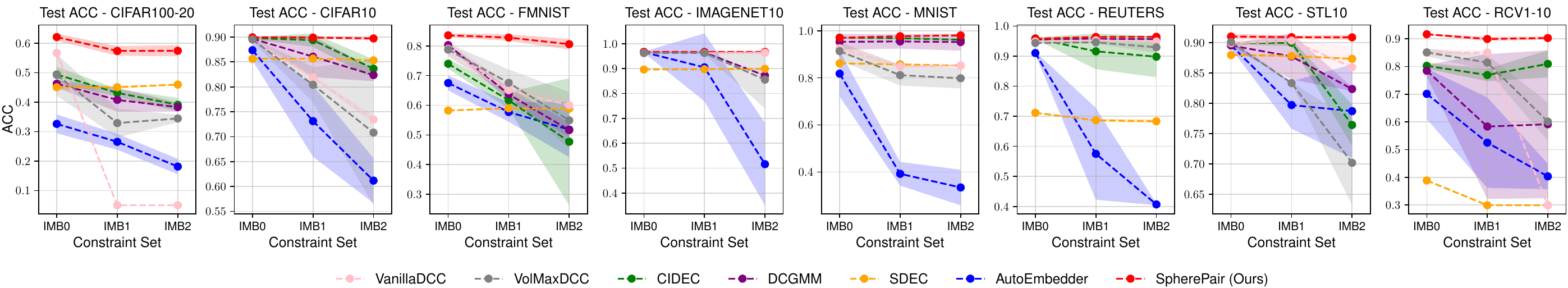}}
    \caption{Test ACC performance (mean$\pm$std over 5 runs) of all models across datasets under the balanced vs. imbalanced constraints setting where ($\mid${\tt IMB0}$\mid$, $\mid${\tt IMB1}$\mid$, $\mid${\tt IMB2}$\mid$) = (10k, 50k, 100k).}
    \label{fig:imb_10K_samll}
    \end{center}
    \vskip -0.3in
\end{figure}

Fig.~\ref{fig:imb_10K_samll} presents the test ACC under the imbalanced constraint setting outlined in Sect.~\ref{subsect:ex-setting}. 
As the imbalance increases, all DCC baselines experience significant performance degradation across most datasets.
In contrast, our SpherePair demonstrates remarkable stability and consistently outperforms 
\begin{wrapfigure}{r}{0.72\textwidth}
    \centering
    \vskip -0in
    \includegraphics[width=0.72\textwidth]{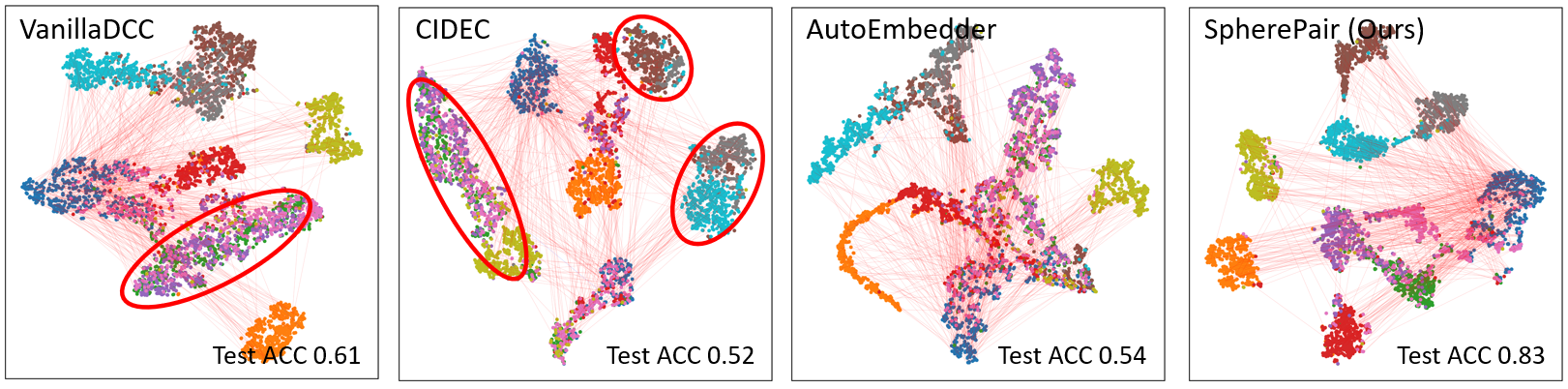}
    \vskip -0.0in
    \caption{t-SNE visualizations of learned FMNIST embeddings under the {\tt IMB2} setting in Fig.~\ref{fig:imb_10K_samll}. Marker colors denote ground-truth categories, and dashed lines represent pairwise constraints. The \textcolor{red}{red circles} highlight the misclustered instances.}
    \label{fig:tSNE_fmnist}
    \vskip -0.05in
\end{wrapfigure}
all baselines across the tasks.
\Cref{fig:tSNE_fmnist} visualizes the learned FMNIST embeddings, illustrating how SpherePair preserves coherent structures while forming more discriminative clusters compared to representative baselines. 
Notably, while the imbalanced sets contain more constraints than the balanced set, the performance of baselines worsens with greater imbalance, posing an open question for further investigation. 
Additional results and visualizations are in \cref{sec:IMB_appendix}, further showcasing the robustness of our approach in real-world scenarios with prevalent imbalanced constraints.

\subsubsection{Unknown cluster number}
\label{sec:unkownK}

\begin{figure}[htbp]
    \vskip -0.0in
    \begin{center}
    \centerline{\includegraphics[width=0.85\textwidth]{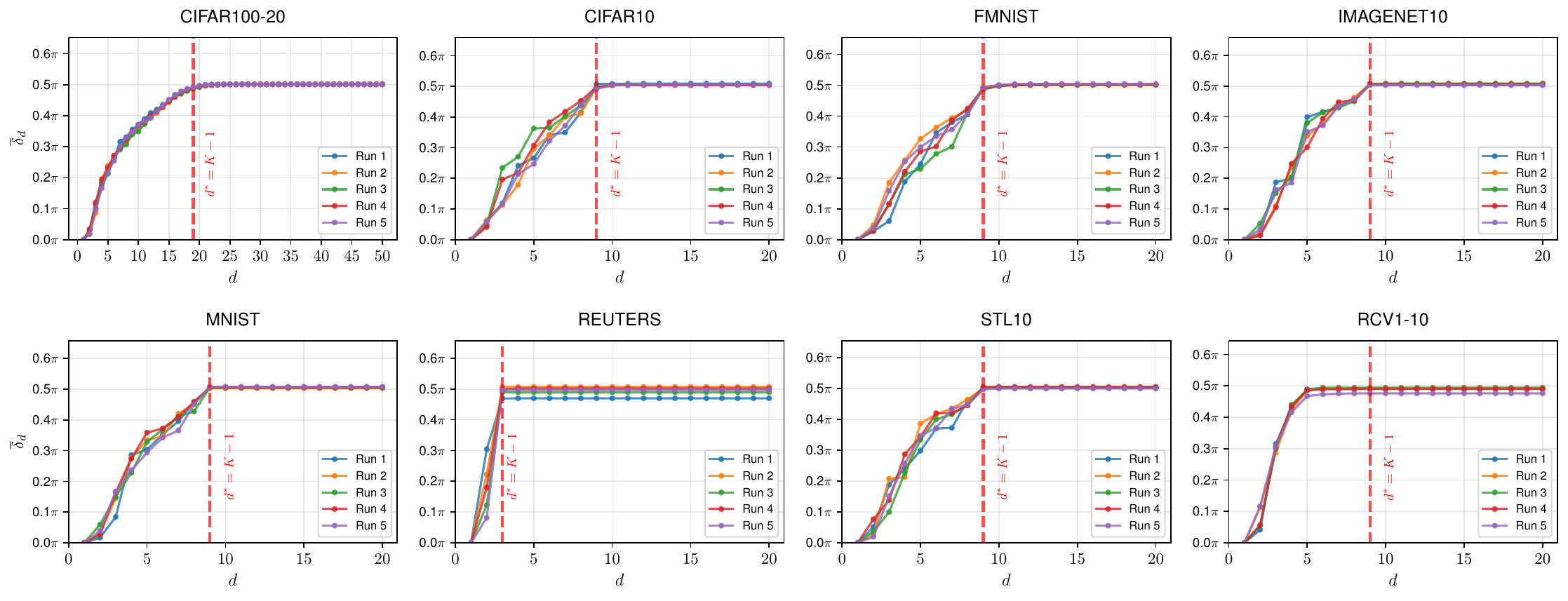}}
    \vskip -0.075in
    \caption{Tail-averaged minimal inter-cluster angle $\overline{\delta}_d$ vs. PCA subspace dimension $d$, obtained from SpherePair embeddings learned with 10k constraints across 5 runs. 
    The \textcolor{red}{red lines} indicate the ground-truth intrinsic dimensions $d^\ast = K{-}1$.}
    \label{fig:K_merge_main_10k}
    \end{center}
    \vskip -0.2in
\end{figure}

We evaluate our cluster-number inference on SpherePair embeddings, simulating the unknown-$K$ case with large embedding dimensions ($D=50$ for CIFAR-100-20 and $D=20$ for others). 
As shown in Fig.~\ref{fig:K_merge_main_10k}, the tail-averaged minimal inter-cluster angle $\overline{\delta}_d$ rises from $0$ as the PCA subspace dimension $d$ increases and, for most datasets, reaches a plateau at $\overline{\delta}_{K-1}$, which clearly reveals the true cluster number $K$. 
On CIFAR-100-20, the plateau is less distinct, yet the estimated $K$ falls within $18$-$22$ around the true $K=20$. 
The only notable deviation is observed on RCV1-10, where strong class imbalance results in only six dominant clusters being identified, underscoring the inherent difficulty of $K$-inference in such settings.
Nevertheless, our approach demonstrates overall robustness and efficiency, as further validated in \cref{sec:unknown_K} with additional results under varied constraint levels, comparisons to alternative $K$-inference methods, and evaluations against other DCC methods.

\begin{figure}[t]
    \vskip -0in
    \begin{center}
    \centerline{\includegraphics[width=0.84\textwidth]{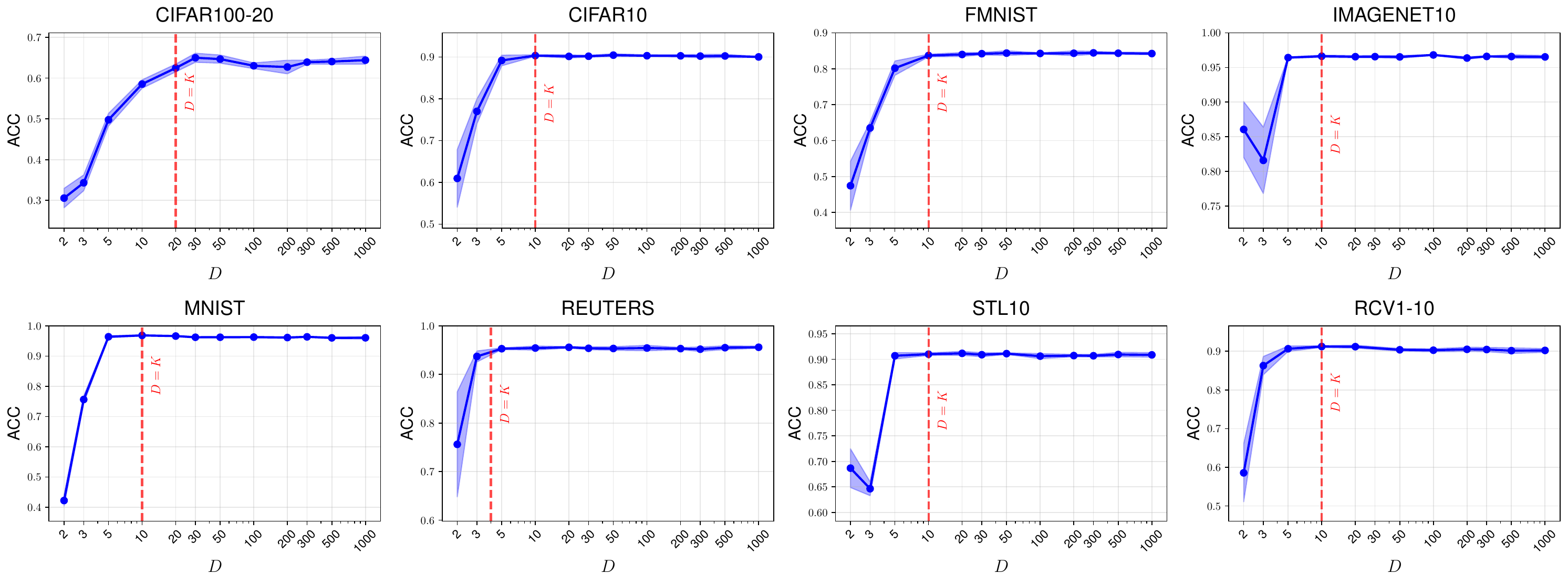}}
    \vskip -0.075in
    \caption{SpherePair test ACC (mean$\pm$std, 5 runs) vs. embedding dimension $D$ across datasets (10k constraints).
    The \textcolor{red}{red lines} indicate the theoretical boundary between insufficient and sufficient $D$.}
    \label{fig:D_10k_acc}
    \end{center}
    \vskip -0.3in
\end{figure}

\subsubsection{Empirical validation and hyperparameter sensitivity analysis}
\label{sec:hyperparameter}

\paragraph{Embedding dimension $D$.}
We study the impact of embedding dimension $D$, the only hyperparameter to be specified to obtain a conflict-free angular embedding, to empirically validate our theoretical insights.
Fig.~\ref{fig:D_10k_acc} presents performance with respect to $D$ under 10k balanced constraints (see \cref{sec:omega_and_D} for results under more constraint levels and clustering metrics), showing that:
(i) SpherePair is robust to choices of $D \geq K$, even up to $D=1000$, which corresponds to $50$--$250\times$ the cluster number $K$ across different datasets. This provides clear and easily satisfied practical guidance.
(ii) Even selecting $D$ slightly below the theoretical threshold minimally impacts performance, offering flexibility when $K$ is unavailable. Additional results in \cref{sec:omega_and_D} further support this $D$-flexibility through comparisons with baselines on CIFAR-100-20.
(iii) Ablation-like comparisons between theoretically sufficient and insufficient $D$ illustrate the effectiveness of our conflict-free constraint embedding in angular space, and empirically validate our theoretical insights in \cref{sec:theoretical}.

\paragraph{Other hyperparameters.}
We further conduct sensitivity analysis on the regularization strength $\lambda$ over $[0,1]$ and the tail ratio $\rho$ over $[0.01,0.2]$ (see \cref{sec:lambda_robustness,sec:rho_sensitivity}).
Key observations are: 
(i) SpherePair embedding is broadly robust to $\lambda$, though an appropriate $\lambda$ can be beneficial, particularly under limited constraints or random initialization. We recommend a default of $\lambda=0.02$ for consistent performance in most cases;
(ii) in cluster-number inference, smaller $\rho$ sharpens rises before $\overline{\delta}_{K-1}$, while larger $\rho$ stabilizes the subsequent plateau; $\rho \in [0.03, 0.1]$ generally offers a favorable trade-off for clearer $K$ estimation.\footnote{Robustness can be enhanced by evaluating consistency across multiple $\rho$ values within this range.}

In summary, the experimental findings provide robust validation of our contributions. Notably, our method also demonstrates significant potential for real-world applications in terms of its learning efficiency (see \cref{sec:efficiency}) and insensitivity to model structures (see \cref{sec:model_selection}).

\section{Conclusion}
\label{sect:conclude}

In this paper, we propose SpherePair, a novel representation learning approach for constrained clustering. 
It learns effective representations from pairwise constraints in an angular space, supported by theoretical guarantees.  
Extensive experiments on real-world and benchmark datasets demonstrate that SpherePair, when integrated with a simple clustering algorithm such as K-means,  consistently outperforms various state-of-the-art DCC baselines. 
Furthermore, SpherePair is anchor‐free, requires minimal hyperparameter tuning, offers robustness with theoretical guarantees, and readily handles an unknown number of clusters while rapidly inferring their quantity, making it highly applicable to real‐world constrained clustering tasks.

Our approach has two limitations:
(i) It currently supports only single-view unstructured data;
(ii) It does not address incomplete or noisy constraint annotations.
To address these challenges, we are extending SpherePair to handle semi-structured, structured, and multi-view data, as well as introducing mechanisms to manage noisy or incomplete annotations. 
We also aim to combine end-to-end and deep constraint embedding frameworks to capture higher-order correlations and improve scalability, particularly for applications requiring a deeper understanding of local and global relationships. 
We anticipate that addressing these challenges will result in more robust SpherePair models, improving the applicability and performance across diverse real-world data.

\section*{Acknowledgment}
We are grateful to the anonymous reviewers for their comments, which improved the presentation of this paper. S.J. Zhang’s work was supported by the UoM-CSC scholarship.


\bibliographystyle{unsrt}
\bibliography{reference}

\newpage
\appendix

\begin{center}
\LARGE \textbf{APPENDICES}
\end{center}
\vspace*{1cm}
\section{Deep constraint clustering formulations}
\label{sec:formulation}

The goal of constrained clustering (CC) is to partition a dataset $\mathcal{X} = \{\boldsymbol{x}_j\}_{j=1}^{|\mathcal{X}|}$ into $K$ clusters $\mathcal{S} = \{\mathcal{S}_k\}_{k=1}^K$ while satisfying a set of pairwise constraints $\mathcal{C} = \{(a_i, b_i, y_i)\}_{i=1}^{|\mathcal{C}|}$. Each constraint $(a_i, b_i, y_i)$ requires that instances $\boldsymbol{x}_{a_i}$ and $\boldsymbol{x}_{b_i}$ be assigned to the same cluster if $y_i = 1$, or to different clusters if $y_i = 0$.  
To solve this task, deep constraint clustering (DCC) methods employ a deep encoder $f_{\boldsymbol{\phi}}: \mathcal{X} \to \mathcal{Z} \subset \mathbb{R}^D$, parameterized by $\boldsymbol{\phi}$, to embed each instance into a latent representation $\boldsymbol{z}_j = f_{\boldsymbol{\phi}}(\boldsymbol{x}_j)$, thereby forming the representation set $\mathcal{Z} = \{\boldsymbol{z}_j\}_{j=1}^{|\mathcal{X}|}$.  
Depending on how the latent representations $\mathcal{Z}$ are utilized to satisfy constraints and perform clustering, existing DCC methods can be categorized into two paradigms: \textit{end-to-end DCC} and \textit{deep constraint embedding}.

\vspace*{7mm}

\paragraph{End-to-end DCC.}
End-to-end DCC introduces additional anchors $\mathcal{A}$ to structure $\mathcal{Z}$, enabling soft cluster assignment $\mathcal{Q}$ that satisfy constraints $\mathcal{C}$. An activation function $\sigma_{\mathcal{A}}: \mathcal{Z} \to \mathcal{Q}$ maps each $\boldsymbol{z}_j$ to a soft assignment $\boldsymbol{q}_j = \sigma_{\mathcal{A}}(\boldsymbol{z}_j) \in \Delta^{K-1}$, where $\Delta^{K-1}$ is the probability simplex. Typically, $\mathcal{A}$ consists of $K$ class weight vectors in a classification layer, and is combined with $f_{\boldsymbol{\phi}}$ to form the classifier $h_{\boldsymbol{\phi}, \mathcal{A}}(\boldsymbol{x}) = \sigma_{\mathcal{A}}(f_{\boldsymbol{\phi}}(\boldsymbol{x}))$. Alternatively, $\mathcal{A}$ can be independent learnable parameters requiring specific initialization. A generic anchor-based pairwise loss function used in end-to-end DCC is defined as:
\[
\mathcal{L}(\sigma_{\mathcal{A}}, f_{\boldsymbol{\phi}}; \mathcal{C}) = \frac{1}{|\mathcal{C}|} \sum_{i=1}^{|\mathcal{C}|} \ell\left(\sigma_{\mathcal{A}}(f_{\boldsymbol{\phi}}(\boldsymbol{x}_{a_i})), \sigma_{\mathcal{A}}(f_{\boldsymbol{\phi}}(\boldsymbol{x}_{b_i})), y_i\right).
\]
This loss function measures how well $\mathcal{Q}$ satisfies $\mathcal{C}$, enabling joint optimization of $\boldsymbol{\phi}$ and $\mathcal{A}$.

\vspace*{7mm}

\paragraph{Deep constraint embedding.}
Deep constraint embedding methods independently trains $f_{\boldsymbol{\phi}}$ to learn a latent embedding representation $\mathcal{Z}$ that satisfies constraints $\mathcal{C}$, by minimizing the distances between positive pairs and maximizing those between negative pairs. 
Such representation learning operates without anchors and is driven by a generic anchor-free pairwise loss function, defined as:
\[
    \mathcal{L}(f_{\boldsymbol{\phi}}; \mathcal{C}) = \frac{1}{|\mathcal{C}|} \sum_{i=1}^{|\mathcal{C}|} \ell\left(f_{\boldsymbol{\phi}}(\boldsymbol{x}_{a_i}), f_{\boldsymbol{\phi}}(\boldsymbol{x}_{b_i}), y_i\right)
    \label{eq:anchor-free-loss}
\]
Optimizing this loss encourages $\mathcal{Z}$ to faithfully encode all pairwise constraints, thereby enabling CC to be treated as an unsupervised clustering task to determine $\mathcal{S}$.

In both DCC paradigms, the encoder \( f_{\boldsymbol{\phi}} \) can be paired with a decoder \( g_{\boldsymbol{\phi}'}: \mathcal{Z} \to \mathcal{X} \), parameterized by \( \boldsymbol{\phi}' \), to form a deep autoencoder. This configuration leverages unconstrained instances in \( \mathcal{X} \) to enrich the representation in \( \mathcal{Z} \).

\vspace*{15mm}

\section{Algorithms}
\label{sec:alg}

In this appendix, we present our SpherePair CC algorithm and the PCA-based cluster number inference algorithm introduced in Section \ref{sec:spherepair}, 
as detailed in \cref{alg:SpherePairs,alg:inferK}.

\newpage

\begin{algorithm}[htbp]
    \caption{SpherePair constraint clustering}
    \label{alg:SpherePairs}
\begin{algorithmic}
    \STATE \textbf{I. Angular constraint embedding learning}
    \STATE {\bfseries Input:} Training dataset $\mathcal{X}$, constraints $\mathcal{C}$, training epochs $T$, batch sizes $|\mathcal{B}_c|$ and $|\mathcal{B}_x|$, 
    embedding dimension $D$,  trade-off factor $\lambda$,  parametrized autoencoder with encoder $f_{\boldsymbol{\phi}}$ and decoder $g_{\boldsymbol{\phi}'}$
    
    \STATE Initialize the autoencoder parameters: $\boldsymbol{\phi}_0$ and $\boldsymbol{\phi}'_0$
    \FOR{$t = 1, 2, \dots, T$}
        \STATE Sample mini-batches $\mathcal{B}_c=\{(a_i, b_i, y_i)\}_{i=1}^{|\mathcal{B}_c|}$ from $\mathcal{C}$, and $\mathcal{B}_c =\{\boldsymbol{x}_j\}_{j=1}^{|\mathcal{B}_x|}$ from $\mathcal{X}$
        \STATE Obtain latent embeddings:  $\boldsymbol{z}_j = f_{\boldsymbol{\phi}}(\boldsymbol{x}_j), \forall \boldsymbol{x}_j \in \mathcal{X}$
        \STATE Obtain reconstructions $\hat{\boldsymbol{x}}_j, \forall \boldsymbol{x}_j \in \mathcal{X}$ by Eq.~\ref{eq:x_j_hat}
        \STATE Compute gradients $\frac{\partial \mathcal{L}}{\partial \boldsymbol{\phi}}$ and $\frac{\partial \mathcal{L}}{\partial \boldsymbol{\phi}'}$ with $\mathcal{L}$ (in Eq.~\ref{eq:L}) via $\mathcal{L}_{\text{ang}}$ (in Eq.~\ref{eq:angular-pairs}) and $\mathcal{L}_{\text{recon}}$ (in Eq.~\ref{eq:L_recon})
        \STATE Update $\boldsymbol{\phi}$, $\boldsymbol{\phi}'$ using stochastic gradient descent ($\tt SGD$): 
        $\boldsymbol{\phi}_t,{\boldsymbol{\phi}'}_t \leftarrow {\tt SGD}
        (
        \boldsymbol{\phi}_{t-1},{\boldsymbol{\phi}'}_{t-1}, 
        \frac{\partial \mathcal{L}}{\partial \boldsymbol{\phi}}, 
        \frac{\partial \mathcal{L}}{\partial \boldsymbol{\phi}'}
        )
        $
    \ENDFOR
    \STATE {\bfseries Output:} Optimal parameters,  $\boldsymbol{\phi}^*$ and $\boldsymbol{\phi}'^*$
    \vspace*{3mm}
    \STATE \textbf{II. Clustering on spherical representations}
    \STATE Obtain the optimal representations $\mathcal{Z}^* = \{f_{\boldsymbol{\phi}^*}(\boldsymbol{x}_j)\}_{j=1}^{|\mathcal{X}|}$
    \STATE Obtain the spherical representations $\mathcal{Z}_{\text{sphere}}$ by Eq.~\ref{eq:Z_sphere}
    \STATE Clustering with a chosen algorithm ${\tt Clustering}(\cdot)$:
    $\mathcal{S} \leftarrow {\tt Clustering}(\mathcal{Z}_{\text{sphere}}),~ \mathcal{S} =\{\mathcal{S}_k\}_{k=1}^K$
     \vspace*{3mm}
    \STATE \textbf{III. Prediction of unseen instances}
    \STATE {\bfseries Input:} Test dataset $\tilde{\mathcal{X}}$ 
    ($\tilde{\mathcal{X}} \cap \mathcal{X} = \varnothing$)
    \STATE Compute latent centroids $\{\boldsymbol{\mu}_k\}_{k=1}^K$ of $\mathcal{S}$ based on $\mathcal{Z}_{\text{sphere}}$
    \FOR{$\forall \tilde{\boldsymbol{x}} \in \tilde{\mathcal{X}} $}
        \STATE Obtain the spherical representation $\tilde{\boldsymbol{z}}_{\text{sphere}} = \text{Norm} \left( f_{\boldsymbol{\phi}^*}(\tilde{\boldsymbol{x}}) \right)$
        \STATE Assign $\tilde{\boldsymbol{z}}$ to cluster $k^*$ where $k^* = \arg\min_{k} \theta_{\tilde{\boldsymbol{z}}_{\text{sphere}}, \boldsymbol{\mu}_k}$
    \ENDFOR
\end{algorithmic}
\end{algorithm}

\vspace*{-2mm}

\begin{algorithm}[H]
    \caption{PCA-based cluster-number inference}
    \label{alg:inferK}
\begin{algorithmic}
    \STATE \textbf{Input:} Spherical representations $\mathcal{Z}_{\text{sphere}}\subset\mathbb{R}^D$, 
    training negative constraint subset $\mathcal{C}^-\!=\!\{(a_i,b_i,0)\}\!\subseteq\!\mathcal{C}$ covering all true clusters, tail ratio hyperparameter $\rho$ with $0 < \rho \ll 1$
    \vspace{1mm}
    \STATE Obtain subset of $\mathcal{Z}_{\text{sphere}}$ involved by $\mathcal{C}^-$: $\mathcal{Z}_{\text{sphere}}^{-} = \{\boldsymbol{z}_{a_i},\,\boldsymbol{z}_{b_i} \mid (a_i,b_i,0)\in\mathcal{C}^-\}\subseteq\mathcal{Z}_{\text{sphere}}$
    \STATE Conduct PCA on $\mathcal{Z}_{\text{sphere}}^{-}$ to obtain all top-$d$ subspace projections $\{\mathcal{Z}_{\mathrm{pca}}^{-\!,(d)}\}_{d=1}^D$
    \vspace{1mm}
    \FOR{$d = 1$ to $D$}
        \STATE Compute inter-cluster angles within $\mathcal{Z}_{\mathrm{pca}}^{-\!,(d)}$, obtain
        $\Theta^{(d)} = \{\theta_{\tilde{\boldsymbol{z}}_{a_i}^{(d)}\!,\tilde{\boldsymbol{z}}_{b_i}^{(d)}}\mid (a_i,b_i,0)\in\mathcal{C}^-\}$ 
        \STATE Select the $\rho$-fraction smallest inter-cluster angles 
        $\mathcal{M}^{(d)}={\tt Sort}_{\rho}(\Theta^{(d)})$,
        where ${\tt Sort}_{\rho}(\Theta^{(d)})$ is an operation that sorts all members of the set $\Theta^{(d)}$, retaining only the $\lceil \rho\times|\mathcal{C}^-|\rceil$ minimal elements.
        \STATE Compute tail average $\overline{\delta}_d = \operatorname{mean}(\mathcal{M}^{(d)})$
    \ENDFOR
    \vspace{1mm}
    \STATE Identify the onset of the plateau in sequence $\{\overline{\delta}_d\}_{d=1}^D$ as $d^\ast$
    \STATE \textbf{Output:} Estimated cluster number $\widehat{K} = d^\ast + 1$
\end{algorithmic}
\end{algorithm}

\section{Proofs}
\label{sec:proof_appendix}

In this appendix, we provide proofs for \cref{prop:z^*}, \cref{prop:requirement}, \cref{corollary:perturbation}, \cref{thm:bound}, \cref{corollary:optimal_omega}, \cref{thm:pca-angle-invariance}, and \cref{corollary:min-neg-angle}.

\subsection{Proof of \cref{prop:z^*}}
\label{sec:proof1}
    \begin{proof}
    Recall that $\mathcal{L}_{\text{ang}}$ in Eq.~\ref{eq:angular-pairs} takes the form
    \begin{equation*}
        \mathcal{L}_{\text{ang}} = -\frac{1}{|\mathcal{C}|} \sum_{i=1}^{|\mathcal{C}|} \Bigl( y_i \,\log \text{Sim}(a_i, b_i) \;+\; \bigl(1 - y_i\bigr)\,\log \bigl(1 - \text{Sim}(a_i, b_i)\bigr) \Bigr),
    \end{equation*}
    where
    \begin{equation*}
        \text{Sim}(a_i, b_i) \;=\;\frac{1}{2}
        \begin{cases}
        \cos \bigl(\theta_{\boldsymbol{z}_{a_i}, \boldsymbol{z}_{b_i}} \bigr) + 1, & \text{if } y_i = 1,\\
        \cos \bigl(\min \bigl(\omega \,\theta_{\boldsymbol{z}_{a_i}, \boldsymbol{z}_{b_i}}, \,\pi\bigr)\bigr) + 1, & \text{if } y_i = 0.
        \end{cases}
    \end{equation*}
    Suppose there is an $\mathcal{Z}^* = \{\boldsymbol{z}_j^*\}_{j=1}^{|\mathcal{X}|}\subset\mathbb{R}^D$ such that $\mathcal{L}_{\text{ang}}=0$ thus satisfies all pairwise relationships $\{(a_i, b_i, y_i)\}$ derived from the ground truth partition $\mathcal{S}^*$ without causing conflicts. Consider two cases:
    \begin{enumerate}
        \item $\boldsymbol{x}_{a_i},\boldsymbol{x}_{b_i}\in\mathcal{S}_k^*$ with $y_i=1$.  
        To achieve zero loss for the corresponding positive pair term in $\mathcal{L}_{\text{ang}}$, we must have $\text{Sim}(a_i,b_i)=1$, and this implies $\cos \bigl(\theta_{\boldsymbol{z}_{a_i}^*, \boldsymbol{z}_{b_i}^*}\bigr) \;=\; 1$. Hence,
        \[
            \theta_{\boldsymbol{z}_{a_i}^*, \boldsymbol{z}_{b_i}^*} \;=\; 0,
        \]
        implying that any two instances in the same cluster $\mathcal{S}_k^*$ must lie at angle of zero degree in the optimal angular representation~$\mathcal{Z}^*$.
    
        \item $\boldsymbol{x}_{a_i}\in\mathcal{S}_k^*,\,\boldsymbol{x}_{b_i}\in\mathcal{S}_{k'}^*$ with $k\neq k'$ and $y_i=0$.  
        To achieve zero loss for this negative pair, we must have $\text{Sim}(a_i,b_i)=0$. By definition, this simplifies to $\cos \bigl(\min(\omega\,\theta_{\boldsymbol{z}_{a_i}^*, \boldsymbol{z}_{b_i}^*},\,\pi)\bigr) = -1$.
        Thus, $\min(\omega\,\theta_{\boldsymbol{z}_{a_i}^*, \boldsymbol{z}_{b_i}^*},\,\pi) = \pi$. Consequently, $\omega\,\theta_{\boldsymbol{z}_{a_i}^*, \boldsymbol{z}_{b_i}^*} \geq \pi$, leading to
        \[
            \theta_{\boldsymbol{z}_{a_i}^*, \boldsymbol{z}_{b_i}^*} 
            \;\geq\; \frac{\pi}{\omega}.
        \]
        Therefore, any two instances that belong to different ground-truth clusters must have their feature vectors separated by an angle of at least $\frac{\pi}{\omega}$ in $\mathcal{Z}^*$.
    \end{enumerate}
    
    Since these two conditions hold for every pair derived from $\mathcal{S}^*$ and collectively yield zero loss, it follows that for each $\boldsymbol{x}_j\in\mathcal{S}_k^*$ and $\boldsymbol{x}_{j'}\in\mathcal{S}_{k'}^*$,
    \[
        \theta_{\boldsymbol{z}_j^*, \boldsymbol{z}_{j'}^*}
        \begin{cases}
        = 0, & \text{if } k = k',\\
        \ge \frac{\pi}{\omega}, & \text{if } k \neq k'.
        \end{cases}
    \]
    \end{proof}

\subsection{Proof of \cref{prop:requirement}}
\label{sec:proof2}
    \begin{proof}
   We establish the proof of conditions (i) and (ii) by showing both necessity and sufficiency.
    
    \noindent
    \paragraph{Necessity.}\;
    Assume that there exists some $\mathcal{Z}^*=\{\boldsymbol{z}_j^*\}_{j=1}^{|\mathcal{X}|}\subset \mathbb{R}^D$ 
    satisfying the conflict-free condition in \cref{prop:z^*} and that, 
    for each constraint $(a_i,b_i,y_i)$ from $\mathcal{S}^*$, 
    the angle $\theta_{\boldsymbol{z}_{a_i}^*,\boldsymbol{z}_{b_i}^*}$ 
    can be uniquely determined by $(a_i,b_i,y_i)$. 
    We show that all cross-cluster angles 
    $\{\theta_{\boldsymbol{z}_j^*, \boldsymbol{z}_{j'}^*}\mid 
     \boldsymbol{x}_j\in\mathcal{S}_k^*, \boldsymbol{x}_{j'}\in\mathcal{S}_{k'\neq k}^*\}$
    must be the same.
    
    By \cref{prop:z^*}, any negative pair $(\boldsymbol{x}_j,\boldsymbol{x}_{j'},0)$ 
    from different clusters satisfies 
    $\theta_{\boldsymbol{z}_j^*,\boldsymbol{z}_{j'}^*}\geq \tfrac{\pi}{\omega}$. 
    Suppose, for contradiction, 
    that $\mathcal{Z}^*$ is not equidistant among clusters. 
    Then there exist two distinct cross-cluster pairs whose angles differ; 
    let 
    \[
       \theta_{\boldsymbol{z}_{p}^*,\boldsymbol{z}_{q}^*}
       \;=\;
       \min \Bigl\{
          \theta_{\boldsymbol{z}_j^*, \boldsymbol{z}_{j'}^*}:
          \boldsymbol{x}_j\in\mathcal{S}_k^*,\,
          \boldsymbol{x}_{j'}\in\mathcal{S}_{k'\neq k}^*
       \Bigr\}
    \]
    be the smallest cross-cluster angle, 
    and let 
    $\theta_{\boldsymbol{z}_{u}^*,\boldsymbol{z}_{v}^*}
     > \theta_{\boldsymbol{z}_{p}^*,\boldsymbol{z}_{q}^*}$ 
    be a strictly larger cross-cluster angle. 
    By conflict-free condition, 
    $\theta_{\boldsymbol{z}_{p}^*,\boldsymbol{z}_{q}^*} \ge \tfrac{\pi}{\omega}$,
    so certainly
    $\theta_{\boldsymbol{z}_{u}^*,\boldsymbol{z}_{v}^*} > \tfrac{\pi}{\omega}$.
    For the negative pair $(u,v,0)$, 
    the constraint alone enforces an angular separation at least $\tfrac{\pi}{\omega}$ 
    but does not expand the angle to $\theta_{\boldsymbol{z}_{u}^*,\boldsymbol{z}_{v}^*}$. 
    Hence the separation 
    $\theta_{\boldsymbol{z}_{u}^*,\boldsymbol{z}_{v}^*}$ in $\mathcal{Z}^*$
    cannot be uniquely determined by the negative constraint $(u,v,0)$, 
    contradicting the hypothesis.
    Therefore, it is necessary that $\mathcal{Z}^*$ be equidistant among clusters.

    \noindent
    \paragraph{Sufficiency.}\;
    Conversely, assume that $\mathcal{Z}^*$ is equidistant among clusters: 
    every cross-cluster pair has the same angle $\theta^*>0$. 
    Then: 
    (i) For each negative pair $(j,j',0)$, we have 
    $\theta_{\boldsymbol{z}_j^*,\boldsymbol{z}_{j'}^*} = \theta^*$.
    If we set $\omega = \omega^* = \frac{\pi}{\theta^*}$,
    then by \cref{prop:z^*}, every cross-cluster angle satisfies a separation 
    $\tfrac{\pi}{\omega}=\theta^*$.  
    Hence each negative constraint $(j,j',0)$ 
    can uniquely determines each angular separation $\theta_{\boldsymbol{z}_j^*,\boldsymbol{z}_{j'}^*}$ in this equidistant $\mathcal{Z}^*$;  
    (ii) As for any positive constraint $(j,j',1)$,
    \cref{prop:z^*} forces each intra-cluster angle to be $0$.  
    Thus each positive constraint $(j,j',1)$ 
    can also uniquely determines each intra-cluster angle in $\mathcal{Z}^*$.  
     
    Hence, the necessity and sufficiency of conditions (i) and (ii) in \cref{prop:requirement} are shown.
    
    \end{proof}

\subsection{Proof of \cref{corollary:perturbation}}
\label{sec:proof3}
\begin{proof}
Recall from Eq.~\ref{eq:angular-pairs} that the average angular loss is
\begin{equation*}
    \mathcal{L}_{\mathrm{ang}}
    = -\frac{1}{|\mathcal{C}|} \sum_{i=1}^{|\mathcal{C}|}
    \Bigl( y_i \,\log \text{Sim}(a_i,b_i) \;+\; (1-y_i)\,\log\bigl(1 - \text{Sim}(a_i,b_i)\bigr) \Bigr),
\end{equation*}
where
\begin{equation*}
    \text{Sim}(a_i,b_i) =
    \frac{1}{2}
    \begin{cases}
        \cos\!\bigl(\theta_{z_{a_i},z_{b_i}}\bigr) + 1, & \text{if } y_i = 1, \\[1ex]
        \cos\!\bigl( \min(\omega\,\theta_{z_{a_i},z_{b_i}},\pi) \bigr) + 1, & \text{if } y_i = 0.
    \end{cases}
\end{equation*}

Given $\mathcal{L}_{\mathrm{ang}} \le \varepsilon$ with $0 < \varepsilon \ll 1$.

\noindent
\paragraph{Positive constraints.}
For any $(a_i,b_i,1) \in \mathcal{C}$, its individual loss term is
\[
    \ell_i^+ \;=\; -\log\!\left( \frac{\cos(\theta_{z_{a_i},z_{b_i}})+1}{2} \right).
\]
Since the average loss is at most $\varepsilon$, each term satisfies $\ell_i^+ \le |\mathcal{C}|\,\varepsilon$, hence
\[
    \frac{\cos(\theta_{z_{a_i},z_{b_i}})+1}{2} \;\ge\; e^{-|\mathcal{C}|\varepsilon}
    \quad\Longrightarrow\quad
    \cos(\theta_{z_{a_i},z_{b_i}}) \;\ge\; 2e^{-|\mathcal{C}|\varepsilon} - 1.
\]
Therefore
\[
    0 \;\le\; \theta_{z_{a_i},z_{b_i}}
    \;\le\; \arccos\!\bigl(2e^{-|\mathcal{C}|\varepsilon}-1\bigr)
    \;=:\; \Delta^+(\varepsilon).
\]
For small $|\mathcal{C}|\varepsilon$, a Taylor expansion of $e^{-|\mathcal{C}|\varepsilon}$ and $\arccos$ gives $\Delta^+(\varepsilon) \approx 2\sqrt{|\mathcal{C}|\varepsilon}$.

\noindent
\paragraph{Negative constraints.}
For any $(a_i,b_i,0) \in \mathcal{C}$ with $\theta_{z_{a_i},z_{b_i}} < \tfrac{\pi}{\omega}$, the loss term is
\[
    \ell_i^- \;=\; -\log\!\left( 1 - \frac{\cos(\omega\,\theta_{z_{a_i},z_{b_i}})+1}{2} \right).
\]
The bound $\ell_i^- \le |\mathcal{C}|\,\varepsilon$ implies
\[
    1 - \frac{\cos(\omega\,\theta_{z_{a_i},z_{b_i}})+1}{2} \;\ge\; e^{-|\mathcal{C}|\varepsilon}
    \quad\Longrightarrow\quad
    \cos(\omega\,\theta_{z_{a_i},z_{b_i}}) \;\le\; 1 - 2e^{-|\mathcal{C}|\varepsilon}.
\]
Since $\theta_{z_{a_i},z_{b_i}}$ is close to $\frac{\pi}{\omega}$, we write
\[
    \omega\,\theta_{z_{a_i},z_{b_i}} \;\ge\; \pi - \arccos\!\bigl(1 - 2e^{-|\mathcal{C}|\varepsilon}\bigr),
\]
which rearranges to
\[
    0 \;\le\; \frac{\pi}{\omega} - \theta_{z_{a_i},z_{b_i}}
    \;\le\; \frac{1}{\omega}\,\arccos\!\bigl(1 - 2e^{-|\mathcal{C}|\varepsilon}\bigr)
    \;=:\; \Delta^-(\varepsilon).
\]
Again, for small $|\mathcal{C}|\varepsilon$, Taylor expansion yields $\Delta^-(\varepsilon) \approx \frac{2\sqrt{|\mathcal{C}|\varepsilon}}{\omega}$.

\noindent
Combining the two cases gives the desired bounds in \cref{corollary:perturbation}.

\end{proof}

\subsection{Proof of \cref{thm:bound}}
\label{sec:proof4}
    \begin{proof}
    We reformulate the statement as a geometric problem of arranging $K$ distinct unit vectors in $\mathbb{R}^D$, ensuring that the angle between any two distinct vectors remains constant. This arrangement satisfies the "equidistant clusters" condition described in \cref{prop:requirement}.

    Let us denote these $K$ vectors as
    \[
      \{\boldsymbol{u}_1, \boldsymbol{u}_2, \dots, \boldsymbol{u}_K\} \;\subset\; \mathbb{R}^D,
      \quad \|\boldsymbol{u}_k\|=1,
    \]
    with the property that
    \[
      \theta_{\boldsymbol{u}_k,\;\boldsymbol{u}_{k'}}
      \;=\;\theta^*
      \quad\text{for all } 1 \le k \neq k' \le K.
    \]
    We show that such a uniform arrangement exists if and only if $D \ge K-1$, and derive the bounds of $\theta^*$ for different $D$.
    
        \noindent \paragraph{Case (i): $D < K-1$.}  
            Assume, for contradiction, that there exist $K$ unit vectors 
            $\{\boldsymbol{u}_1,\dots,\boldsymbol{u}_K\}\subset \mathbb{R}^D$ with $D < K-1$ 
            and a common angle~$\theta^*>0$. 
            Let $c=\cos\theta^*$.  
            Consider the $K\times K$ Gram matrix $G$ of these vectors, where
            \[
              G_{ij} 
              \;=\; 
              \boldsymbol{u}_i\cdot \boldsymbol{u}_j
              \;=\;
              \begin{cases}
              1, & \text{if } i = j,\\
              c, & \text{if } i \neq j.
              \end{cases}
            \]
            We may rewrite $G$ as:
            \[
              G 
              \;=\;
              (1 - c)\,I \;+\; c\,J,
            \]
            where $I$ is the $K\times K$ identity matrix and $J$ is the $K\times K$ matrix of all ones.
            
            To identify the eigenvalues of $G$, note that $J$ has one eigenvalue $K$ 
            (with eigenvector $\boldsymbol{1}=(1,\dots,1)^\top$) 
            and $K-1$ eigenvalues equal to $0$.  
            Hence $G$ inherits:
            \begin{itemize}
            \item One eigenvalue 
            \(
              (1-c)\cdot 1 + c\cdot K 
              \;=\; 1 + c\,(K-1),
            \)
            associated with $\boldsymbol{1}$.
            \item $K-1$ eigenvalues 
            \(
              (1-c)\cdot 1 + c\cdot 0 
              \;=\; 1-c,
            \)
            corresponding to any vector orthogonal to $\boldsymbol{1}$.
            \end{itemize}
            Since $c = \cos\theta^* < 1$ (ensuring the $K$ vectors are pairwise distinct), 
            it follows that $1-c > 0$.  
            Thus $G$ has at least $K-1$ strictly positive eigenvalues, 
            which implies 
            \[
              \operatorname{rank}(G) 
              \;\ge\; K - 1.
            \]
            On the other hand, because all $\boldsymbol{u}_i$ lie in $\mathbb{R}^D$, 
            the dimension of the subspace spanned by them is at most $D$, 
            so $\operatorname{rank}(G)\le D$.  
            Therefore we must have 
            \[
              K-1 
              \;\le\; \operatorname{rank}(G) 
              \;\le\; D,
            \]
            contradicting $D < K-1$.  
            Hence, there can be no such $K$ unit vectors in $\mathbb{R}^D$ whose pairwise angles are all equal, 
            and consequently, no valid $\omega$ (i.e.\ no angle $\theta^*$) realizes the equidistant configuration 
            when $D < K-1$.\\
    
        \noindent \paragraph{Case (ii): $D = K-1$.}  
            In a space of dimension exactly $K-1$, 
            it is both necessary and sufficient that the $K$ vectors form a regular simplex. 
            The same Gram matrix argument above now forces $\operatorname{rank}(G) = K-1$, 
            hence the unique way for $K$ vectors to remain all equiangular is 
            \[
              1 + c (K-1) = 0 
              \quad\Longrightarrow\quad
              c = -\,\frac{1}{\,K-1\,},
              \quad
              \theta^* \;=\;\arccos \bigl(-\frac{1}{\,K-1}\bigr).
            \]
            In terms of \cref{prop:requirement}, the uniform cross-cluster angle is $\theta^*$, 
            so $\omega = \omega^*$ must satisfy 
            \[
              \theta^* 
              \;=\; 
              \frac{\pi}{\,\omega^*}
              \;\;\Longrightarrow\;\;
              \omega^*
              \;=\;
              \frac{\pi}{\theta^*}
              \;=\;
              \frac{\pi}{\,\arccos \bigl(-\tfrac{1}{K-1}\bigr)}.
            \]
            This $\omega^*$ is unique for $D=K-1$.\\
    
        \noindent \paragraph{Case (iii): $D \ge K$.}\; 
            Suppose we wish to place $K$ unit vectors 
            $\{\boldsymbol{u}_1,\dots,\boldsymbol{u}_K\}\subset \mathbb{R}^D$ 
            so that every pair of distinct vectors has a common angle~$\theta^*>0$.  
            Let $c=\cos\theta^*$, and form the corresponding $K\times K$ Gram matrix as before.
            As in the preceding cases, we may rewrite $G$ as $G=(1-c)\,I + c\,J$, 
            where $I$ is the $K\times K$ identity and $J$ is the $K\times K$ all-ones matrix.  
            Its eigenvalues are
            \[
               \lambda_1 = 1 + c\,(K-1),
               \quad
               \lambda_2=\cdots=\lambda_K=1-c.
            \]
            For $G$ to be a valid Gram matrix of real vectors, all eigenvalues must be nonnegative:
            \[
               1-c \;\ge\;0
               \quad\Longrightarrow\quad
               c \;\le\;1,
               \qquad
               1 + c\,(K-1) \;\ge\;0
               \quad\Longrightarrow\quad
               c \;\ge\;-\,\frac{1}{\,K-1\,}.
            \]
            Since $c<1$ when the vectors are mutually distinct, we combine these to conclude
            \[
               -\frac{1}{\,K-1\,}
               \;\le\;
               c
               \;<\;
               1
               \quad\Longrightarrow\quad
               0
               \;<\;
               \theta^* 
               \;\le\;
               \arccos \bigl(-\frac{1}{K-1}\bigr).
            \]
            Thus, whenever $D\ge K$, any common angle $\theta^*$ up to 
            $\arccos(-\frac{1}{K-1})$ is feasible.  
            Equivalently, in terms of 
            $\omega=\tfrac{\pi}{\theta^*}$,
            we obtain
            \[
               \omega
               \;=\;
               \frac{\pi}{\,\theta^*\,}
               \;\;\ge\;\;
               \frac{\pi}{\,\arccos \bigl(-\frac{1}{K-1}\bigr)}.
            \]
            Geometrically, this reflects the fact that we only need a subspace of dimension 
            $K-1$ to place $K$ equiangular vectors, and 
            having additional dimensions ($D \ge K$) does not tighten the required angle---it simply allows the same arrangement (or more flexible ones) to fit in higher-dimensional ambient space.  
            Hence, any $\omega$ above the threshold $\pi \big/ \arccos(-\frac{1}{K-1})$ remains valid when $D \ge K$.
    
    Combining all three cases leads to the desired conclusions:  
\begin{itemize}
    \item[(i)] When $D < K-1$, such a valid $\omega$ does not exist.  
    \item[(ii)] When $D = K -1$, the unique valid $\omega$ is $\pi \big/ \arccos(-\frac{1}{K-1})$. 
    \item[(iii)] When $D \geq K$, the range of valid $\omega$ values is relaxed to $\omega \geq \pi \big/ \arccos(-\frac{1}{K-1})$.  
\end{itemize}
\end{proof}

\subsection{Proof of \cref{corollary:optimal_omega}}
\label{sec:proof5}
\begin{proof}
Given $D \ge K$, by \cref{thm:bound}, the admissible set is $\omega \ge \pi \big/ \arccos\!\bigl(-\tfrac{1}{K-1}\bigr)$.
Hence the minimal admissible value is
\[
\omega_{\min}^*(K)\;=\;\frac{\pi}{\arccos\!\bigl(-\tfrac{1}{K-1}\bigr)}.
\]

We now establish the stated properties:

\noindent \paragraph{Bounds.}\;
For $K=2$, $\arccos(-1)=\pi$, thus $\omega_{\min}^*(2)=\pi/\pi=1$.
For every $K>2$, $-\tfrac{1}{K-1}\in(-1,0)$, hence
\[
\arccos\!\Bigl(-\tfrac{1}{K-1}\Bigr)\in\Bigl(\tfrac{\pi}{2},\,\pi\Bigr),
\quad\Rightarrow\quad
1 \;<\; \omega_{\min}^*(K) \;<\; 2.
\]
Therefore $1 \le \omega_{\min}^*(K) < 2$ for all $K>1$.

\noindent \paragraph{Monotonicity.}\;
If $2\le K_1<K_2$, then $-\tfrac{1}{K_1-1}<-\tfrac{1}{K_2-1}$.
Since $\arccos(\cdot)$ is strictly decreasing on $[-1,1]$,
\[
\arccos\!\Bigl(-\tfrac{1}{K_1-1}\Bigr)
\;>\;
\arccos\!\Bigl(-\tfrac{1}{K_2-1}\Bigr),
\]
and thus
\[
\omega_{\min}^*(K_1)=\frac{\pi}{\arccos(-\tfrac{1}{K_1-1})}
\;<\;
\frac{\pi}{\arccos(-\tfrac{1}{K_2-1})}
=\omega_{\min}^*(K_2).
\]
Hence $\omega_{\min}^*(K)$ is strictly increasing in $K$.

\noindent \paragraph{Limit.}\;
As $K\to\infty$, we have $-\tfrac{1}{K-1}\to 0$, so by continuity
\[
\arccos\!\Bigl(-\tfrac{1}{K-1}\Bigr)\to \arccos(0)=\tfrac{\pi}{2},
\qquad
\Rightarrow\qquad
\omega_{\min}^*(K)=\frac{\pi}{\arccos(-\tfrac{1}{K-1})}\to 2.
\]

Combining the above completes the proof.

\end{proof}

\subsection{Proof of \cref{thm:pca-angle-invariance}}
\label{sec:proof6}
\begin{proof}
For notation, let $\{\boldsymbol{u}_k\}_{k=1}^K\subset\mathbb{R}^D$ be the $K$ cluster-representative unit vectors on the sphere (i.e., the common location of each cluster in $\mathcal{Z}_{\mathrm{sphere}}$), let $p_k:=|\mathcal{S}_k^*|/{|\mathcal{X}|}>0$ with $\sum_{k=1}^K p_k=1$ be the cluster frequencies, and set the sample mean
\[
\boldsymbol{m}:=\sum_{k=1}^K p_k\,\boldsymbol{u}_k.
\]
After standard centering by $\boldsymbol{m}$, the $k$-th cluster maps to the common centered vector
\[
\boldsymbol{v}_k:=\boldsymbol{u}_k-\boldsymbol{m}\quad (k=1,\dots,K).
\]
For an optimal spherical embedding $\mathcal{Z}_{\mathrm{sphere}}$, all instances in cluster $\mathcal{S}_k^*$ coincide at $\boldsymbol{u}_k$, therefore after centering they all coincide at $\boldsymbol{v}_k$. Denote
\[
U:=\operatorname{span}\{\boldsymbol{v}_1,\dots,\boldsymbol{v}_K\}\subset\mathbb{R}^D.
\]

\medskip
\noindent\textbf{Step 1 (Centered data lie in a $(K\!-\!1)$–dimensional subspace and $\operatorname{Im}(\Sigma)=U$).}
Since $\sum_{k=1}^K p_k\boldsymbol{v}_k=\sum_k p_k(\boldsymbol{u}_k-\boldsymbol{m})=\boldsymbol{0}$, the family $\{\boldsymbol{v}_k\}$ is linearly dependent and hence $\dim U\le K-1$. On the other hand, $\{\boldsymbol{u}_k\}$ are the $K$ affinely independent vertices of a regular simplex, whose affine hull has dimension $K-1$; translating this affine hull by $-\boldsymbol{m}$ yields the linear subspace $U$ through the origin. Hence $\dim U=K-1$.

Let the covariance after centering be
\[
\Sigma=\frac{1}{{|\mathcal{X}|}}\sum_{j=1}^{|\mathcal{X}|} (\operatorname{Norm}(\boldsymbol{z}_j^*)-\boldsymbol{m})(\operatorname{Norm}(\boldsymbol{z}_j^*)-\boldsymbol{m})^\top
=\sum_{k=1}^K p_k\,\boldsymbol{v}_k\boldsymbol{v}_k^\top.
\]
For any $\boldsymbol{w}\perp U$ we have $\boldsymbol{v}_k^\top\boldsymbol{w}=0$ and thus $\Sigma\boldsymbol{w}=\boldsymbol{0}$, so $\operatorname{Im}(\Sigma)\subseteq U$, where $\operatorname{Im}(\Sigma)$ is the column space of $\Sigma$ . Conversely, for any $\boldsymbol{u}\in U\setminus\{\boldsymbol{0}\}$, since $\{\boldsymbol{v}_k\}$ span $U$, there exists $k$ with $\boldsymbol{v}_k^\top\boldsymbol{u}\neq 0$, and therefore
\[
\boldsymbol{u}^\top \Sigma \boldsymbol{u}
=\sum_{k=1}^K p_k\,(\boldsymbol{v}_k^\top \boldsymbol{u})^2>0.
\]
Thus $\Sigma$ is positive definite on $U$ and vanishes on $U^\perp$, which implies $\operatorname{Im}(\Sigma)=U$. This conclusion does not depend on the specific values of $\{p_k\}$ beyond $p_k>0$.

\medskip
\noindent\textbf{Step 2 (For $d\ge K-1$, PCA projection preserves all pairwise angles).}
Let $W_d$ be the $d$–dimensional PCA subspace spanned by the top $d$ eigenvectors of $\Sigma$. By Step~1, $\Sigma$ is positive definite on $U$ and zero on $U^\perp$, hence its nonzero eigenspace is exactly $U$. Therefore, for any $d\ge K-1=\dim U$,
\[
U\subseteq W_d\subseteq U\oplus U^\perp=\mathbb{R}^D,
\]
where $\oplus$ denotes the direct sum of subspaces. $\{\operatorname{Norm}(\boldsymbol{z}_j^*)-\boldsymbol{m}\}$ lie in $U$ and each $\operatorname{Norm}(\boldsymbol{z}_j^*)-\boldsymbol{m}$ equals some $\boldsymbol{v}_k$. Consequently, for any $\boldsymbol{v}_k$, the orthogonal projection onto $W_d$ satisfies $P_d\boldsymbol{v}_k=\boldsymbol{v}_k$ because $\boldsymbol{v}_k\in U\subseteq W_d$. PCA then applies an orthogonal change of basis inside $W_d$ (and, when $d=D$, an orthogonal transform in the full space). Orthogonal transforms preserve inner products, norms, and hence angles between any nonzero pair of vectors. Therefore, for any pair $(j,j')$ with nonzero projections and any $d,d' \ge K-1$,
\[
\theta_{\tilde{\boldsymbol{z}}_{j}^{(d)},\tilde{\boldsymbol{z}}_{j'}^{(d)}}
= \theta_{\tilde{\boldsymbol{z}}_{j}^{(d')},\tilde{\boldsymbol{z}}_{j'}^{(d')}},
\]
proving part (i).

\medskip
\noindent\textbf{Step 3 (For $d<K-1$, global invariance cannot hold).}
We work in the nondegenerate regime where the pairs compared satisfy $\tilde{\boldsymbol{z}}_j^{(d)}\neq\boldsymbol{0}\neq\tilde{\boldsymbol{z}}_{j'}^{(d)}$, i.e., $P_d\boldsymbol{v}_k\neq \boldsymbol{0}$ for the involved clusters. Since $\dim U=K-1>d$, one can choose indices $\mathcal{I}=\{k_1,\dots,k_{d+1}\}$ so that $P_d\boldsymbol{v}_{k_{\tau}}\neq \boldsymbol{0}$ for all ${\tau}$, and $\{\boldsymbol{v}_{k_{\tau}}\}_{{\tau}=1}^{d+1}$ are linearly independent in $U$. Form the matrix $X=[\,\boldsymbol{v}_{k_1}\ \cdots\ \boldsymbol{v}_{k_{d+1}}\,]\in\mathbb{R}^{D\times(d+1)}$ with Gram matrix $G:=X^\top X\in\mathbb{R}^{(d+1)\times(d+1)}$, which is positive definite (hence $\operatorname{rank}(G)=d+1$). Let $\widetilde{G}:=(P_dX)^\top(P_dX)=X^\top P_d X$, whose rank is at most $d$ because $P_dX$ has columns in the $d$–dimensional subspace $W_d$.

Assume, for contradiction, that the angle invariance of part (i) holds for \emph{all} admissible pairs among the set $\{\boldsymbol{v}_{k_{\tau}}\}_{{\tau}=1}^{d+1}$, i.e., for every ${\tau}\neq {\tau}'$ with $P_d\boldsymbol{v}_{k_{\tau}},P_d\boldsymbol{v}_{k_{\tau'}}\neq \boldsymbol{0}$,
\[
\frac{\langle P_d\boldsymbol{v}_{k_{\tau}},P_d\boldsymbol{v}_{k_{\tau'}}\rangle}
{\|P_d\boldsymbol{v}_{k_{\tau}}\|\,\|P_d\boldsymbol{v}_{k_{\tau'}}\|}
=
\frac{\langle \boldsymbol{v}_{k_{\tau}},\boldsymbol{v}_{k_{\tau'}}\rangle}
{\|\boldsymbol{v}_{k_{\tau}}\|\,\|\boldsymbol{v}_{k_{\tau'}}\|}.
\]
Setting $h_{\tau}:=\|P_d\boldsymbol{v}_{k_{\tau}}\|>0$ and $H:=\mathrm{diag}(h_1,\dots,h_{d+1})$, the above identities imply
\[
\widetilde{G}_{{\tau}{\tau'}}=\langle P_d\boldsymbol{v}_{k_{\tau}},P_d\boldsymbol{v}_{k_{{\tau}'}}\rangle
=h_{\tau} h_{{\tau}'}\,\langle \boldsymbol{v}_{k_{\tau}},\boldsymbol{v}_{k_{{\tau}'}}\rangle,
\quad\text{hence}\quad
\widetilde{G}=H\,G\,H.
\]
Since $H$ is invertible, $\operatorname{rank}(\widetilde{G})=\operatorname{rank}(G)=d+1$, which contradicts $\operatorname{rank}(\widetilde{G})\le d$. Therefore, when $d<K-1$, the invariance across $d$ in (i) cannot hold for all admissible pairs, proving part (ii).

\end{proof}

\subsection{Proof of \cref{corollary:min-neg-angle}}
\label{sec:proof7}
\begin{proof}

We follow the notation in \cref{thm:pca-angle-invariance}: let $\{\boldsymbol{u}_k\}_{k=1}^K\subset\mathbb{R}^D$ be the $K$ cluster-representative unit vectors on the sphere y, let $p_k=|\mathcal{S}_k^*|/{|\mathcal{X}|}>0$ with $\sum_{k=1}^K p_k=1$ be the cluster frequencies, and let the sample mean be $\boldsymbol{m}:=\sum_{k=1}^K p_k\,\boldsymbol{u}_k$.
After centering by $\boldsymbol{m}$, each cluster $\mathcal{S}_k^*$ collapses to a common vector $\boldsymbol{v}_k:=\boldsymbol{u}_k-\boldsymbol{m}$ where $k=1,\dots,K$.
Let $U=\operatorname{span}\{\boldsymbol{v}_1,\dots,\boldsymbol{v}_K\}\subset\mathbb{R}^D$.

From \cref{sec:proof6}, we know that $\dim U=K-1$ and the image space of the covariance matrix after
centering is exactly $U$. Hence when $d\ge K-1$, the PCA subspace $W_d$ contains $U$ and therefore $P_d\boldsymbol{v}_k=\boldsymbol{v}_k$. Subsequent orthogonal transformations preserve all pairwise angles.

\medskip
\noindent\textbf{(i) Case $K=2$.}
Here $\dim U=1$. Since
$\boldsymbol{m}=p_1\boldsymbol{u}_1+p_2\boldsymbol{u}_2$ with $p_2=1-p_1$, we obtain
\[
\boldsymbol{v}_1=(1-p_1)(\boldsymbol{u}_1-\boldsymbol{u}_2),\qquad
\boldsymbol{v}_2=-p_1(\boldsymbol{u}_1-\boldsymbol{u}_2).
\]
Thus $\boldsymbol{v}_1$ and $\boldsymbol{v}_2$ are collinear in opposite
directions, so their angle is always $\pi$. Any PCA projection with
$d\ge 1=K-1$ only applies an isometry within this line (and possibly an
orthogonal transform in the full space), leaving the angle unchanged.
Hence $\delta_d=\pi$ for all $d\ge 1$.

\medskip
\noindent\textbf{(ii) Case $K>2$: proving $\delta_1=0$ and invariance for $d\ge K-1$.}

\begin{enumerate}

\item \textit{$\delta_1=0$}: In the one-dimensional PCA projection, all vectors $\boldsymbol{v}_k$ are mapped to a single line. Since $K>2$ and each $p_k>0$, there must exist two distinct clusters whose projections lie on the same side and are nonzero, yielding an angle of $0$. Hence $\delta_1=0$.

\item \textit{Invariance for $d\ge K-1$}: By \cref{thm:pca-angle-invariance}, for any admissible pair $(j,j')$, when $d\ge K-1$ the angle $\theta_{\tilde{\boldsymbol{z}}_{j}^{(d)}\!,\tilde{\boldsymbol{z}}_{j'}^{(d)}}$
equals its counterpart at $d=K-1$ (or $D$). Therefore the minimal cross-cluster angle $\delta_d$ is constant for all $d\ge K-1$; denote this constant by $\delta_\star$.

\item \textit{Upper bound of $\delta_\star$}: Let $c=\cos\theta^*$ be the common inner product among $\{\boldsymbol{u}_k\}$. The Gram matrix of $U=[\,\boldsymbol{u}_1\ \cdots\ \boldsymbol{u}_K\,]$ is $G=(1-c)I+c\,\boldsymbol{1}\boldsymbol{1}^\top$. 
For any $k\neq k'$, consider $\boldsymbol{v}_k=\boldsymbol{u}_k-\boldsymbol{m}$ and $\boldsymbol{v}_{k'}=\boldsymbol{u}_{k'}-\boldsymbol{m}$.
Derivable from the Gram structure, angle $\theta_{\boldsymbol{v}_k,\boldsymbol{v}_{k'}}$ depends only on $\{p_i\}$: letting $S=\sum_{i=1}^K p_i^2$, one obtains
\begin{equation}\label{eq:cos-formula}
\cos\theta_{\boldsymbol{v}_k,\boldsymbol{v}_{k'}}
=\frac{S-p_k-p_{k'}}{\sqrt{(1-2p_k+S)(1-2p_{k'}+S)}}.
\end{equation}
When $p_1=\cdots=p_K=\frac{1}{K}$, $S=\frac{1}{K}$, and \eqref{eq:cos-formula} gives $\theta_{\boldsymbol{v}_k,\boldsymbol{v}_{k'}}=\arccos(-\frac{1}{K-1})$. 
All cross-cluster angles coincide, so $\delta_\star=\arccos(-\frac{1}{K-1})$, yielding the upper bound.
This bound is attained when clusters are balanced.

\item \textit{Lower bound of $\delta_\star$}: 
For a given pair $(k,k')$, write $r=1-p_k-p_{k'}=\sum_{i\neq k,k'}p_i>0$ and $s=\sum_{i\neq k,k'}p_i^2$, so $0<s\le r^2$.
Equation~\eqref{eq:cos-formula} can be rewritten as a function $f(s)$.
One checks that $f(s)$ is nondecreasing in $s$ on the relevant interval.
Hence the maximal value of $\cos\theta_{\boldsymbol{v}_k,\boldsymbol{v}_{k'}}$ (i.e., the minimal angle) occurs at $s=r^2$, which corresponds to the case when all remaining probability mass $r$ concentrates in a single cluster.
In this regime, one can show $\cos\theta_{\boldsymbol{v}_k,\boldsymbol{v}_{k'}}\le \frac{1}{2}$, thus $\cos\theta_{\boldsymbol{v}_k,\boldsymbol{v}_{k'}}\ge\arccos(\frac{1}{2})=\tfrac{\pi}{3}$.
Equality is never attained when all $p_k>0$, but as some $p_\ell\to 1$ (the others tending to $0$), $\theta_{\boldsymbol{v}_k,\boldsymbol{v}_{k'}}$ approaches $\tfrac{\pi}{3}$.
Therefore $\delta_\star>\tfrac{\pi}{3}$, with the infimum $\frac{\pi}{3}$ approached when one cluster dominates.

\end{enumerate}

\medskip
Thus, for $K=2$, we proved $\delta_d=\pi$ for all $d\ge 1$.
For $K>2$, we established $\delta_1=0$; for $d\ge K-1$, $\delta_d$ takes
a constant value $\delta_\star$; and
\[
\delta_\star \in (\tfrac{\pi}{3},\ \arccos(-\tfrac{1}{K-1})],
\]
with the upper bound $\arccos(-\tfrac{1}{K-1})$ attained in the balanced case and the lower bound $\tfrac{\pi}{3}$ approached in the highly imbalanced case.

\end{proof}

\section{Visualization of SpherePair representation learning in angular space}
\label{sec:3D_visualization_appendix}

To demonstrate how SpherePair learns equidistant spherical embeddings for different numbers of clusters in angular space, we perform a 3D visualisation experiment using subsets of the Reuters dataset \cite{REUTERS}, following the preprocessing steps outlined in the work \cite{DEC}. Specifically, the original training partition consists of 10,000 instances from four root categories: \emph{Corporate/Industrial}~(CCAT), \emph{Economics}~(ECAT), \emph{Government/Social}~(GCAT), and \emph{Markets}~(MCAT). Their respective instance counts in this training portion are 4,066 (CCAT), 2,888 (ECAT), 2,202 (GCAT), and 844 (MCAT). We form three data subsets by taking the first 2, 3, and all 4 categories, thus yielding scenarios with $K=2, 3, 4$. For each subset, we randomly sample 10k instance pairs and derive pairwise constraints based on whether the two instances belong to the same ground-truth category. Naturally, the constraint distribution varies across different $K$ due to the differing subset compositions.

Using the three data subsets and their respective 10k constraints, we train a SpherePair autoencoder comprising a fully connected encoder with hidden layers of sizes $500$, $500$, and $2000$ (mirrored by a symmetric decoder) and a 3-dimensional embedding layer ($D=3$). As guaranteed by \cref{thm:bound}, we set the negative-zone factor $\omega = \pi \big/ \arccos(-\frac{1}{K-1})$ as it is always valid when $D \ge K-1$, to ensure that the learned embeddings can form an equidistant arrangement on the unit sphere.

\begin{figure}[htbp]
    \begin{center}
    \centerline{\includegraphics[width=0.89\textwidth]{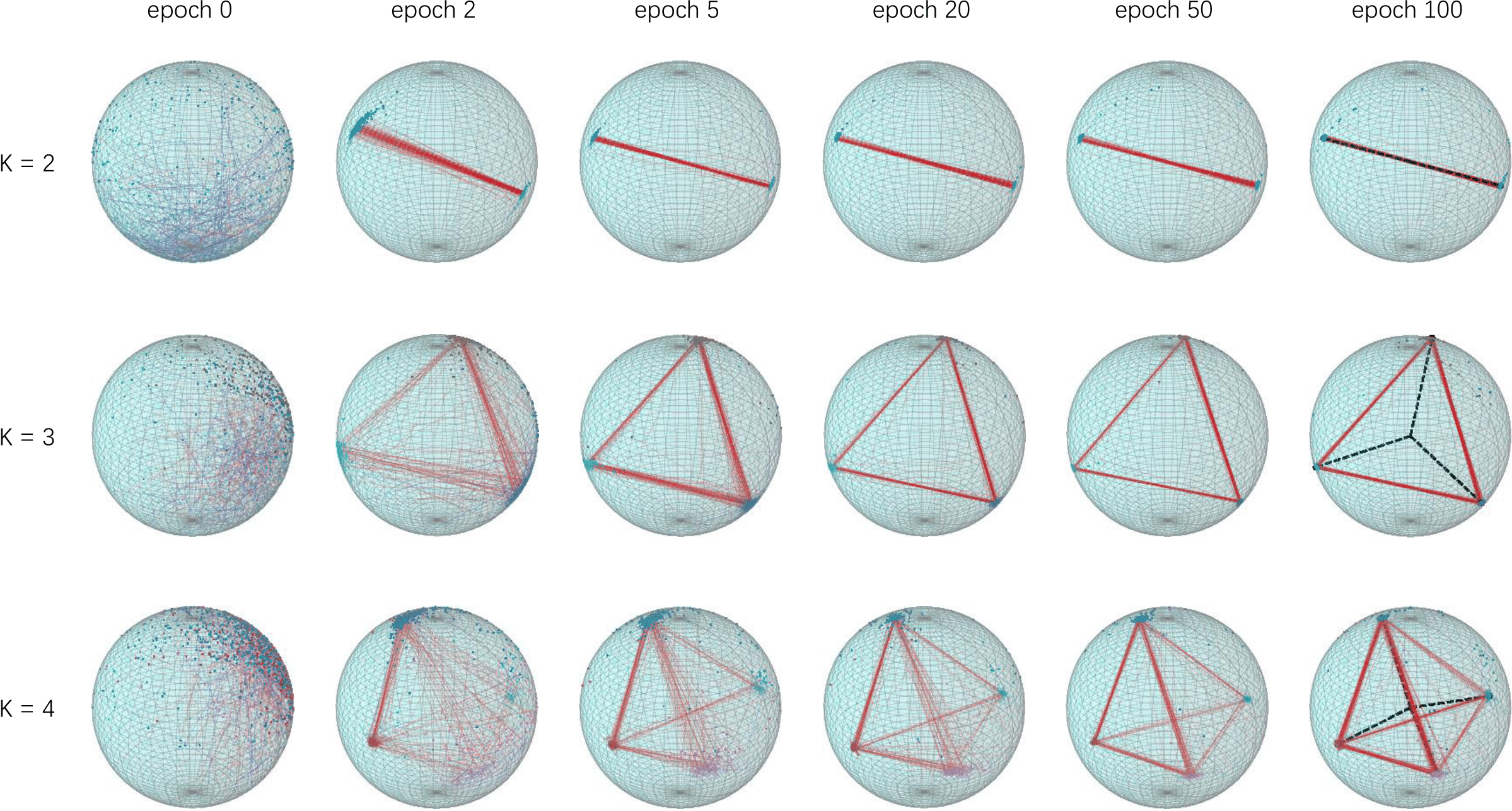}}
    \caption{Evolution of SpherePair embeddings on a 3D unit sphere for datasets with $K=2$, $3$, and $4$ clusters. Each row corresponds to one of the three subsets derived from the Reuters dataset, and each column shows a snapshot of the embeddings at a particular epoch. Different marker colors denote different ground-truth categories, and dashed lines represent randomly sampled pairwise constraints (blue for positive, red for negative). In the final column of each row, we illustrate the converged embeddings and highlight, in black dashed lines, the connections from the origin to each cluster centroid obtained by running K-means on $\mathcal{Z}_{\text{sphere}}$.}
    \label{fig:3d}
    \end{center}
    \vskip -0.2in
\end{figure}

Fig.~\ref{fig:3d} illustrates the evolution of the resulting spherical representations $\mathcal{Z}_{\text{sphere}}$ over the course of training. As the training progresses, the data points gradually separate on the 3D sphere, and the final embeddings form nearly regular simplex configurations in the $(K-1)$-dimensional subspace of $\mathbb{R}^3$, consistent with our theoretical insights in Sect.~\ref{sec:theoretical}. Worth noting, despite the imbalance in instance counts among different categories, the formation of equidistant spherical clusters is not contingent on ground truth clusters being of equal size. Thus, this 3D visualisation vividly illustrates and validates the theoretical insights into our proposed SpherePair representation learning framework.

\section{Details of experimental settings}
\label{sec:setting_appendix}

In this appendix, we provide a detailed description of our experimental settings to ensure that all necessary information is included and our results are fully replicable.

\subsection{Datasets}
\label{sec:datasets_appendix}

We provide detailed descriptions of the benchmark datasets employed for evaluation in experiments.

\begin{itemize}
    \item \textbf{CIFAR-10} \footnote{CIFAR-10 webpage: \url{https://www.cs.toronto.edu/~kriz/cifar.html}} \cite{CIFAR}: 
    Consists of 60,000 real-world 32$\times$32 color images spanning 10 classes, each class containing 6,000 images. CIFAR-10 has been widely used as a benchmark in DCC research \cite{SDEC,AutoEmbedder,VolMaxDCC}.

    \item \textbf{CIFAR-100-20} \footnote{CIFAR-100-20 webpage: \url{https://www.cs.toronto.edu/~kriz/cifar.html}} \cite{CIFAR}:
    A more complex extension of CIFAR-10, also having 60,000 real-world 32$\times$32 color images. CIFAR-100 originally contains 100 fine-grained classes, which can be grouped into 20 superclasses. In our experiments, we use these 20 superclasses (each containing 3,000 images) as the ground truth rather than the 100 classes. 
    
    \item \textbf{FashionMNIST} \footnote{FashionMNIST repository: \url{https://github.com/zalandoresearch/fashion-mnist}} \cite{FMNIST}:
    Contains 70,000 grayscale images of Zalando fashion products (10 categories) with a size of 28$\times$28 each. It is pre-split into 60,000 training images (6,000 per class) and 10,000 test images (1,000 per class). FashionMNIST also serves as a benchmark for DCC evaluation such as the methods \cite{AutoEmbedder,CIDEC1,DCGMM}.
    
    \item \textbf{ImageNet-10} \cite{Imagenet10}: 
    A subset of ImageNet \footnote{ImageNet webpage: \url{https://image-net.org/}} \cite{imagenet} with 10 classes, each containing 1,300 randomly selected color images (13,000 in total). The chosen classes are \texttt{n02056570}, \texttt{n02085936}, \texttt{n02128757}, \texttt{n02690373}, \texttt{n02692877}, \texttt{n03095699}, \texttt{n04254680}, \texttt{n04285008}, \texttt{n04467665}, \texttt{n07747607}. This ImageNet-10 is commonly used as a benchmark in both DCC \cite{VolMaxDCC} and unsupervised deep clustering \cite{Imagenet10,ContrastiveClustering}.
    
    \item \textbf{MNIST} \footnote{MNIST webpage: \url{http://yann.lecun.com/exdb/mnist/}} \cite{MNIST}:
    Contains 70,000 grayscale images of handwritten digits (0--9), each of size 28$\times$28. It is pre-split into 60,000 training images (6,000 per digit) and 10,000 test images (1,000 per digit). MNIST serves as a benchmark in DCC evaluation such as the methods \cite{SDEC,AutoEmbedder,CIDEC1,DCGMM}.

    \item \textbf{Reuters} \cite{REUTERS}:
    A subset of the RCV1\footnote{\label{fn:rcv1}RCV1 webpage: \url{https://trec.nist.gov/data/reuters/reuters.html}} corpus (a large-scale newswire collection of 804,414 English stories).
    We use the preprocessed version\footnote{Preprocessed Reuters repository: \url{https://github.com/piiswrong/dec}}
     from \cite{DEC}, which provides tf–idf features on the 2,000 most frequent words across documents sampled from four root categories: \emph{Corporate/Industrial} (CCAT), \emph{Economics} (ECAT), \emph{Government/Social} (GCAT), and \emph{Markets} (MCAT).
    The dataset is pre-split into 10,000 training samples and 2,000 test samples, with the four categories containing 4,066/2,888/2,202/844 training and 774/582/471/173 test samples, respectively.
    We treat these four categories as the ground-truth clusters in our experiments. Reuters has been widely adopted as a benchmark in DCC research \cite{AutoEmbedder,CIDEC1,DCGMM}.
    
    \item \textbf{STL-10} \footnote{STL-10 webpage: \url{https://cs.stanford.edu/~acoates/stl10/}} \cite{STL10}:
    Comprises 13,000 96$\times$96 color images from 10 classes, each class having 1,300 images. STL-10 is adopted by various DCC works \cite{SDEC,DCGMM,VolMaxDCC}.

    \item \textbf{RCV1-10} \footnote{We provide the preprocessed RCV1-10 subset: \url{https://github.com/spherepaircc/SpherePairCC/tree/main}}:
    Another subset of RCV1\footref{fn:rcv1} with highly imbalanced class distribution.
    We randomly selected 10 categories (C14, C18, C313, C42, E21, E311, GDEF, GODD, GWELF, M13) from the 103 available topics and removed documents carrying multiple labels within this set, yielding 177,669 single-label articles with the following class counts: 6,634 (C14), 51,145 (C18), 1,042 (C313), 10,954 (C42), 40,950 (E21), 1,679 (E311), 8,492 (GDEF), 2,743 (GODD), 903 (GWELF), and 53,127 (M13).
    We treat these 10 categories as the ground-truth clusters in our experiments. Tf–idf features were computed on the 2,000 most frequent word stems, yielding the RCV1-10 subset as a realistic benchmark for imbalanced constrained clustering.
\end{itemize}

\subsection{Baselines}
\label{sec:baselines_appendix}

We provide the detailed information on the state-of-the-art deep constrained clustering baselines used in our comparative study.

\paragraph{VanillaDCC}\cite{MCL}: A straightforward end-to-end anchor-based DCC model grounded in the Meta Classification Likelihood (MCL) loss:
\[
    \mathcal{L}_{\text{MCL}} 
    = 
    -\frac{1}{|\mathcal{C}|} \sum_{i=1}^{|\mathcal{C}|} \Bigl(y_i \,\log P_{\text{co}_i} + (1 - y_i)\,\log \bigl(1 - P_{\text{co}_i}\bigr)\Bigr),
\]
where $P_{\text{co}_i} = \boldsymbol{q}_{a_i} \boldsymbol{q}_{b_i}^\top \in [0,1]$ is a pairwise co-occurrence likelihood integrated into a logistic loss. Here, $\boldsymbol{q}_{a_i}$ and $\boldsymbol{q}_{b_i}$ are the soft assignments of constrained data pair $(\boldsymbol{x}_{a_i}, \boldsymbol{x}_{b_i})$ to $K$ clusters. By minimizing this loss, VanillaDCC learns a cluster assignment matrix $\mathcal{Q} = \{\boldsymbol{q}_j\}_{j=1}^N$ that respects the constraint set $\mathcal{C}$.

\paragraph{VolMaxDCC}\cite{VolMaxDCC}: An extension of VanillaDCC that modifies the MCL-based loss to handle confused memberships and incorporates an additional geometric regularization term controlled by a trade-off factor $\lambda$. Specifically, VolMaxDCC introduces a matrix $B$ (treated as an optimization variable and derived from a confusion matrix) into the pairwise co-occurrence likelihood and adds a volume maximization regularization term:
\[
\mathcal{L}_{\text{VolMax}} = -\frac{1}{|\mathcal{C}|} \sum_{i=1}^{|\mathcal{C}|} \Bigl(y_i \log P'_{\text{co}_i} + (1 - y_i) \log \bigl(1 - P'_{\text{co}_i}\bigr)\Bigr) - \lambda \log \det\bigl(\mathcal{Q}^\top\mathcal{Q}\bigr).
\]
Here $P'_{\text{co}_i} = \boldsymbol{q}_{a_i} B \, \boldsymbol{q}_{b_i}^\top \in [0,1]$ is the modified pairwise co-occurrence likelihood, and $\mathcal{Q}$ is treated as the cluster assignment matrix. The first term in $\mathcal{L}_{\text{VolMax}}$ adjusts the similarity measure to account for membership confusion, while the second term serves as a geometric regularization that encourages maximization of the volume of the Gram matrix $\mathcal{Q}^\top\mathcal{Q}$, thereby enhancing the separability and distinguishability of clusters. Both $B$ and $\mathcal{Q}$ are optimized during training. The choice of the trade-off factor $\lambda$ requires tuning to balance these effects. Overall, $\mathcal{L}_{\text{VolMax}}$ enables VolMaxDCC to handle noisy constraints caused by annotation confusion while promoting well-separated and distinguishable cluster assignments.

\paragraph{CIDEC}\cite{CIDEC1}: 
An end-to-end method that balances unsupervised representation learning and constrained clustering within a multi-task joint optimization framework. Specifically, CIDEC first encodes data into a latent space via a deep autoencoder and initializes $K$ learnable cluster anchors using K-means, where $K$ corresponds to the known number of ground-truth classes. It then alternates between supervised and unsupervised training phases during each epoch. In each iteration, the model jointly refines the autoencoder parameters and cluster assignments by combining the the following learning objectives to form a multi-objective loss function:

\begin{enumerate}
\item[(i)] The deep embedding clustering objective from \cite{DEC,IDEC}, which minimizes the KL divergence 
    \[
      \mathcal{L}_{\text{DEC}} = \sum_{j=1}^{|\mathcal{X}|} \sum_{k=1}^K p_{jk} \log\frac{p_{jk}}{q_{jk}},
    \] 
    between the soft assignment distributions $\boldsymbol{q}_j = (q_{j1},q_{j2},\ldots,q_{jK})$ and a target distribution $\boldsymbol{p}_j=(p_{j1},p_{j2},\ldots,p_{jK})$. Specifically, for each sample $j$, the target distribution is calculated as
    \[
      p_{jk} 
      = 
      \frac{q_{jk}^2 / f_k}{\sum_{k'=1}^K q_{jk'}^2 / f_{k'}}, 
      \quad\text{where}\; f_k = \sum_{j=1}^{|\mathcal{X}|} q_{jk},
    \]
    enhancing the influence of assignments with higher confidence,
   \item[(ii)] $\mathcal{L}_{\text{MCL}}$ for incorporating pairwise constraints, and
   \item[(iii)] A reconstruction loss to preserve the intrinsic data structure.
\end{enumerate}
 
This process employs two hyperparameters: $\lambda_1$ to balance the clustering and reconstruction losses, and $\lambda_2$ to weight the contributions of positive and negative constraints within the MCL-based loss. When no constraints are available, CIDEC reduces to the unsupervised clustering model IDEC~\cite{IDEC}.

\paragraph{DCGMM}\cite{DCGMM}: DCGMM combines a deep generative model (i.e., a variational autoencoder-like architecture) with a conditional Gaussian mixture framework to handle pairwise constraints. Specifically, DCGMM models each cluster as a Gaussian mixture component, where the number of components is set to the known ground-truth class count $K$. It incorporates constraint information by conditioning on positive and negative pairs, thereby reshaping the latent variable distributions. Additionally, DCGMM assigns a weight $|W_{i,j}|$ to each pairwise constraint between $\boldsymbol{x}_i$ and $\boldsymbol{x}_j$, reflecting the degree of certainty in that constraint. During training, it jointly optimizes (i) the variational likelihood of the autoencoder and (ii) a constraint-based term weighted by $|W_{i,j}|$ that pushes instances from positive pairs into the same mixture component and instances from negative pairs into different components. When no constraints are given, DCGMM reduces to VaDE~\cite{VaDE}, an unsupervised clustering model.

\paragraph{SDEC}\cite{SDEC}: SDEC combines an anchor-free constraint loss with the anchor-based deep embedding clustering objective from \cite{DEC} to learn cluster assignments that satisfy constraints. Specifically, SDEC minimizes a weighted sum of the unsupervised clustering loss $\mathcal{L}_{\text{DEC}}$, and a Euclidean distance-based constraint loss. The latter encourages positive pairs to be close in the latent space while pushing negative pairs apart:
\[
    \mathcal{L}_{\text{Euclidean}} 
    = \frac{1}{|\mathcal{C}|} \sum_{i=1}^{|\mathcal{C}|} \bigl(2y_i - 1\bigr)\,d\bigl(\boldsymbol{z}_{a_i}, \boldsymbol{z}_{b_i}\bigr)^2,
\]
where $d(\boldsymbol{z}_{a_i}, \boldsymbol{z}_{b_i}) = \|\boldsymbol{z}_{a_i} - \boldsymbol{z}_{b_i}\|_2$. SDEC introduces a trade-off factor $\lambda$ to balance these two objectives, optimizing the combined loss
\[
\mathcal{L}_{\text{SDEC}} = \mathcal{L}_{\text{DEC}} + \lambda \mathcal{L}_{\text{Euclidean}}.
\]
When no constraints are available, SDEC degenerates to the unsupervised clustering model DEC~\cite{DEC}.

\paragraph{AutoEmbedder}\cite{AutoEmbedder}: AutoEmbedder also learns pairwise embeddings in Euclidean space but does not include any unsupervised clustering loss, making it purely anchor-free for deep constraint embedding. AutoEmbedder uses an MSE loss based on a truncated Euclidean margin:
\[
 \mathcal{L}_{\text{MSE}} 
 = 
 \frac{1}{|\mathcal{C}|} \sum_{i=1}^{|\mathcal{C}|} 
 \Bigl(\min \{\max(0, d(\boldsymbol{z}_{a_i}, \boldsymbol{z}_{b_i})), \alpha \} \;-\; \alpha (1 - y_i)\Bigr)^2,
\]
where $d(\boldsymbol{z}_{a_i}, \boldsymbol{z}_{b_i}) = \|\boldsymbol{z}_{a_i} - \boldsymbol{z}_{b_i}\|_2$, and $\alpha$ is a manually chosen margin for the Euclidean distance. With $\mathcal{L}_{\text{MSE}}$, AutoEmbedder learn representations that pull positive pairs together while ensuring that negative pairs remain at least a margin $\alpha$ apart.

\subsection{Protocol}
\label{sec:protocol_appendix}

We provide the details of the experimental protocol used in our experiments.

\paragraph{Data splitting.}
In accordance with the protocol specified in the main text (Sect.~\ref{subsect:ex-setting}), we adhere to the original pre-split training/test partitions for FashionMNIST, MNIST, and the Reuters subset. 
For the remaining benchmarks (CIFAR-10, CIFAR-100-20, STL-10, ImageNet-10, and RCV1-10), we randomly split each dataset into 80\% for training and 20\% for testing, resulting in 48,000/12,000 samples for training/testing in CIFAR-10 and CIFAR-100-20, 10,400/2,600 in STL-10 and ImageNet-10, and 142,135/35,534 in RCV1-10. 
Following \cite{VolMaxDCC}, we then reserve 1,000 samples from each training split to form a validation set, used solely for hyperparameter tuning in baselines that require it (e.g., VolMaxDCC~\cite{VolMaxDCC}, AutoEmbedder~\cite{AutoEmbedder}).

\begin{figure}[t]
    \centering
    \includegraphics[width=0.9\textwidth]{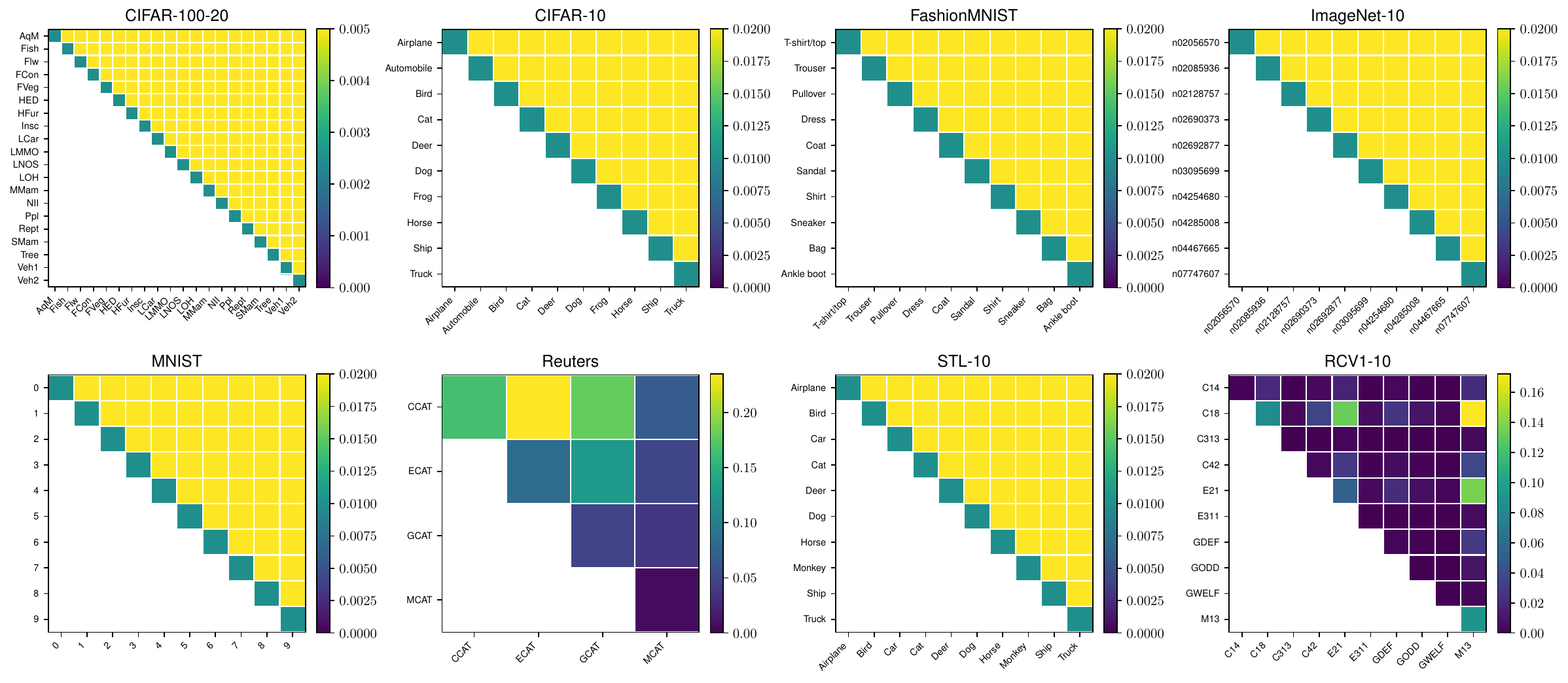}
    \caption{Distribution of randomly sampled constraints across ground-truth clusters. Each heatmap is a $K \times K$ symmetric matrix (represented by its upper triangular part) that illustrates the fraction of constraints originating from each corresponding pair of clusters $(k, k')$. The fractions in all matrix entries sum to 1 for each dataset.}
    \label{fig:constraint_heatmaps}
\end{figure}

\paragraph{Constraint set generation.}
For our standard experiments, we generate pairwise constraints via random sampling of training data pairs according to their ground-truth clusters. Concretely, for a training set with $N$ samples partitioned into $K$ ground-truth clusters $\mathcal{S}^* = \{\mathcal{S}_k^*\}_{k=1}^K$, we randomly select pairs of samples and assign a \emph{positive} constraint $(y_i=1)$ if both samples lie in the same cluster (i.e., $\boldsymbol{x}_{a_i}, \boldsymbol{x}_{b_i}\in \mathcal{S}_k^*$), or a \emph{negative} constraint $(y_i=0)$ if they come from different clusters. In principle, there are up to $\binom{|\mathcal{X}|}{2} = \frac{{|\mathcal{X}|}({|\mathcal{X}|}-1)}{2}$ possible constraints, among which each cluster $\mathcal{S}_k^*$ can yield up to $\frac{|\mathcal{S}_k^*|\,(|\mathcal{S}_k^*|-1)}{2}$ positive constraints and each cluster pair $(\mathcal{S}_k^*, \mathcal{S}_{k'}^*)$ with $k' \neq k$ can yield up to $|\mathcal{S}_k^*|\cdot|\mathcal{S}_{k'}^*|$ negative constraints. Consequently, the proportion of positive and negative constraints from different ground-truth clusters reflects the underlying class distribution. Fig.~\ref{fig:constraint_heatmaps} visualizes the distribution of the randomly sampled constraints (of any chosen size, e.g.\ 1k, 5k, or 10k) in the form of $K\times K$ heatmaps for the eight datasets. Unless otherwise stated, our experiments use this random sampling strategy for constructing constraint sets.

\begin{figure}[htbp]
    \centering
    \includegraphics[width=0.7\textwidth]{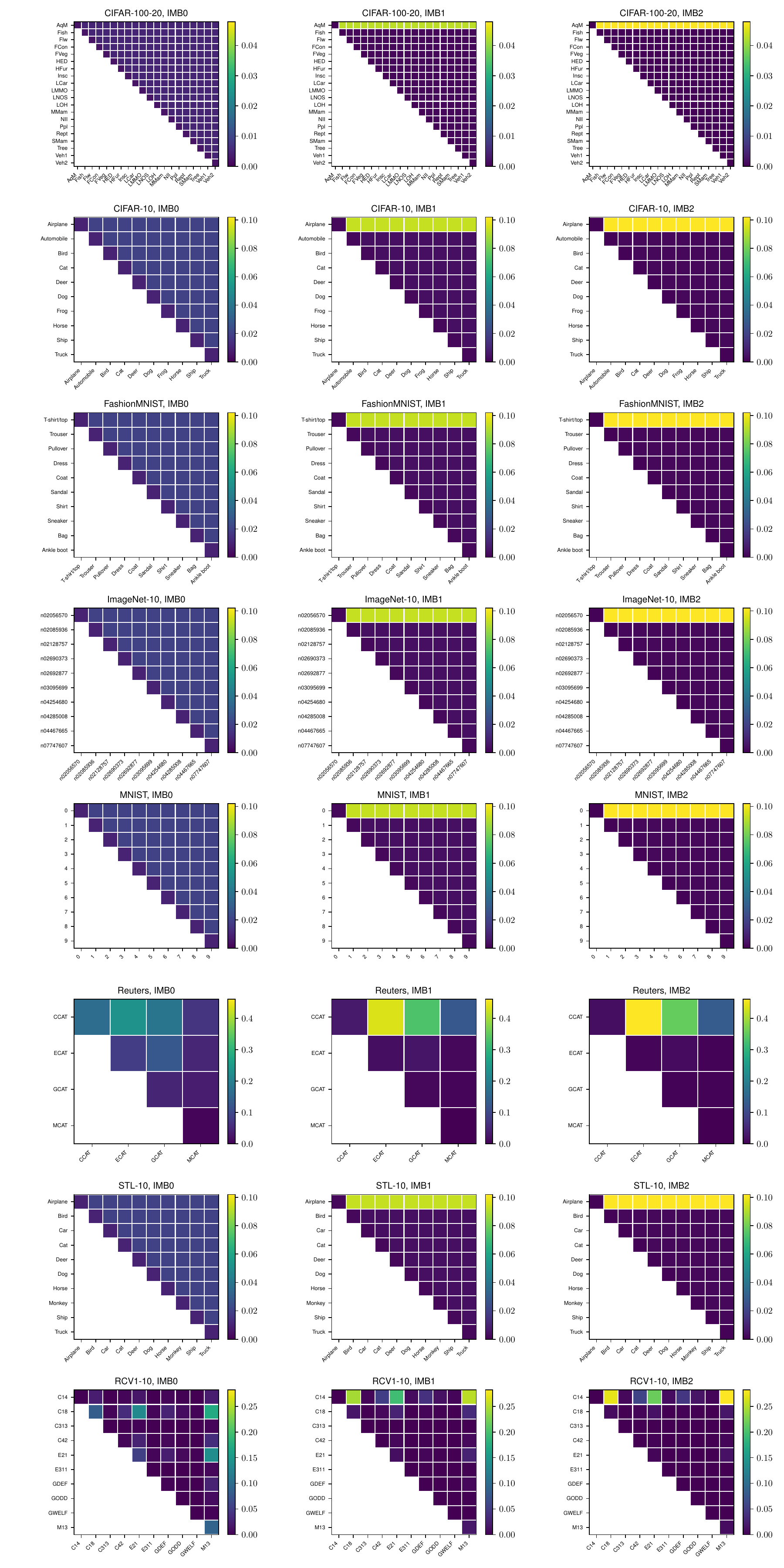}
    \caption{Heatmaps of imbalanced constraint distributions, $K\times K$ symmetric matrices (represented by its upper triangular part), for eight datasets. Rows correspond to datasets, while the three columns represent {\tt IMB0}, {\tt IMB1}, and {\tt IMB2}, with sizes in the ratio of $1:5:10$. To form imbalanced constraint sets, a single "IMB cluster," corresponding to the first row in each heatmap, is selected to receive additional negative constraints connecting it to other clusters. The heatmap intensity at entry $(k,k')$ indicates the fraction of constraints connecting ground-truth clusters $k$ and $k'$, reflecting how {\tt IMB1} and {\tt IMB2} become increasingly dominated by the IMB cluster’s negative constraints.}
    \label{fig:imb_distributions}
\end{figure}

\paragraph{Imbalanced constraint set generation.}
For each imbalanced-constraint trial, we generate a group of three sets, {\tt IMB0}, {\tt IMB1} and {\tt IMB2}, to evaluate performance under progressively skewed constraint distributions. We fix a size ratio
$|\mathtt{IMB0}| : |\mathtt{IMB1}| : |\mathtt{IMB2}| = 1 : 5 : 10,$
and designate the first ground-truth class in each dataset as the “IMB cluster,” from which additional negative constraints are predominantly sampled. Concretely, we form {\tt IMB0} of size
$|{\tt IMB0}|$
by randomly sampling pairs of training data, labeling them positive if both samples lie in the same ground-truth cluster or negative otherwise (the same procedure as in Fig.~\ref{fig:constraint_heatmaps}). We then obtain {\tt IMB1} by adding 
$\bigl(|\mathtt{IMB1}| - |\mathtt{IMB0}|\bigr)$ 
extra negative constraints linking the IMB cluster to other clusters, and further enlarge {\tt IMB1} to {\tt IMB2} by appending $\bigl(|\mathtt{IMB2}| - |\mathtt{IMB1}|\bigr)$  similarly imbalanced constraints. As a result, {\tt IMB0} $\subset$ {\tt IMB1} $\subset$ {\tt IMB2}, ensuring that any performance decline from {\tt IMB0} to {\tt IMB1} or {\tt IMB2} cannot be attributed to removing earlier constraints.

Since we select the same IMB cluster and preserve the above ratio for each dataset, every group of three sets exhibits a consistent inter-cluster distribution. Fig.~\ref{fig:imb_distributions} illustrates this for the eight datasets, with each row corresponding to one dataset and the three columns showing {\tt IMB0}, {\tt IMB1}, {\tt IMB2} as heatmaps of size $K\times K$; the highlighted area in each {\tt IMB1} and {\tt IMB2} heatmap involves the IMB cluster. We can observe that {\tt IMB1} and {\tt IMB2} gradually become dominated by negative constraints involving the IMB cluster. In Fig.~\ref{fig:imb_10K_samll}, we report empirical results for $\bigl(|\mathtt{IMB0}|, |\mathtt{IMB1}|, |\mathtt{IMB2}|\bigr) = (10\mathrm{k}, 50\mathrm{k}, 100\mathrm{k}),$ while additional experiments with varying constraint set sizes can be seen in \cref{sec:IMB_appendix}. These additional constraint sets are generated using the same methodology and exhibit similar cluster-wise constraint distributions.

\paragraph{Cluster number inference.}
To evaluate our PCA-based cluster-number inference method (\cref{alg:inferK}), we apply standard mean-centered PCA \cite{PCA} to the $\ell_2$-normalized embeddings of instances involved in negative pairs. 
For comparison with alternative $K$-inference methods and to assess the applicability of different deep constraint embeddings, we adopt two alternative strategies, specifically “K-means + Silhouette Coefficient” and “Agglomerative Clustering + $K$-cluster lifetime”, which are applied to both the SpherePair and AutoEmbedder representations:

\begin{itemize}
    \item \textbf{K-means + Silhouette Coefficient (SC).}
    For each candidate $K'$, we run K-means clustering 5 times with random initializations and compute the average SC score on a random subset of 5,000 instances. Specifically, for each instance $j$, we calculate the average intra-cluster distance $d_j^{\text{intra}}$ between its embedding $\boldsymbol{z}_j$ (for AutoEmbedder) or $\text{Norm}(\boldsymbol{z}_j)$ (for SpherePair) and the embeddings of all other instances from its assigned cluster $\mathcal{S}_k$. We also determine $d_j^{\text{inter}}$, the minimum average distance between $\boldsymbol{z}_j$ (or $\text{Norm}(\boldsymbol{z}_j)$) and the embeddings of instances from any other assigned cluster $\mathcal{S}_{k'}$, where $k' \neq k$. The SC score $s_j$ for instance $j$ is then computed as $s_j = \left(d_j^{\text{inter}} - d_j^{\text{intra}}\right) \big/ \max\bigl(d_j^{\text{intra}},\,d_j^{\text{inter}}\bigr)$. The overall SC score is the mean of $s_j$ across all sampled instances. A higher SC indicates better-defined clusters with greater separation and lower intra-cluster dispersion. We select the number of clusters $K'$ as $K^*$ that corresonds to the maximal SC score.
   
    \item \textbf{Agglomerative clustering + $K$-cluster lifetime.}
 We employ agglomerative clustering with Ward linkage \cite{everitt2011cluster}, which iteratively merges the closest clusters while recording the corresponding linkage distances. For each candidate number of clusters, $K'$, let $d_{K'}$ denote the linkage distance when the dataset is partitioned into $K'$ clusters.
The \textit{$K$-cluster lifetime} at level $K'$ is defined as $\Delta d_{K'} = d_{K'} - d_{K'+1}$.
A larger $\Delta d_{K'}$ indicates a more significant increase in linkage distance, or equivalently, a longer $K$-cluster lifetime. This suggests that the $K'$-cluster partition is more stable and well-separated. To determine the optimal number of clusters, $K^*$, we select  $K^*=\text{argmax}_{K'} (\Delta d_{K'})$. In \cref{fig:SpherePair_lifetime,fig:AutoEmbedder_lifetime}, each $\Delta d_{K'}$ is normalized using $d_2$ for readability, resulting in ${\Delta \hat{d}}_{K'} = \frac{\Delta d_{K'}}{d_2}$.   
\end{itemize}

\subsection{Implementation}
\label{sec:implementation_appendix}

We provide the implementation details\footnote{Our source code is available on GitHub: \url{https://github.com/spherepaircc/SpherePairCC/tree/main}} for our experiments.

\paragraph{Software and hardware.}
We implement all methods (except DCGMM\footnote{We use the authors' implementation for DCGMM: \url{https://github.com/lauramanduchi/DC-GMM}}) in PyTorch 1.5.1 \footnote{PyTorch 1.5.1: \url{https://github.com/pytorch/pytorch/releases/tag/v1.5.1}} with Python 3.7. For SpherePair and AutoEmbedder, we use scikit-learn’s K-means implementation\footnote{Scikit-learn webpage: \url{https://scikit-learn.org/0.19/documentation.html}} and fastcluster's efficient hierarchical clustering implementation\footnote{fastcluster library: \url{https://pypi.org/project/fastcluster/}} for clustering. Experiments are conducted on Tesla V100 GPU with 16 GB of memory.

\paragraph{Data preprocessing.}
For Reuters and RCV1-10 subsets, we directly use the preprocessed tf-idf vectors. For all other datasets, we convert images into feature vectors to facilitate fair comparisons using fully connected networks. Specifically, for the 28$\times$28 grayscale images in MNIST and FashionMNIST, we reshape each image into a 784-dimensional vector, mirroring the methods of \cite{SDEC,CIDEC1,DCGMM}. For color images in CIFAR-10, CIFAR-100-20, STL-10, and ImageNet-10, we adopt the unsupervised feature extraction strategy proposed in \cite{ContrastiveClustering}; in particular, we train a ResNet-34 \cite{resnet} model for 1,000 epochs and utilize the resulting 512-dimensional latent representations. This preprocessing is consistent with the method employed in \cite{VolMaxDCC}. This vectorization step enables us to apply the same standard fully connected architectures across all datasets, ensuring consistency and fairness in evaluating the baselines. 

\paragraph{Network architectures and pretraining.}
For all methods except VanillaDCC and VolMaxDCC, we employ a fully connected encoder with hidden layers of size 500--500--2000 (following \cite{SDEC,CIDEC1,DCGMM}), using an embedding dimension of $D=20$ for CIFAR-100-20 and $D=10$ for all other datasets, unless stated otherwise.
Notably, while the original AutoEmbedder~\cite{AutoEmbedder} utilizes a pre-trained MobileNet-based CNN, we adapt it to use our standardized fully connected network to ensure fair comparison across all models. In contrast, VanillaDCC and VolMaxDCC~\cite{VolMaxDCC} adopt a distinct architecture comprising two hidden layers of size 512--512 followed by a classification layer corresponding to the number of clusters $K$, as recommended in \cite{VolMaxDCC}.

Except for the end-to-end VanillaDCC and VolMaxDCC, all other methods undergo unsupervised pretraining on training sets. Specifically, for SpherePair, SDEC, CIDEC, and AutoEmbedder, we utilize a two-stage stacked denoising autoencoder (SDAE) pretraining approach \cite{hinton2006reducing,vincent2008denoising}, consistent with the works \cite{SDEC,CIDEC1}:
\begin{enumerate}
    \item[(i)] \emph{Layer-wise Pretraining}: Each hidden layer is individually pretrained as a single-layer denoising autoencoder for 300 epochs using 20\% random masking noise and MSE loss. During this phase, the output of each encoder serves as the input to the subsequent layer, progressively refining the weights of each layer.
    \item[(ii)] \emph{End-to-End Fine-Tuning}: After completing layer-wise pretraining, the entire network is jointly optimized for an additional 500 epochs. This phase continues to apply 20\% masking noise to the inputs.
\end{enumerate}
A key distinction for SpherePair during pretraining is the normalization of latent embeddings before decoding, as specified in Eq.~\ref{eq:x_j_hat}. This ensures that the pre-trained autoencoder retains angular information critical for our clustering objectives. For DCGMM~\cite{DCGMM}, we follow the authors' setting by pretraining a variational autoencoder (VAE) for 10 epochs, aligning with their implementation strategy.

\paragraph{Hyperparameters and optimization.}
We provide detailed hyperparameter settings and optimization configurations for our SpherePair method and all baseline models.

\begin{itemize}
    \item \textbf{SpherePair (Ours).} We fix the negative-zone factor $\omega$ at 2 according to our theoretical analysis, and set the reconstruction loss weighting parameter $\lambda=0.02$ in all experiments unless varied for sensitivity analysis.
    For optimization, we employ the standard Adam optimizer with a learning rate of 0.001. The constraint mini-batch size is set to $|\mathcal{B}_c| = 256$, and consequently, the instance mini-batch size is determined by $|\mathcal{B}_x| = {|\mathcal{X}|} / {\frac{|\mathcal{C}|}{|\mathcal{B}_c|}}$, where ${|\mathcal{X}|}$ represents the dataset size and $|\mathcal{C}|$ denotes the constraint set size. Training is conducted for a maximum of 300 epochs. An early stopping criterion is applied after the first 100 epochs, terminating training if the relative change in loss $\mathcal{L}$ (Eq.~\ref{eq:L}) remains less than 0.1 for 5 consecutive epochs. 
    For clustering on the learned spherical representations $\mathcal{Z}_{\text{sphere}}$, we utilize either (i) the K-means algorithm with 20 random initializations or (ii) hierarchical clustering using the Ward linkage method. 
    For cluster-number inference, we set the tail ratio $\rho=0.05$ when computing the tail-averaged minimal inter-cluster angle $\overline{\delta}_d$.

    \item \textbf{VanillaDCC.} VanillaDCC is implemented straightforwardly by optimizing $\mathcal{L}_{\text{MCL}}$ using the standard Adam optimizer with a learning rate of 0.001. The batch size is set to 256. Training is conducted for a maximum of 300 epochs, with early stopping triggered if the relative change in the soft cluster assignments for training samples falls below 0.001 over 2 consecutive epochs. This early stopping strategy is widely adopted in end-to-end deep clustering \cite{DEC,IDEC} and deep constrained clustering \cite{SDEC,CIDEC1}
    
    \item \textbf{VolMaxDCC.} Following the VolMaxDCC paper\footnote{While we strictly follow the authors' optimal hyperparameter search strategy, which uses ground-truth class label information in the validation data, such information is typically unavailable in constrained clustering tasks.} \cite{VolMaxDCC}, we parameterize the optimization variable $B$ such that each element $B_{ij} = \frac{1}{1 + \exp(-B'_{ij})}$, where $B'_{ij}$ is a trainable parameter initialized to 1 if $i = j$ and to -1 otherwise. We utilize the SGD optimizer with learning rates of 0.5 for the network parameters and 0.1 for $B'$, respectively, and set the batch size to 128. The trade-off factor (geometric regularization weight) $\lambda$ is selected by searching over the range \{0, $10^{-1}$, $10^{-2}$, $10^{-3}$, $10^{-4}$, $10^{-5}$\} based on the model's best accuracy on the validation set. The optimal $\lambda$ values identified for the datasets CIFAR-100-20, CIFAR-10, FMNIST, ImageNet-10, MNIST, Reuters, STL-10, and RCV1-10 are $10^{-4}$, $10^{-4}$, $10^{-4}$, $10^{-5}$, $10^{-2}$, $10^{-4}$, $10^{-2}$, and $10^{-4}$, respectively. 
    
    \item \textbf{CIDEC.} We follow the authors' recommendations by setting $\lambda_1 = 1$ to balance the clustering and reconstruction losses and $\lambda_2 = 0.1$ to weight the contributions of positive constraints within the MCL-based loss. The K-means algorithm is executed 20 times to initialize the $K$ cluster anchors. Optimization is performed using the standard Adam optimizer with a learning rate of 0.001 and a batch size of 256. Training proceeds for up to 300 epochs, with early stopping invoked if the soft cluster assignments exhibit a relative change of less than 0.001 over consecutive epochs.
    
    \item \textbf{DCGMM.} Utilizing the authors' implementation, we set the constraint weights $|W_{ij}| = 10,\!000$ and adhere to their optimization configurations. However, we pretrain a variational autoencoder (VAE) for 10 epochs exclusively on the instances in the training set, rather than using all instances from both the training and test sets.\footnote{Based on the authors' source code, we observed that their experiments involved pretraining the VAE on all instances from both the training and test sets. This constitutes a transductive setting, which is unsuitable for scenarios requiring inductive learning, as is the case in all our experiments.}
    
    \item \textbf{SDEC.} We align with the authors' recommendations by setting the constraint loss weight $\lambda = 10^{-5}$. The K-means algorithm is executed 20 times to initialize the $K$ cluster anchors used for unsupervised clustering. Optimization is carried out using the SGD optimizer with a learning rate of 0.01 and a batch size of 256. Training continues for a maximum of 300 epochs, with early stopping triggered if the soft cluster assignments change by less than 0.001 over consecutive epochs.
    
    \item \textbf{AutoEmbedder.} We implement AutoEmbedder using the same fully connected encoder architecture as SpherePair, ensuring consistency across models. The optimization settings mirror those of SpherePair: an Adam optimizer with a learning rate of 0.001 and a batch size of 256. Training is conducted for up to 300 epochs, with an early stopping criterion applied after the first 100 epochs, terminating training if the relative change in loss remains below 0.1 for 5 consecutive epochs. Unlike the original AutoEmbedder~\cite{AutoEmbedder}, which is based on a pre-trained CNN network with a well-structured embedding space, our fully connected implementation requires modifications to the loss function. Specifically, the MSE loss $\mathcal{L}_{\text{MSE}}$ can lead to scenarios where the embedding distance for positive pairs exceeds the margin $\alpha$, causing gradient issues. To address this, we introduce separate margins $\alpha_1$ and $\alpha_2$ for negative and positive constraints, respectively. We search for $\alpha_1$ within \{1, 10, 50, 100, 500, 1000, 5000\} and $\alpha_2$ within \{100, 1000, 10000\}, performing hyperparameter tuning based on validation set performance, similar to the approach used for VolMaxDCC. 
    The optimal $\alpha_1$ values identified for the datasets CIFAR-100-20, CIFAR-10, FMNIST, ImageNet-10, MNIST, Reuters, STL-10, and RCV1-10 are 500, 500, 50, 10, 100, 500, 50, and 10, respectively, while the optimal $\alpha_2$ is consistently 10,\!000 across all datasets.
    This outcome is expected, as the margin for positive constraints does not contribute to the Euclidean clustering objective, and smaller margins render positive constraints ineffective.

\end{itemize}

\section{Additional experimental results}
\label{sec:result_appendix}

In this appendix, we present supplementary experimental results, building upon the experimental settings detailed in \cref{sec:setting_appendix}. Additionally, we provide further insights and findings related to our SpherePair approach.

\subsection{Comparison of hierarchical clustering results}
\label{sec:hierarchical}

We present additional results from our comparative study.
In \cref{table:SpherePair_VS_baselines} presented in the main text, we reported the primary clustering analysis results for two anchor-free deep constraint embedding models, AutoEmbedder and our SpherePair, using K-means applied to their learned representations. Here, we extend the analysis by presenting results for both models using Ward's agglomerative hierarchical clustering, as shown in \cref{table:hier}.

\begin{table}[h]
    \caption{Comparative performance (\%) of ACC, NMI, and ARI across multiple datasets for AutoEmbedder (AE) and SpherePair (Ours) models using 1k, 5k, and 10k constraints. The results are derived from the hierarchical clustering analysis applied to their learned representations. Consistent with the notation used in \cref{table:SpherePair_VS_baselines}, \textcolor{blue}{blue} and black indicate training and test performance, respectively. Better results are highlighted in \textbf{bold}.}
    \label{table:hier}
    \begin{center}
    \begin{scriptsize}
    \renewcommand{\arraystretch}{1.05}
    \setlength{\tabcolsep}{3.1pt} 
    \setlength{\aboverulesep}{1pt}
    \setlength{\belowrulesep}{1pt}
    \setlength{\extrarowheight}{0pt}
    
    \begin{tabular}{ccccc@{\hskip 7pt}>{\hskip 7pt}ccc<{\hskip 7pt}@{\hskip 7pt}ccc}
    \toprule
      \multicolumn{1}{l}{} & & \multicolumn{3}{c}{1k} & \multicolumn{3}{c}{5k} & \multicolumn{3}{c}{10k} \\
    \cline{3-11}\noalign{\vskip 2pt}
      \multicolumn{1}{l}{} & & ACC & NMI & ARI & ACC & NMI & ARI & ACC & NMI & ARI \\
    \midrule

    \multirow{2}{*}{CIFAR100-20} 
    & AE 
    & \textcolor{blue}{19.7}, 19.2 
    & \textcolor{blue}{21.4}, 20.7 
    & \textcolor{blue}{5.9}, 5.8
    & \textcolor{blue}{10.5}, 10.8
    & \textcolor{blue}{7.9}, 8.5
    & \textcolor{blue}{2.1}, 2.2
    & \textcolor{blue}{31.2}, 31.5 
    & \textcolor{blue}{37.3}, 38.5
    & \textcolor{blue}{21.2}, 22.1
    \\ 
    & Ours 
    & \textcolor{blue}{\textbf{45.9}}, \textbf{46.4}
    & \textcolor{blue}{\textbf{46.4}}, \textbf{46.3}
    & \textcolor{blue}{\textbf{29.4}}, \textbf{30.5}
    & \textcolor{blue}{\textbf{55.0}}, \textbf{54.9}
    & \textcolor{blue}{\textbf{50.9}}, \textbf{51.2}
    & \textcolor{blue}{\textbf{37.4}}, \textbf{38.7}
    & \textcolor{blue}{\textbf{61.6}}, \textbf{62.6} 
    & \textcolor{blue}{\textbf{54.8}}, \textbf{54.9}
    & \textcolor{blue}{\textbf{42.6}}, \textbf{44.1}
    \\
    \cline{1-11}\noalign{\vskip 2pt}
    
    \multirow{2}{*}{CIFAR10} 
    & AE 
    & \textcolor{blue}{45.8},  44.2 
    & \textcolor{blue}{51.3},  50.1 
    & \textcolor{blue}{29.4},  28.0 
    & \textcolor{blue}{82.1},  83.0 
    & \textcolor{blue}{77.4},  78.3 
    & \textcolor{blue}{72.3},  73.9 
    & \textcolor{blue}{84.8},  85.3 
    & \textcolor{blue}{79.4},  80.0 
    & \textcolor{blue}{75.2},  76.5 
    \\ 
    & Ours 
    & \textcolor{blue}{\textbf{83.7}},  \textbf{84.3} 
    & \textcolor{blue}{\textbf{77.0}},  \textbf{77.1} 
    & \textcolor{blue}{\textbf{72.3}},  \textbf{73.0}
    & \textcolor{blue}{\textbf{89.5}},  \textbf{89.4} 
    & \textcolor{blue}{\textbf{81.5}},  \textbf{81.1}
    & \textcolor{blue}{\textbf{79.8}},  \textbf{79.6}
    & \textcolor{blue}{\textbf{90.5}},  \textbf{90.2}
    & \textcolor{blue}{\textbf{82.4}},  \textbf{82.0}
    & \textcolor{blue}{\textbf{81.3}},  \textbf{80.8}
    \\
    \cline{1-11}\noalign{\vskip 2pt}

    \multirow{2}{*}{FMNIST}
    & AE 
    & \textcolor{blue}{40.7},  40.2 
    & \textcolor{blue}{42.2},  40.8 
    & \textcolor{blue}{25.7},  25.0 
    & \textcolor{blue}{60.5},  61.2 
    & \textcolor{blue}{59.5},  60.0 
    & \textcolor{blue}{46.8},  48.2 
    & \textcolor{blue}{66.6},  67.1 
    & \textcolor{blue}{65.0},  64.6 
    & \textcolor{blue}{52.4},  53.6 
    \\
    & Ours
    & \textcolor{blue}{\textbf{66.5}},  \textbf{67.8} 
    & \textcolor{blue}{\textbf{64.3}},  \textbf{63.1} 
    & \textcolor{blue}{\textbf{51.2}},  \textbf{52.3} 
    & \textcolor{blue}{\textbf{79.4}},  \textbf{79.2} 
    & \textcolor{blue}{\textbf{71.8}},  \textbf{70.8} 
    & \textcolor{blue}{\textbf{65.4}},  \textbf{65.2} 
    & \textcolor{blue}{\textbf{84.4}},  \textbf{83.7} 
    & \textcolor{blue}{\textbf{75.3}},  \textbf{74.2} 
    & \textcolor{blue}{\textbf{71.5}},  \textbf{70.2} 
    \\
    \cline{1-11}\noalign{\vskip 2pt}

    \multirow{2}{*}{ImageNet10}
    & AE 
    & \textcolor{blue}{63.3},  61.2 
    & \textcolor{blue}{62.5},  58.0 
    & \textcolor{blue}{47.0},  42.6 
    & \textcolor{blue}{96.2},  \textbf{96.4} 
    & \textcolor{blue}{91.4},  \textbf{91.9} 
    & \textcolor{blue}{91.7},  92.1 
    & \textcolor{blue}{96.6},  96.5 
    & \textcolor{blue}{92.0},  \textbf{92.1} 
    & \textcolor{blue}{92.7},  92.4 
    \\
    & Ours
    & \textcolor{blue}{\textbf{95.9}},  \textbf{96.1} 
    & \textcolor{blue}{\textbf{91.1}},  \textbf{91.4} 
    & \textcolor{blue}{\textbf{91.3}},  \textbf{91.6} 
    & \textcolor{blue}{\textbf{96.8}},  \textbf{96.4} 
    & \textcolor{blue}{\textbf{92.3}},  91.7
    & \textcolor{blue}{\textbf{93.2}},  \textbf{92.3} 
    & \textcolor{blue}{\textbf{97.2}},  \textbf{96.6} 
    & \textcolor{blue}{\textbf{93.1}},  \textbf{92.1} 
    & \textcolor{blue}{\textbf{94.0}},  \textbf{92.6} 
    \\
    \cline{1-11}\noalign{\vskip 2pt}

    \multirow{2}{*}{MNIST}
    & AE 
    & \textcolor{blue}{44.2},  42.6 
    & \textcolor{blue}{41.3},  38.3 
    & \textcolor{blue}{27.1},  25.1 
    & \textcolor{blue}{60.4},  61.1 
    & \textcolor{blue}{57.5},  58.6 
    & \textcolor{blue}{47.1},  49.1 
    & \textcolor{blue}{87.3},  89.4 
    & \textcolor{blue}{79.9},  83.0 
    & \textcolor{blue}{77.9},  82.0 
    \\
    & Ours 
    & \textcolor{blue}{\textbf{95.4}},  \textbf{92.3} 
    & \textcolor{blue}{\textbf{89.5}},  \textbf{83.4} 
    & \textcolor{blue}{\textbf{90.3}},  \textbf{83.8} 
    & \textcolor{blue}{\textbf{96.6}},  \textbf{96.1} 
    & \textcolor{blue}{\textbf{91.5}},  \textbf{90.2} 
    & \textcolor{blue}{\textbf{92.7}},  \textbf{91.5} 
    & \textcolor{blue}{\textbf{97.0}},  \textbf{96.9} 
    & \textcolor{blue}{\textbf{92.0}},  \textbf{91.8} 
    & \textcolor{blue}{\textbf{93.6}},  \textbf{93.2} 
    \\
    \cline{1-11}\noalign{\vskip 2pt}

    \multirow{2}{*}{REUTERS}
    & AE 
    & \textcolor{blue}{55.5},  55.1 
    & \textcolor{blue}{24.0},  25.8 
    & \textcolor{blue}{19.3},  19.4 
    & \textcolor{blue}{87.6},  88.7 
    & \textcolor{blue}{67.6},  69.1 
    & \textcolor{blue}{76.4},  78.1 
    & \textcolor{blue}{93.0},  91.9 
    & \textcolor{blue}{79.0},  76.2 
    & \textcolor{blue}{86.2},  83.6 
    \\
    & Ours 
    & \textcolor{blue}{\textbf{91.6}},  \textbf{92.3} 
    & \textcolor{blue}{\textbf{73.1}},  \textbf{75.1} 
    & \textcolor{blue}{\textbf{80.4}},  \textbf{82.4} 
    & \textcolor{blue}{\textbf{96.0}},  \textbf{94.8} 
    & \textcolor{blue}{\textbf{84.4}},  \textbf{81.0} 
    & \textcolor{blue}{\textbf{90.5}},  \textbf{87.4} 
    & \textcolor{blue}{\textbf{97.6}},  \textbf{95.4} 
    & \textcolor{blue}{\textbf{89.7}},  \textbf{82.4} 
    & \textcolor{blue}{\textbf{94.6}},  \textbf{88.6} 
    \\
    \cline{1-11}\noalign{\vskip 2pt}

    \multirow{2}{*}{STL10}
    & AE 
    & \textcolor{blue}{59.3},  59.4 
    & \textcolor{blue}{56.7},  55.9 
    & \textcolor{blue}{41.1},  40.7 
    & \textcolor{blue}{81.8},  83.1 
    & \textcolor{blue}{76.0},  76.8 
    & \textcolor{blue}{68.7},  71.2 
    & \textcolor{blue}{89.6},  89.7 
    & \textcolor{blue}{81.6},  81.3 
    & \textcolor{blue}{79.0},  79.4 
    \\
    & Ours 
    & \textcolor{blue}{\textbf{86.1}},  \textbf{87.4} 
    & \textcolor{blue}{\textbf{77.2}},  \textbf{78.2} 
    & \textcolor{blue}{\textbf{73.3}},  \textbf{75.6} 
    & \textcolor{blue}{\textbf{90.9}},  \textbf{89.9} 
    & \textcolor{blue}{\textbf{82.5}},  \textbf{81.2} 
    & \textcolor{blue}{\textbf{81.6}},  \textbf{79.8} 
    & \textcolor{blue}{\textbf{92.9}},  \textbf{90.7} 
    & \textcolor{blue}{\textbf{85.3}},  \textbf{82.2} 
    & \textcolor{blue}{\textbf{85.2}},  \textbf{81.1} 
    \\
    \cline{1-11}\noalign{\vskip 2pt}

    \multirow{2}{*}{RCV1-10} 
    & AE 
    & \textcolor{blue}{31.7}, 31.6
    & \textcolor{blue}{9.2}, 8.6
    & \textcolor{blue}{6.5}, 6.7
    & \textcolor{blue}{50.9}, 51.9
    & \textcolor{blue}{32.3}, 33.7
    & \textcolor{blue}{35.6}, 38.1
    & \textcolor{blue}{77.5}, 78.6 
    & \textcolor{blue}{55.5}, 57.4
    & \textcolor{blue}{67.2}, 70.5
    \\ 
    & Ours 
    & \textcolor{blue}{\textbf{66.9}}, \textbf{67.8}
    & \textcolor{blue}{\textbf{60.2}}, \textbf{61.8}
    & \textcolor{blue}{\textbf{58.4}}, \textbf{59.8}
    & \textcolor{blue}{\textbf{89.3}}, \textbf{89.6}
    & \textcolor{blue}{\textbf{74.6}}, \textbf{74.4}
    & \textcolor{blue}{\textbf{84.0}}, \textbf{83.8}
    & \textcolor{blue}{\textbf{91.5}}, \textbf{91.5} 
    & \textcolor{blue}{\textbf{77.9}}, \textbf{77.4}
    & \textcolor{blue}{\textbf{87.0}}, \textbf{86.4}
    \\
    
    \bottomrule
    \end{tabular}
    \end{scriptsize}
    \end{center}
\end{table}

From \cref{table:hier}, we observe that SpherePair consistently outperforms AutoEmbedder across nearly all dataset-constraint-metric combinations. The only exception is a minor 0.2\% gap in test NMI when using 5k constraints from ImageNet10. When comparing the hierarchical clustering results of SpherePair in \cref{table:hier} with the K-means results reported in \cref{table:SpherePair_VS_baselines}, we find that SpherePair delivers nearly identical performance, with most deviations under 1\% and a maximum difference of less than 4\%. 
This consistency highlights the robustness of SpherePair’s embeddings across different clustering algorithms. In contrast, AutoEmbedder exhibits more pronounced performance fluctuations when switching from K-means to hierarchical clustering, with some cases showing gaps as large as 15\% (e.g., CIFAR-10 with 1k constraints). These variations suggest that AutoEmbedder struggles to generate well-structured cluster representations, making its downstream partitioning outcomes highly sensitive to the choice of clustering method.

These extended results underscore the versatility of SpherePair’s learned embeddings, which deliver stable and high-quality clustering results regardless of the clustering analysis algorithm used. This flexibility is particularly valuable in practice, as hierarchical clustering can reveal dendrogram structures that provide insights into domain-specific phenomena—something K-means and end-to-end anchor-based DCC methods cannot readily achieve. Consequently, users can confidently replace the clustering step in our framework with a more interpretable or domain-specific method, assured that SpherePair’s representations will continue to deliver strong performance.

\subsection{Imbalanced constraints}
\label{sec:IMB_appendix}

We present additional experimental results that analyze model behavior and visualize learned representations in latent embedding spaces under imbalanced constraint conditions.
The distribution of imbalanced constraints skews toward specific inter-cluster relationships due to experts’ greater familiarity with particular knowledge areas, which is common in real-world scenarios.

\subsubsection{Model behavior}
\label{sec:IMB_behavior}

\begin{figure}[t]
    \begin{center}
    \centerline{\includegraphics[width=\textwidth]{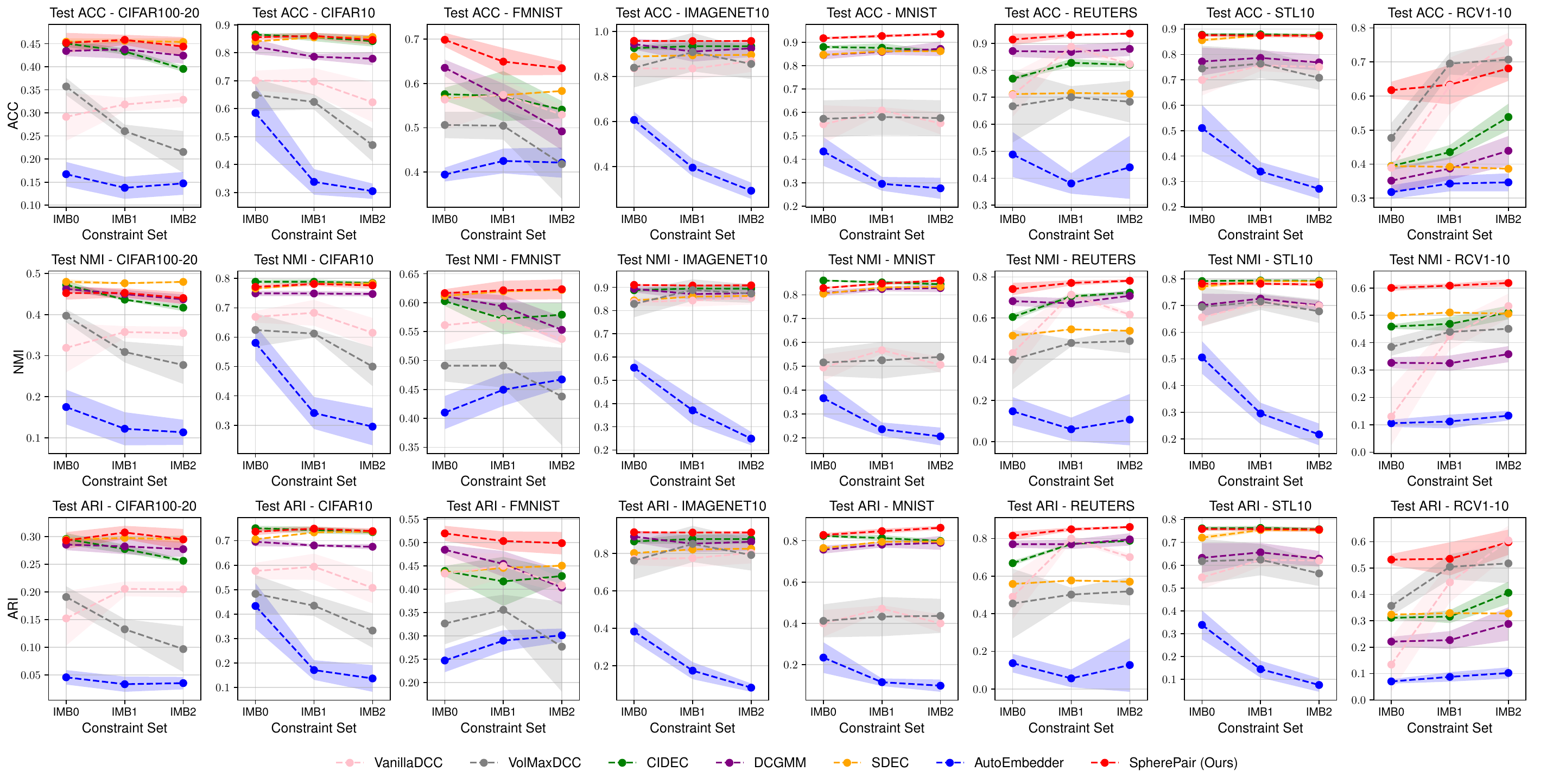}}
    \caption{Test ACC/NMI/ARI performance (mean$\pm$std over 5 runs) of all models across datasets under the imbalanced constraints setting where ($\mid${\tt IMB0}$\mid$, $\mid${\tt IMB1}$\mid$, $\mid${\tt IMB2}$\mid$) = (1k, 5k, 10k).}    
    \label{fig:imb_1k_big}
    \end{center}
    \vskip -0.2in
\end{figure}

\begin{figure}[t]
    \begin{center}
    \centerline{\includegraphics[width=\textwidth]{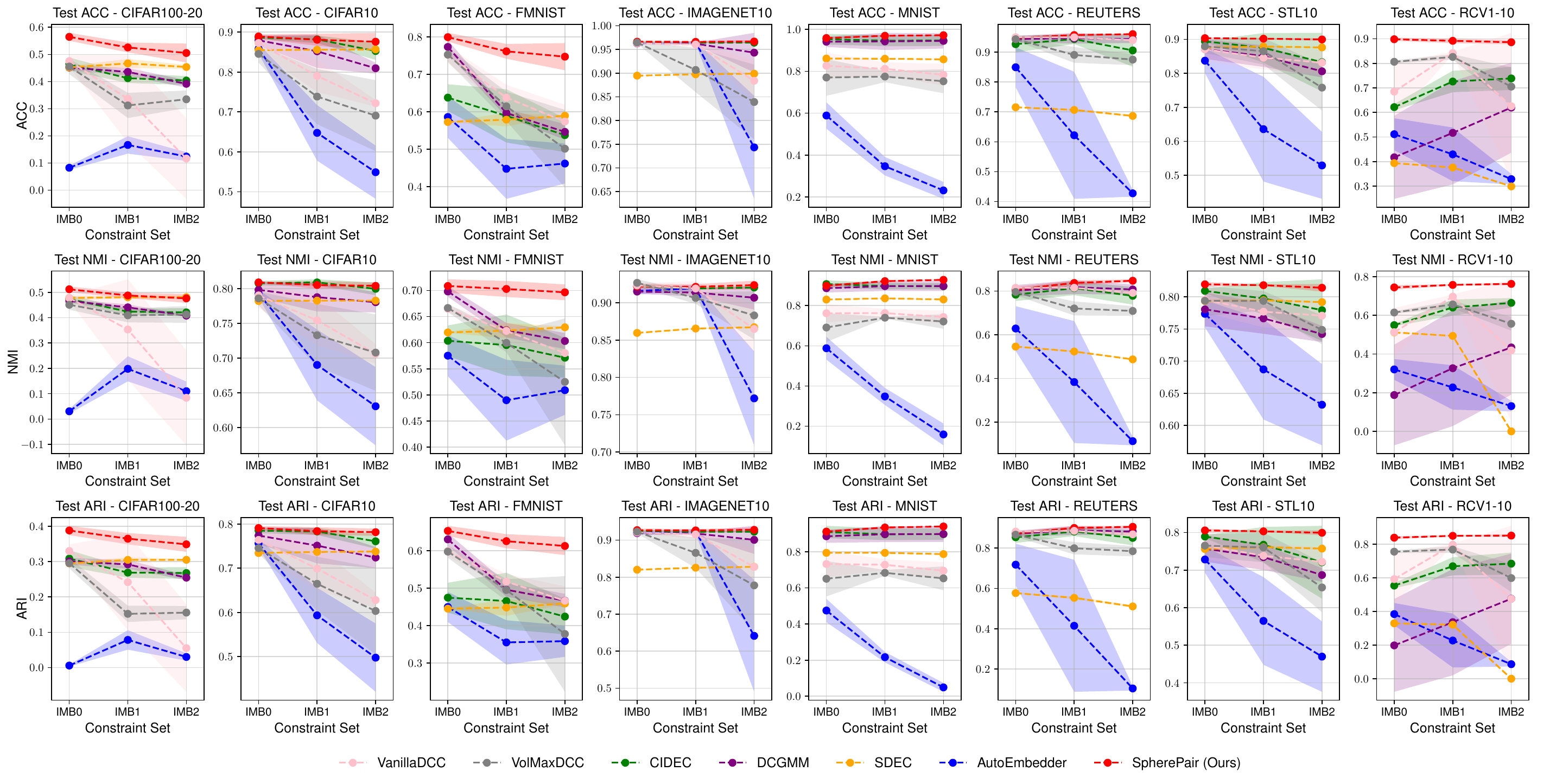}}
    \caption{Test ACC/NMI/ARI performance (mean$\pm$std over 5 runs) of all models across datasets under the imbalanced constraints setting where ($\mid${\tt IMB0}$\mid$, $\mid${\tt IMB1}$\mid$, $\mid${\tt IMB2}$\mid$) = (5k, 25k, 50k).}   
    \label{fig:imb_5k_big}
    \end{center}
    \vskip -0.2in
\end{figure}

\begin{figure}[t]
    \begin{center}
    \centerline{\includegraphics[width=\textwidth]{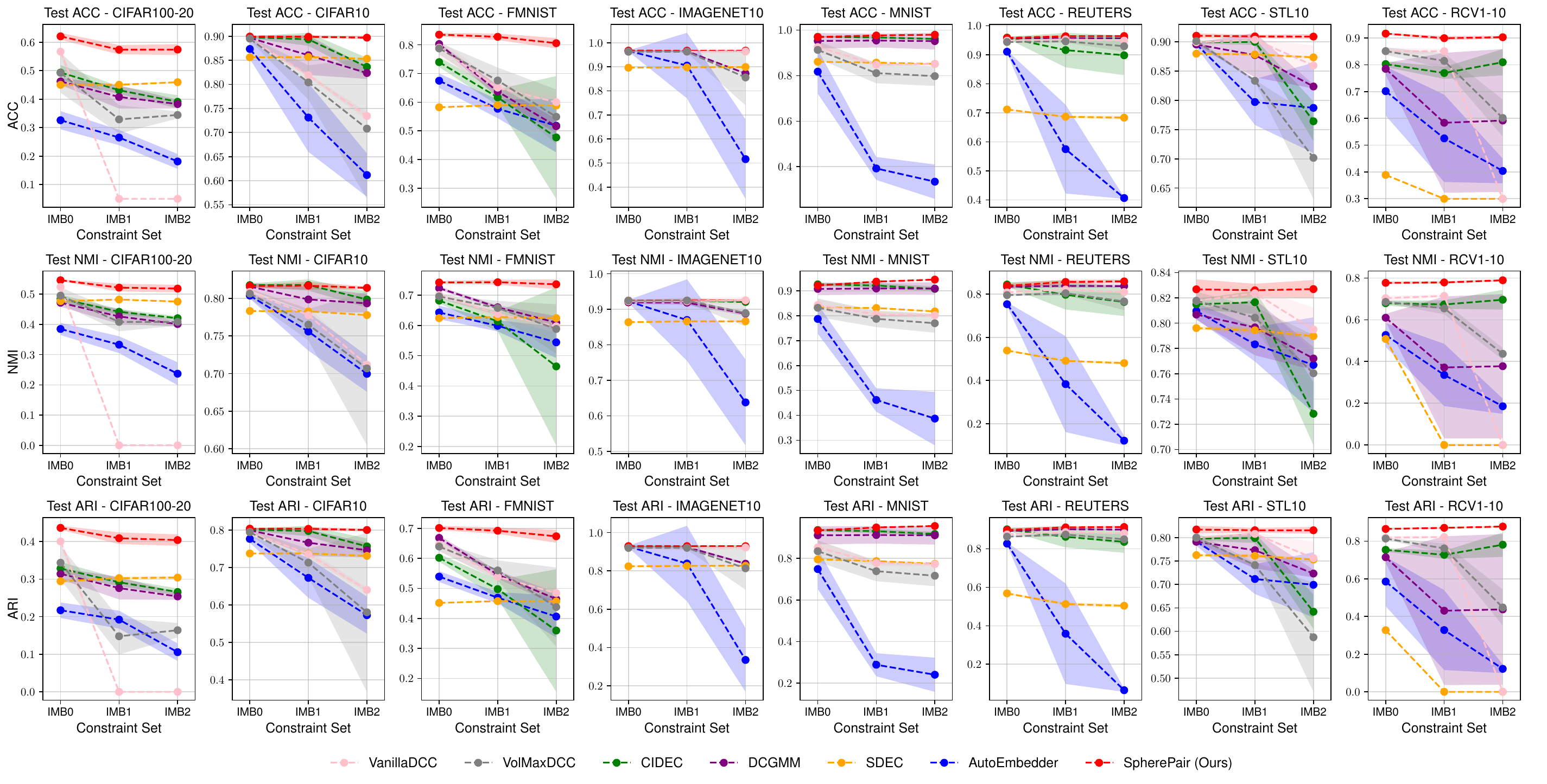}}
    \caption{Test ACC/NMI/ARI performance (mean$\pm$std over 5 runs) of all models across datasets under the imbalanced constraints setting where ($\mid${\tt IMB0}$\mid$, $\mid${\tt IMB1}$\mid$, $\mid${\tt IMB2}$\mid$) = (10k, 50k, 100k).}     
    \label{fig:imb_10k_big}
    \end{center}
    \vskip -0.4in
\end{figure}

To complement Fig.~\ref{fig:imb_10K_samll} in the main text, we provide additional results in \cref{fig:imb_1k_big,fig:imb_5k_big,fig:imb_10k_big}. Each figure illustrates how performance changes as the imbalance level increases from \texttt{IMB0} to \texttt{IMB1} and \texttt{IMB2}, following the generation procedure in \cref{sec:protocol_appendix}, where the nested relation \(\texttt{IMB0} \subset \texttt{IMB1} \subset \texttt{IMB2}\) rules out declines due to reduced constraints. The three figures correspond to different constraint set sizes, namely $(1\mathrm{k},5\mathrm{k},10\mathrm{k})$, $(5\mathrm{k},25\mathrm{k},50\mathrm{k})$, and $(10\mathrm{k},50\mathrm{k},100\mathrm{k})$. Each shows three rows of metrics (ACC, NMI, ARI) across eight datasets, offering a comprehensive view of model behavior under varying imbalance levels.

\paragraph{Baseline behavior.}
Most baseline models exhibit increasing performance degradation as the imbalance level rises from \texttt{IMB0} to \texttt{IMB2}, although the severity of this decline varies depending on the initial size of \(|\texttt{IMB0}|\). 

\begin{itemize}
    \item When \(|\texttt{IMB0}|\) is relatively small, the additional constraints from \texttt{IMB1} or \texttt{IMB2} can have a partially positive impact, mitigating some of the adverse effects of the skewed constraint distribution. For example, in \cref{fig:imb_1k_big}, most baselines on the STL10 dataset experience only limited degradation as the imbalance increases, and on RCV1-10 most baselines even exhibit a noticeable performance gain.
    \item In contrast, when the balanced set size \(|\texttt{IMB0}|\) is already large enough for near-saturation performance, introducing imbalanced constraints from \texttt{IMB1} or \texttt{IMB2} amplifies negative effects. As shown in \cref{fig:imb_10k_big}, most baselines suffer pronounced drops on both STL10 and RCV1-10 compared to the lower-constraint scenarios.
\end{itemize}

An exception is the SDEC model, which remains stable under varying levels of imbalance. This stability likely stems from its reliance on an unsupervised clustering objective, as shown in \cref{table:SpherePair_VS_baselines} presented in the main text, where increasing constraints also leads to minimal performance gains. As a result, SDEC is less susceptible to both the benefits and drawbacks of heavily imbalanced constraint sets, maintaining relatively steady behavior under \texttt{IMB} conditions.

\paragraph{SpherePair behavior.}
SpherePair remains robust across datasets and imbalance scenarios, consistently ranking among the top-performing methods. The only notable degradations occur on CIFAR-100-20 and FMNIST; however, SpherePair is always noticeably less affected than baselines and consistently outperforms them under nearly all settings. In particular, on FMNIST, increasing \(|\texttt{IMB0}|\) enables SpherePair to effectively exploit the richer constraint information to counteract the negative effects of imbalance (see \cref{fig:imb_10k_big}). This resilience can be attributed to its ability to respect and leverage non-dominant local pairwise relationships.

SpherePair's robustness highlights its practical advantage in real-world scenarios, where annotators are more likely to label familiar pairwise relationships while neglecting less familiar ones. This adaptability makes SpherePair particularly suited for imbalanced constraint settings.

\begin{figure}[htbp]
    \centering
    \centerline{\includegraphics[width=0.75\textwidth]{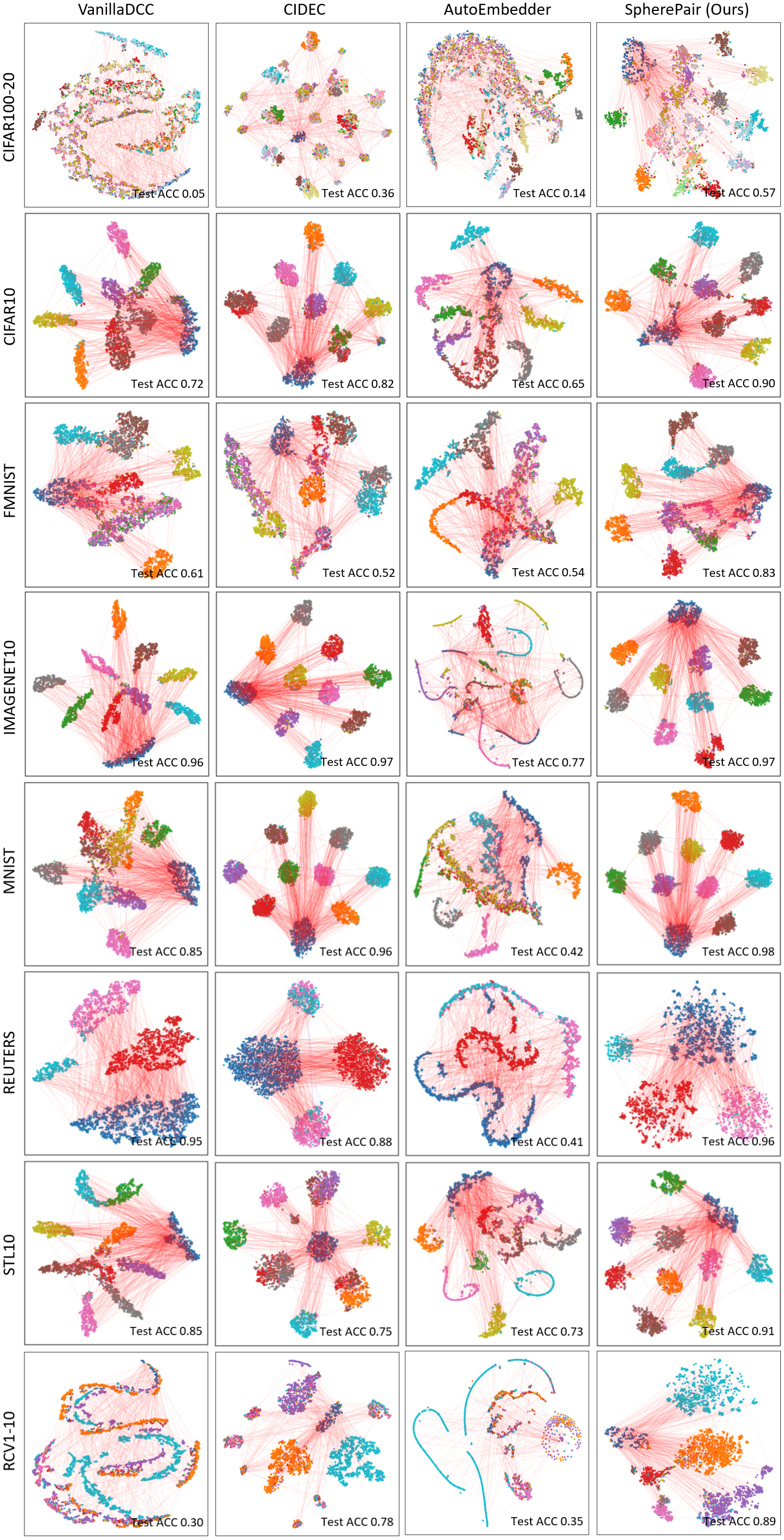}}
    \caption{t-SNE visualization of embeddings under imbalanced constraints (\(\mid\texttt{IMB2}\mid = 100\mathrm{k}\)). Each column corresponds to one model: (a)~VanillaDCC, (b)~CIDEC, (c)~AutoEmbedder, and (d)~SpherePair. We visualize the final hidden-layer output (VanillaDCC), latent embedding (CIDEC, AutoEmbedder), and normalized latent embedding (SpherePair). Different marker colors denote different ground-truth categories.}
    \label{fig:tSNE_merge}
\end{figure}

\subsubsection{Latent embedding visualization}
\label{sec:IMB_visualization}

To gain deeper insights into how a DCC model is influenced by imbalanced constraints, we visualize the learned representations in the latent embedding space under the \texttt{IMB2} configuration shown in \cref{fig:imb_10k_big} (where $|\texttt{IMB2}| = 100\mathrm{k}$). The visualization focuses on four representative models that explicitly learn latent embeddings: \textbf{VanillaDCC} (an end-to-end classification approach), \textbf{CIDEC} (an end-to-end autoencoder approach), \textbf{AutoEmbedder} (a deep constraint embedding method in Euclidean space), and our \textbf{SpherePair} (a deep constraint embedding method in angular space). 

For each model, the embeddings are visualized as follows: the output of the last hidden layer for VanillaDCC, the autoencoder’s latent space for CIDEC and AutoEmbedder, and the unit-normalized spherical embeddings for SpherePair. 
\cref{fig:tSNE_merge} show the t-SNE plots of the resulting representations across different datasets, leading to several key observations:

\paragraph{End-to-end methods.}
Both VanillaDCC and CIDEC exhibit a tendency to mismatch local similarities and global clustering decisions under imbalanced constraints. This often results in instances from different ground-truth categories being incorrectly embedded into tight, misassigned clusters.
These clusters indicate that imbalanced constraints cause the anchors to disproportionately emphasize dominant relationships, neglecting minority local constraints.

\paragraph{AutoEmbedder.}
As an anchor-free Euclidean embedding method, AutoEmbedder generates non-convex and less discriminative clusters, suggesting that pairwise learning in Euclidean space is particularly sensitive to imbalance. While AutoEmbedder occasionally preserves local groupings for specific categories, its clusters can be challenging to partition accurately. For instance, on the Reuters dataset, K-means applied to AutoEmbedder’s features yields an ACC of only $0.41$.

\paragraph{SpherePair.}
In contrast, SpherePair leverages the properties of angular space and its derived \textit{negative zone} to maintain a balance between respecting local relationships and forming sufficiently separable, convex clusters. Even under severe imbalance, SpherePair produces normalized embeddings that form compact, clearly discernible clusters, demonstrating its robustness in representation learning.

Overall, these visualizations highlight that under strong constraint imbalance, anchor-based end-to-end methods like VanillaDCC and CIDEC are prone to misclustering minority classes, while anchor-free Euclidean-based deep constraint embedding methods like AutoEmbedder struggle to form separable clusters. Our SpherePair, by capitalizing on angular distances, preserves coherent local structures and generates stable, well-defined clusters, even under skewed supervision.

\subsection{Unknown cluster number}
\label{sec:unknown_K}
We comprehensively evaluate our PCA-based cluster-number inference combined with SpherePair in terms of its effectiveness across different constraint levels, its comparison with alternative $K$-inference strategies, and the applicability of SpherePair relative to DCC baselines for $K$-inference.

\subsubsection{PCA-based $K$-inference under different constraint levels.}
\label{sec:unknwon_K_constraint_levels}
\begin{figure}[htbp]
    \centering
    \begin{subfigure}{\linewidth}
        \centering
        \includegraphics[width=\textwidth]{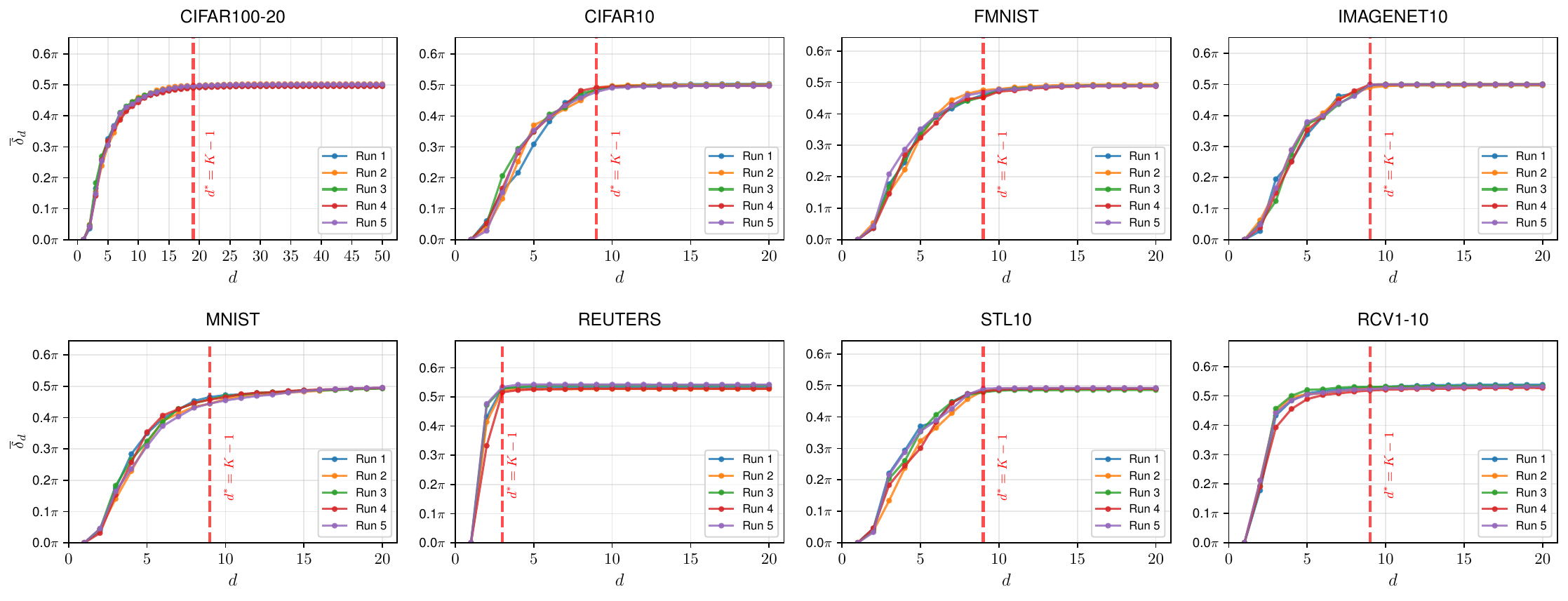}
        \caption{} \label{fig:K_merge_main:1k}
    \end{subfigure}
    \vskip 0.05in

    \begin{subfigure}{\linewidth}
        \centering
        \includegraphics[width=\textwidth]{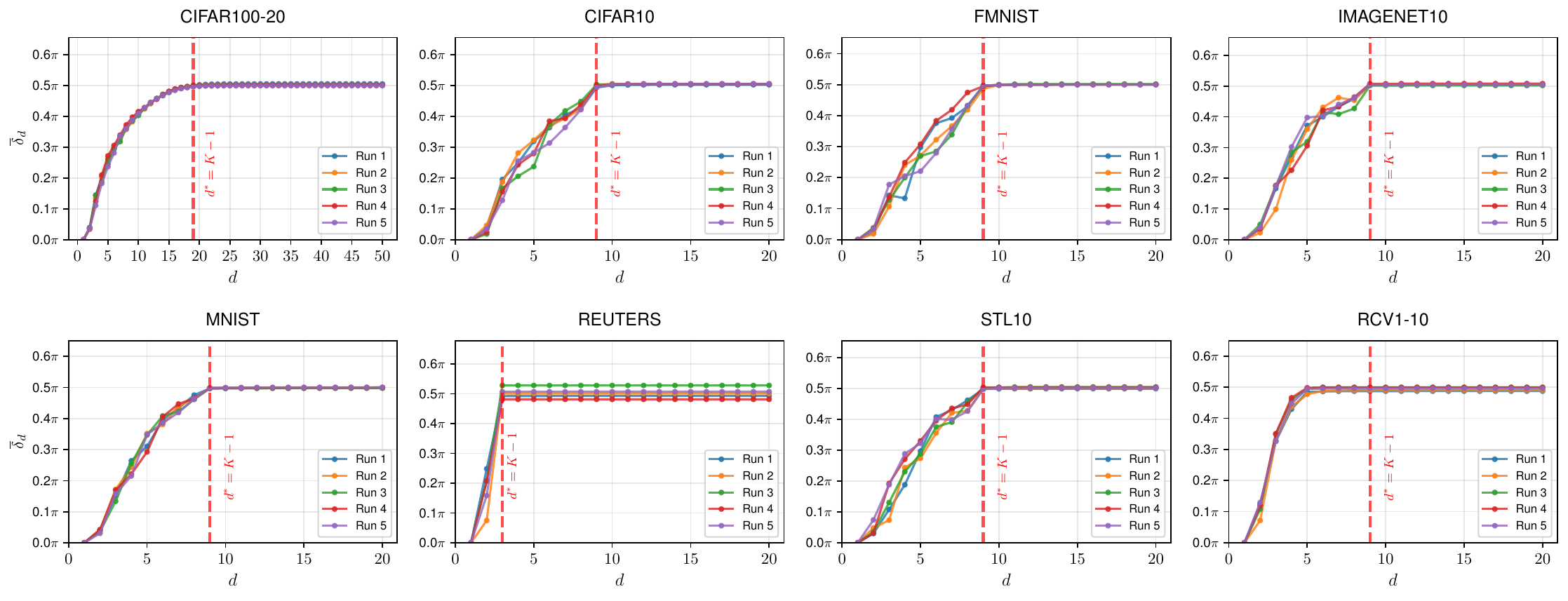}
        \caption{} \label{fig:K_merge_main:5k}
    \end{subfigure}
    \vskip 0.05in

    \begin{subfigure}{\linewidth}
        \centering
        \includegraphics[width=\textwidth]{K_merge_main_10k.pdf}
        \caption{} \label{fig:K_merge_main:10k}
    \end{subfigure}
    \vskip 0.05in

    \caption{Tail-averaged minimal inter-cluster angle $\overline{\delta}_d$ vs. PCA subspace dimension $d$, obtained from SpherePair embeddings learned with (a) 1k, (b) 5k, and (c) 10k constraints across five runs. The \textcolor{red}{red lines} indicate the ground-truth intrinsic dimensions $d^\ast = K{-}1$.}
    
    \label{fig:K_merge_main}
\end{figure}

We extend the evaluation of our $K$-inference from \cref{fig:K_merge_main_10k} to additional constraint levels (1k/5k/10k), as shown in \cref{fig:K_merge_main}. 
Across most datasets, 10k constraints yield clear plateau entries, while 5k constraints produce slightly less pronounced entries that remain sufficient for correct $K$ estimation. 
An exception is RCV1-10, where strong class imbalance poses inherent challenges and leads to inaccurate estimates across constraint settings. 
In the more limited 1k-constraint setting, the plateaus become much less sharp, particularly on CIFAR-100-20, MNIST, and FMNIST, resulting in more frequent inaccuracies. 
Nevertheless, under the 1k-constraint setting, our method still produces correct $K$ estimates on CIFAR-10, ImageNet-10, Reuters, and STL-10 in nearly all cases, with only two minor deviations on CIFAR-10 where the estimate differed from the ground truth by 1.

\begin{figure}[htbp]
    \centering
    \begin{subfigure}{\linewidth}
        \centering
        \includegraphics[width=\textwidth]{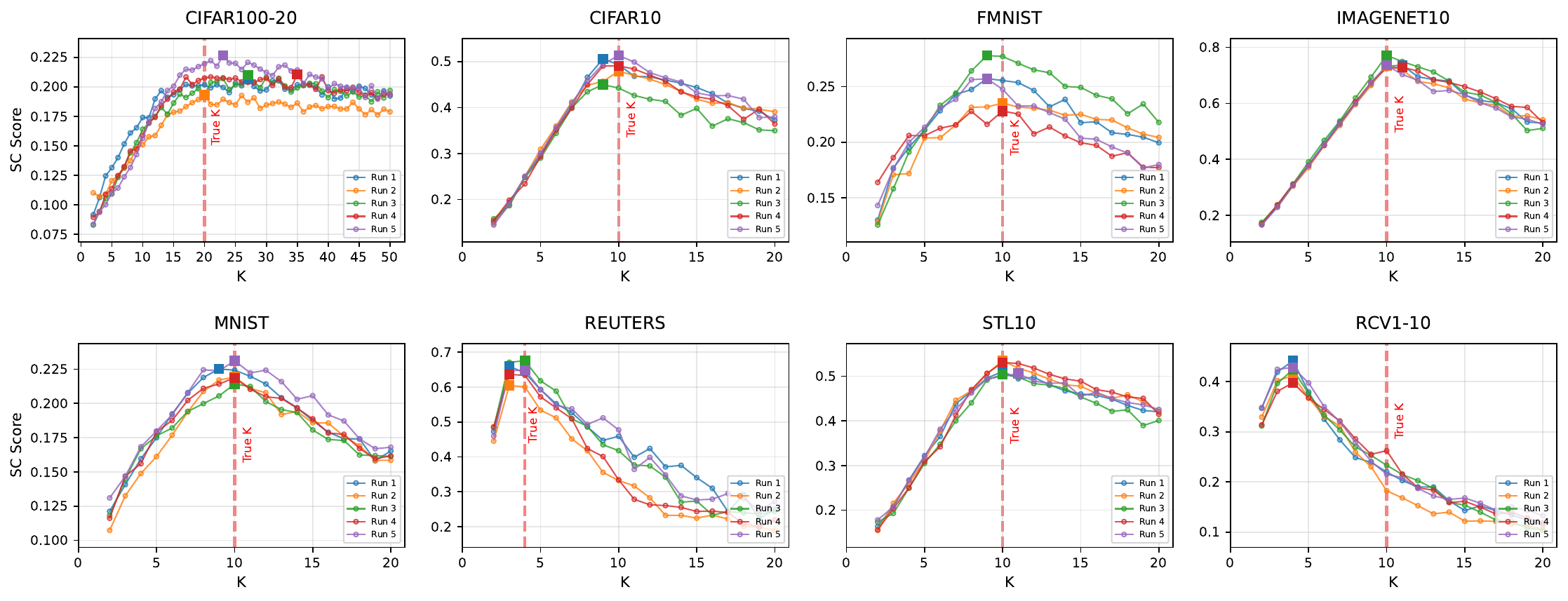}
        \caption{} \label{fig:SpherePair_SC:1k}
    \end{subfigure}
    \vskip 0.05in

    \begin{subfigure}{\linewidth}
        \centering
        \includegraphics[width=\textwidth]{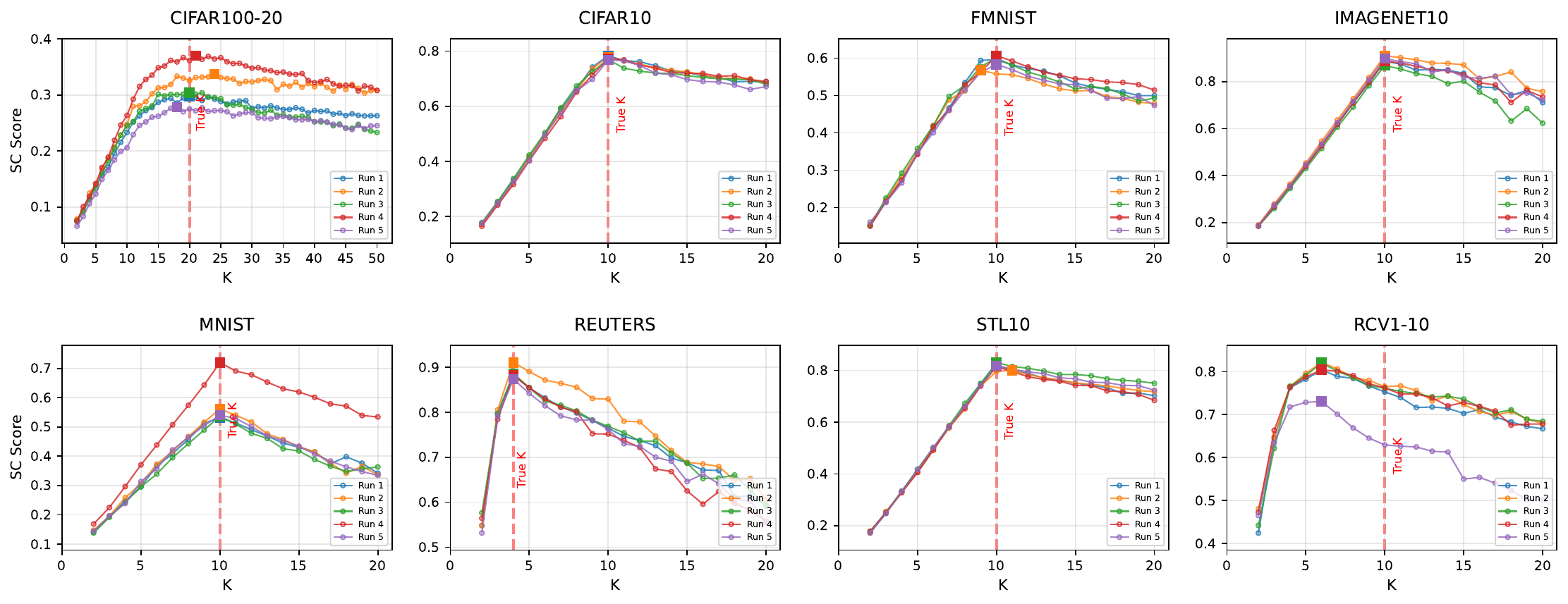}
        \caption{} \label{fig:SpherePair_SC:5k}
    \end{subfigure}
    \vskip 0.05in

    \begin{subfigure}{\linewidth}
        \centering
        \includegraphics[width=\textwidth]{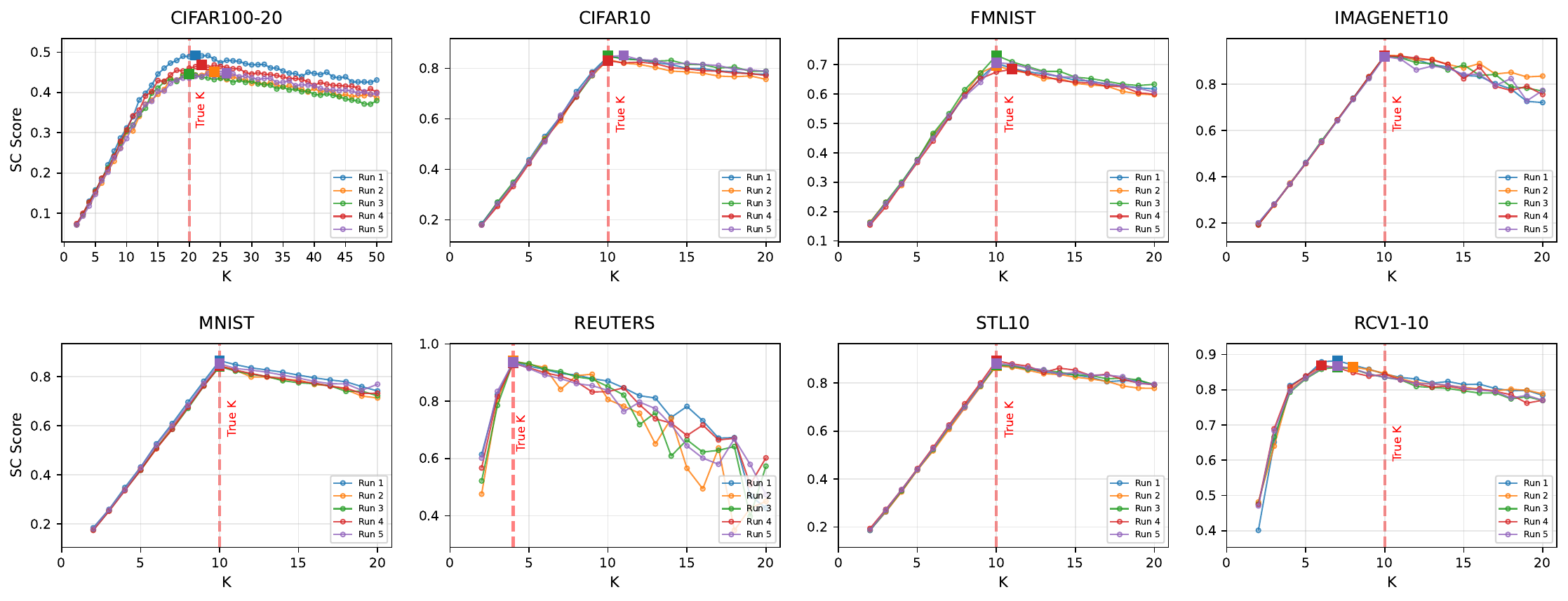}
        \caption{} \label{fig:SpherePair_SC:10k}
    \end{subfigure}
    \vskip 0.05in

    \caption{Silhouette coefficient (SC) with K-means for $K$-inference across five runs, obtained by sweeping candidate $K$ values on SpherePair embeddings learned with (a) 1k, (b) 5k, and (c) 10k constraints. The estimated $K$ corresponds to the maximum SC value (bold solid markers) in each curve. The \textcolor{red}{red lines} indicate the ground-truth $K$.}
    \label{fig:SpherePair_SC}
\end{figure}

\begin{figure}[htbp]
    \centering
    \begin{subfigure}{\linewidth}
        \centering
        \includegraphics[width=\textwidth]{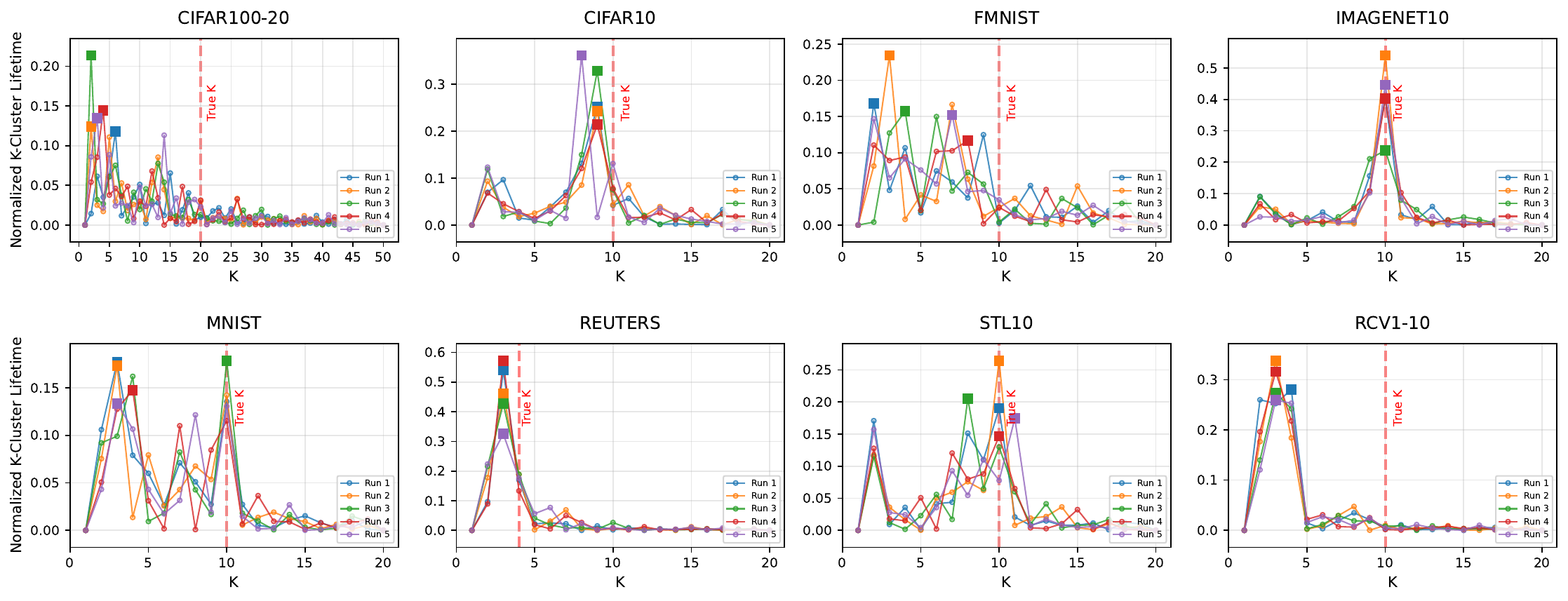}
        \caption{} \label{fig:SpherePair_lifetime:1k}
    \end{subfigure}
    \vskip 0.05in

    \begin{subfigure}{\linewidth}
        \centering
        \includegraphics[width=\textwidth]{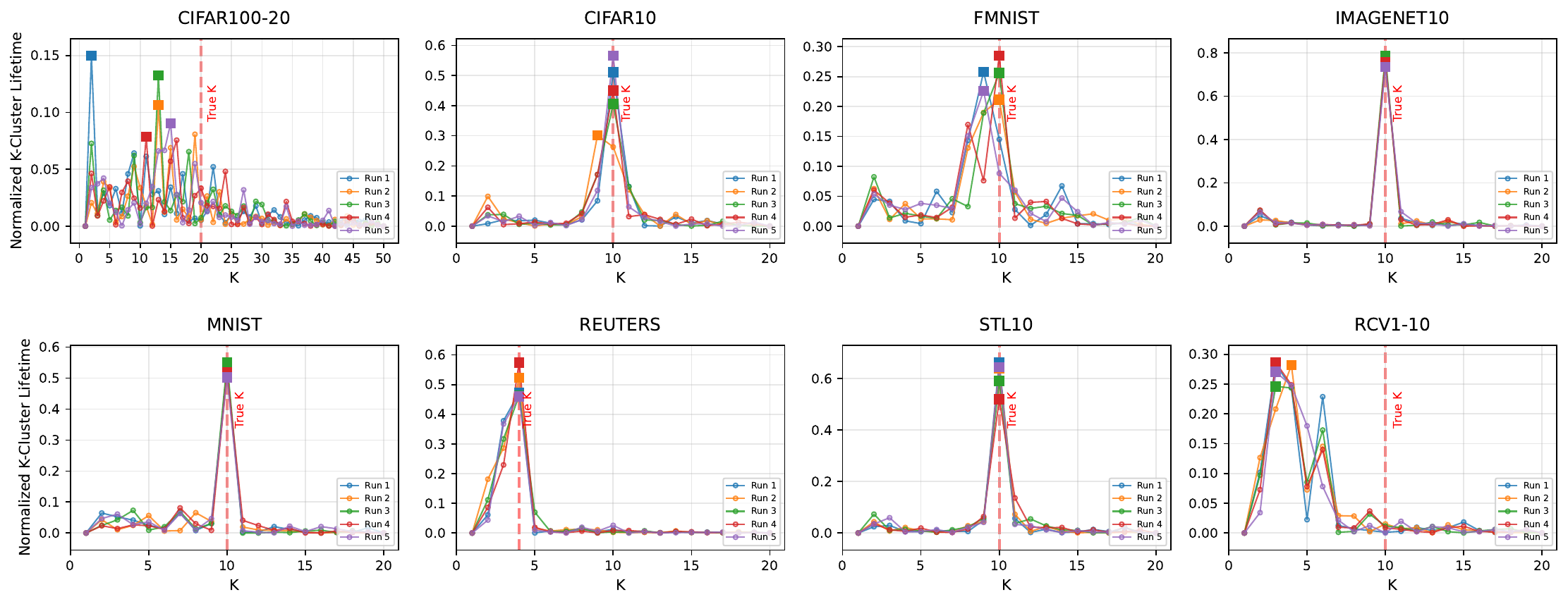}
        \caption{} \label{fig:SpherePair_lifetime:5k}
    \end{subfigure}
    \vskip 0.05in

    \begin{subfigure}{\linewidth}
        \centering
        \includegraphics[width=\textwidth]{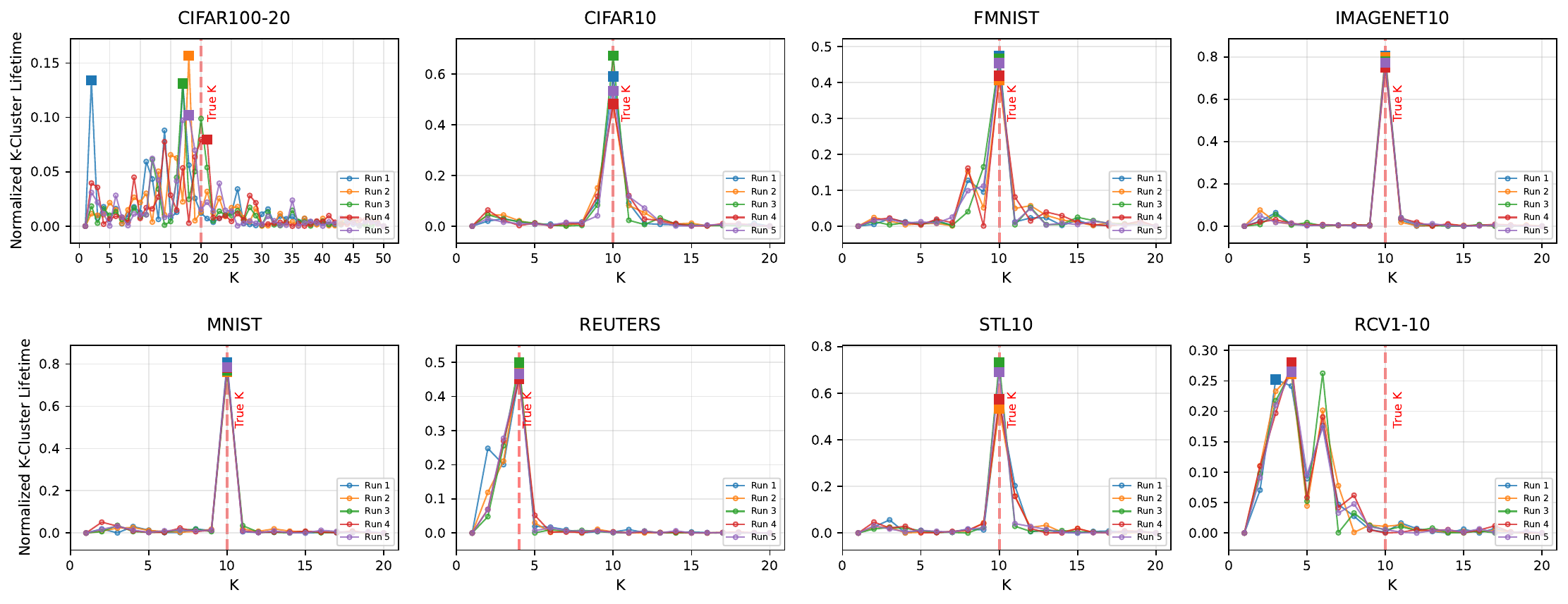}
        \caption{} \label{fig:SpherePair_lifetime:10k}
    \end{subfigure}
    \vskip 0.05in

    \caption{$K$-cluster lifetime with Agglomerative Clustering for $K$-inference across five runs, obtained from SpherePair embeddings learned with (a) 1k, (b) 5k, and (c) 10k constraints. The estimated $K$ corresponds to the maximum lifetime value (bold solid markers) in each curve. The \textcolor{red}{red lines} indicate the ground-truth $K$.}
    \label{fig:SpherePair_lifetime}
\end{figure}

\begin{figure}[htbp]
    \centering
    \begin{subfigure}{\linewidth}
        \centering
        \includegraphics[width=\textwidth]{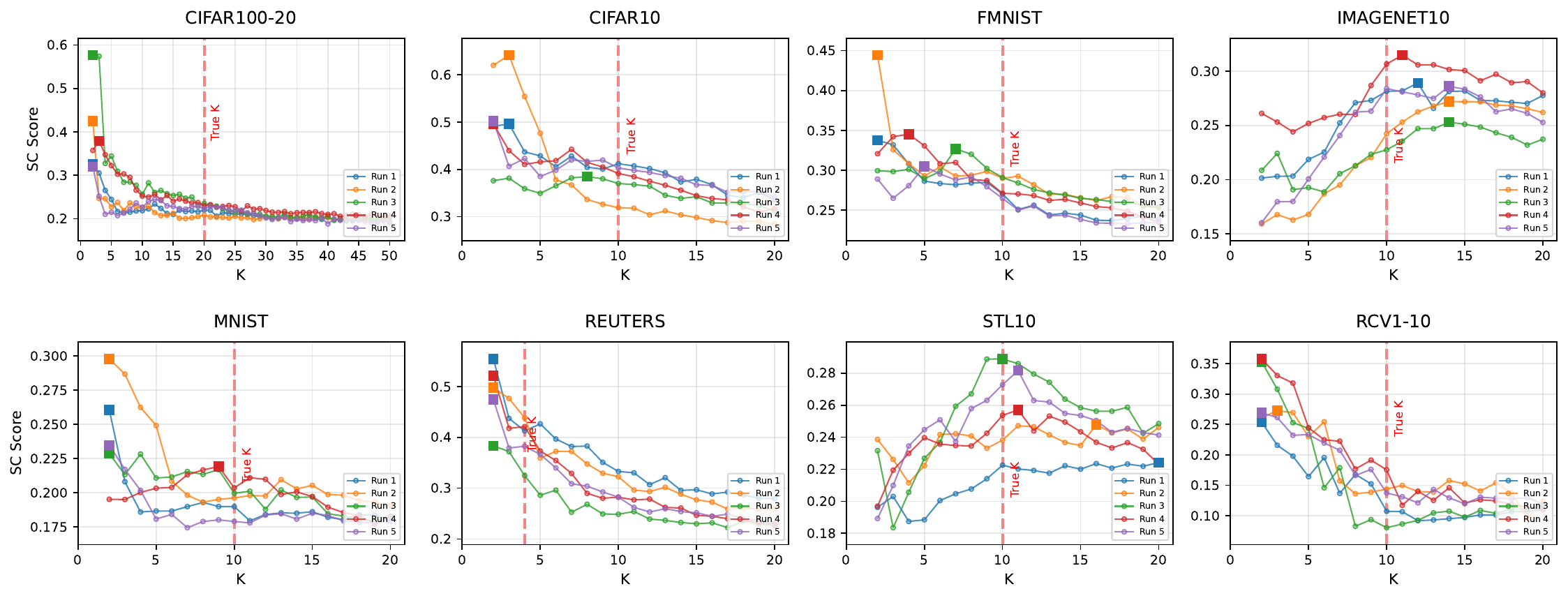}
        \caption{} \label{fig:AutoEmbedder_SC:1k}
    \end{subfigure}
    \vskip 0.05in

    \begin{subfigure}{\linewidth}
        \centering
        \includegraphics[width=\textwidth]{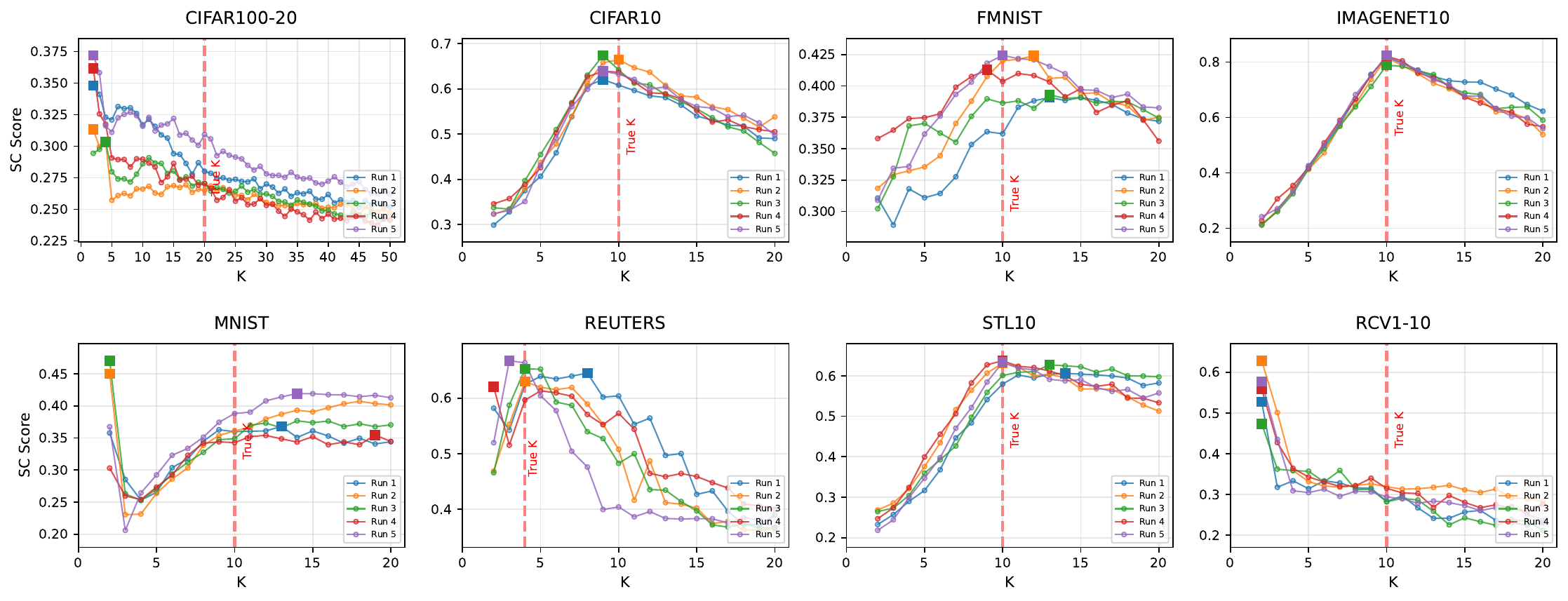}
        \caption{} \label{fig:AutoEmbedder_SC:5k}
    \end{subfigure}
    \vskip 0.05in

    \begin{subfigure}{\linewidth}
        \centering
        \includegraphics[width=\textwidth]{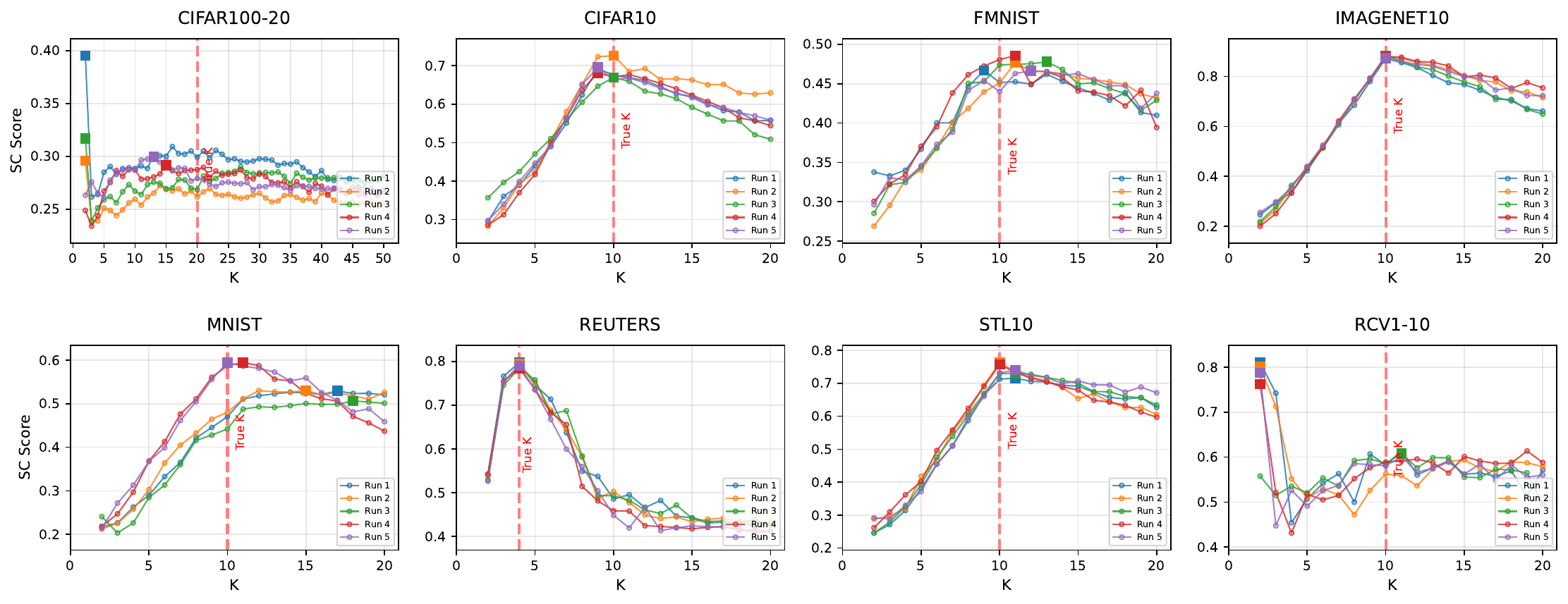}
        \caption{} \label{fig:AutoEmbedder_SC:10k}
    \end{subfigure}
    \vskip 0.05in

    \caption{Silhouette coefficient (SC) with K-means for $K$-inference across five runs, obtained by sweeping candidate $K$ values on AutoEmbedder embeddings learned with (a) 1k, (b) 5k, and (c) 10k constraints. The estimated $K$ corresponds to the maximum SC value (bold solid markers) in each curve. The \textcolor{red}{red lines} indicate the ground-truth $K$.}
    \label{fig:AutoEmbedder_SC}
\end{figure}

\begin{figure}[htbp]
    \centering
    \begin{subfigure}{\linewidth}
        \centering
        \includegraphics[width=\textwidth]{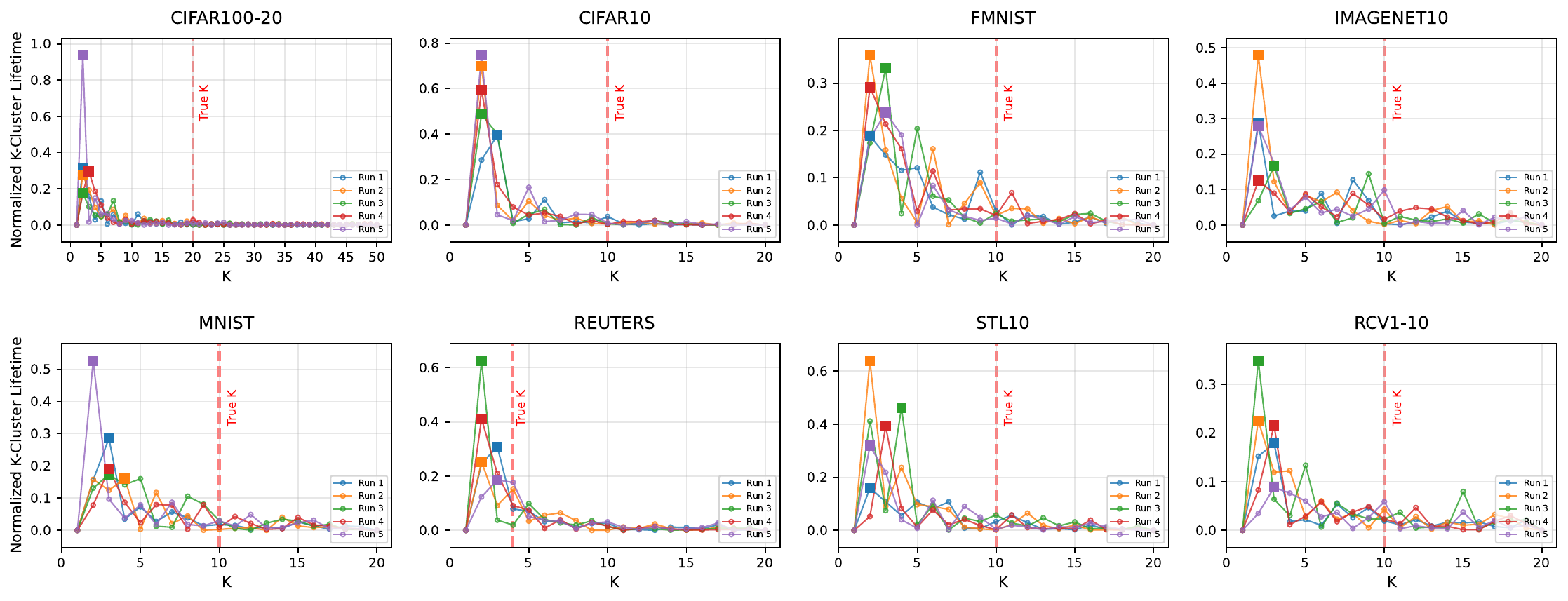}
        \caption{} \label{fig:AutoEmbedder_lifetime:1k}
    \end{subfigure}
    \vskip 0.05in

    \begin{subfigure}{\linewidth}
        \centering
        \includegraphics[width=\textwidth]{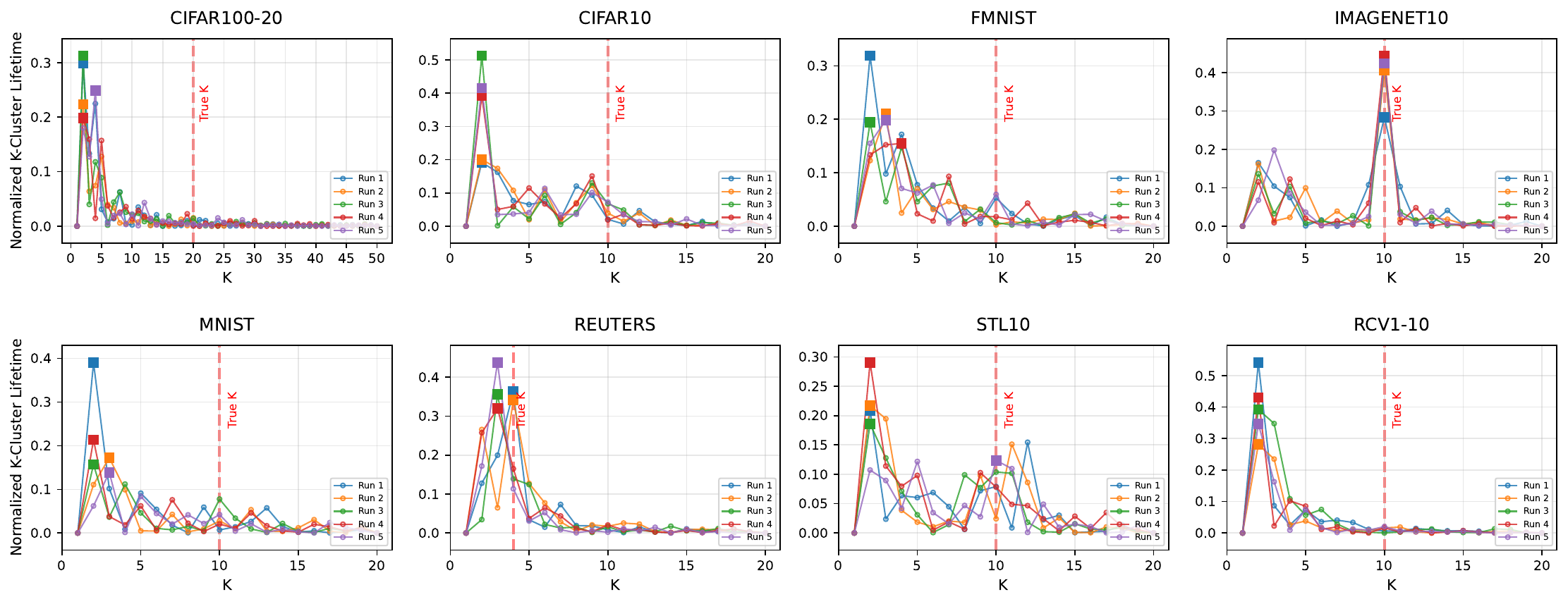}
        \caption{} \label{fig:AutoEmbedder_lifetime:5k}
    \end{subfigure}
    \vskip 0.05in

    \begin{subfigure}{\linewidth}
        \centering
        \includegraphics[width=\textwidth]{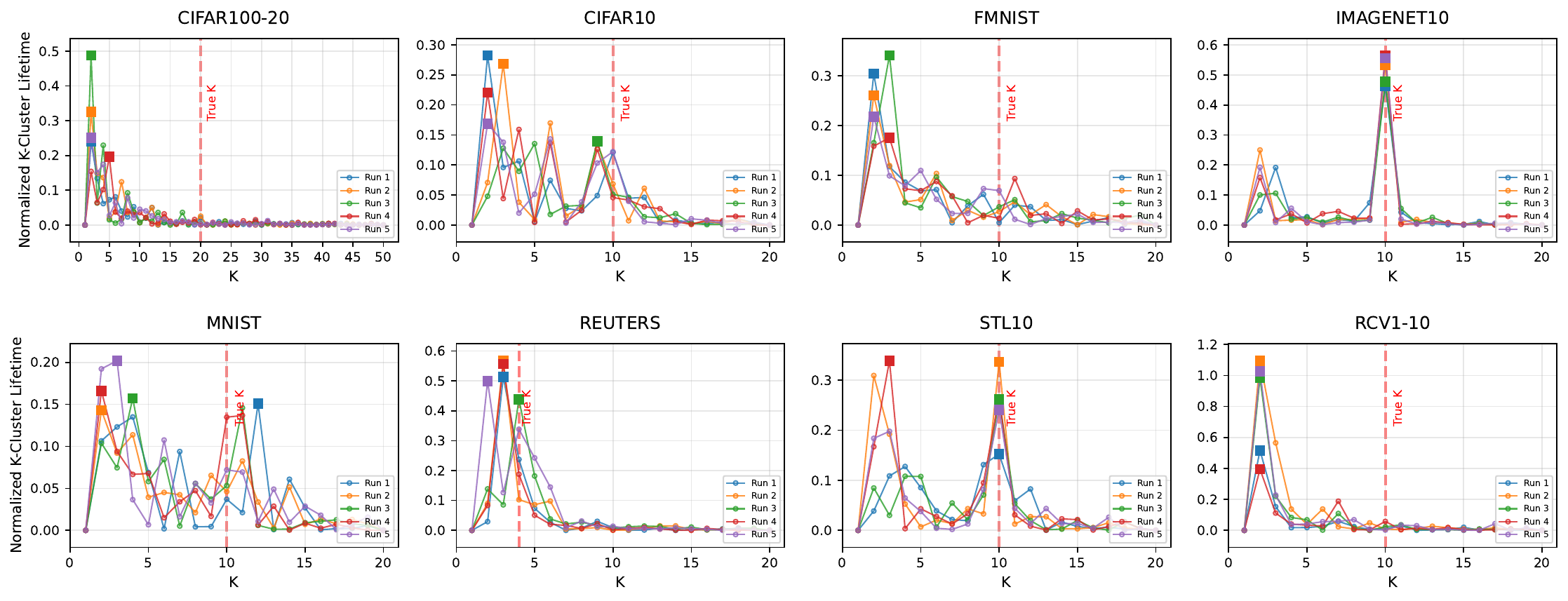}
        \caption{} \label{fig:AutoEmbedder_lifetime:10k}
    \end{subfigure}
    \vskip 0.05in

    \caption{$K$-cluster lifetime with Agglomerative Clustering for $K$-inference across five runs, obtained from AutoEmbedder embeddings learned with (a) 1k, (b) 5k, and (c) 10k constraints. The estimated $K$ corresponds to the maximum lifetime value (bold solid markers) in each curve. The \textcolor{red}{red lines} indicate the ground-truth $K$.}
    \label{fig:AutoEmbedder_lifetime}
\end{figure}

\subsubsection{Comparison with alternative post-clustering $K$-inference.}
\label{sec:unknwon_K_post_clustering}
Under the same experimental setup as in \cref{fig:K_merge_main}, we consider two post-clustering validation strategies as alternatives to our geometric approach. \cref{fig:SpherePair_SC} and \cref{fig:SpherePair_lifetime} show the curves of the silhouette coefficient (SC) with K-means and the $K$-cluster lifetime with Agglomerative Clustering, obtained by sweeping candidate $K$ values on SpherePair embeddings learned with 1k/5k/10k constraints. In both cases, the estimated $K$ is given by the curve maximum. 
Comparing \cref{fig:SpherePair_SC,fig:SpherePair_lifetime} with \cref{fig:K_merge_main}, we find:
(i) SC with K-means is slightly less accurate than our method under 10k and 5k constraints, yielding minor misestimations on CIFAR-10, FMNIST, and STL-10, together with a markedly larger discrepancy from the ground-truth $K$ on CIFAR-100-20. With only 1k constraints, while it outperforms our approach on FMNIST and MNIST, it is unstable on other datasets with around half of the estimates incorrect, and exhibits a deviation of up to 15 on CIFAR-100-20.
(ii) $K$-cluster lifetime with Agglomerative Clustering shows stronger sensitivity to the constraint level, yielding almost no correct $K$ estimates under 1k constraints except on ImageNet-10 and STL-10. Moreover, its estimates on CIFAR-100-20 are consistently unreliable with deviations up to 18 across all constraint levels, reflecting its failure in complex scenarios.

Overall, these results highlight the superior accuracy and robustness of our PCA-based $K$-inference on SpherePair embeddings. \cref{sec:efficiency} further compares computational overhead, where post-clustering methods incur additional cost from repeated clustering (multiple K-means runs or one Agglomerative clustering), while our PCA-based inference is considerably more efficient as it requires only a single closed-form PCA solution.

\subsubsection{Comparison with DCC baselines.}
\label{sec:unknwon_K_baselines}
We separately compare SpherePair with end-to-end DCC and Euclidean constraint embedding methods to highlight its advantage in scenarios with unknown $K$.

\paragraph{Comparison with end-to-end DCC.}
Unlike our deep constraint embedding approach, which allows direct geometric inference or post-clustering inference via rapid sweeping over $K$ on pre-learned representations, all end-to-end anchor-based DCC baselines require training a new model from scratch for each candidate $K$. This makes them impractical for such estimation due to the time-intensive nature of retraining (see learning efficiency in \cref{table:overall_training_time} for training times corresponding to a specific $K$).

\paragraph{Comparison with Euclidean constraint embedding.}
Under the same setup as in \cref{fig:SpherePair_SC,fig:SpherePair_lifetime}, we further apply the two post-clustering $K$-inference strategies to representations learned by the Euclidean constraint embedding baseline, AutoEmbedder, and report the results in \cref{fig:AutoEmbedder_SC,fig:AutoEmbedder_lifetime}. Comparing these with \cref{fig:SpherePair_SC,fig:SpherePair_lifetime} highlights the applicability of different learned representations to $K$-inference. 
Over 40 cases per setting (8 datasets $\times$ 5 runs), AutoEmbedder’s representations consistently struggle at all constraint levels, failing under both “K-means + SC” (39/40, 28/40, 25/40 failures for 1k, 5k, 10k constraints, respectively) and “Agglomerative + $K$-cluster lifetime” (40/40, 32/40, 30/40 failures).
This indicates that AutoEmbedder produces suboptimal embeddings that are not sufficiently structured to support reliable cluster-number inference, underscoring the superiority of our SpherePair-based approaches.

In summary, SpherePair proves highly applicable to real-world scenarios with unknown cluster numbers. By separating representation learning from clustering, it avoids the heavy retraining cost required by end-to-end DCC methods. By producing geometrically well-structured representations that remain clustering-friendly, SpherePair enables both reliable PCA-based $K$-inference and effective post-clustering validation.

\subsection{Empirical validation and hyperparameter sensitivity analysis}
\label{sec:hyperparameter_appendix}
We supplement Sect.~\ref{sec:hyperparameter} with additional results,
providing empirical validation of our theoretical insights and evaluating the robustness of our approach.

\subsubsection{Embedding dimension $D$}
\label{sec:omega_and_D}

\begin{figure}[htbp]
    \centering
    \begin{subfigure}{\linewidth}
        \centering
        \includegraphics[width=\textwidth]{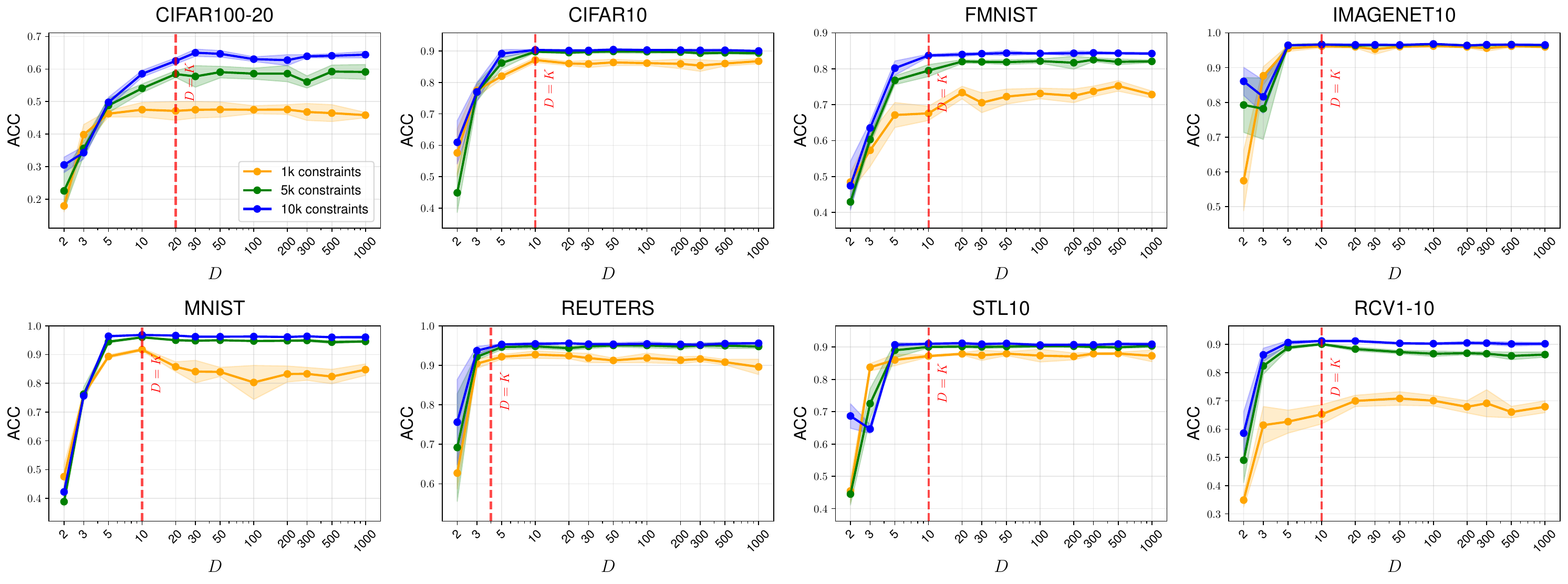}
        \caption{} \label{fig:D_merge:acc}
    \end{subfigure}
    \vskip 0.05in

    \begin{subfigure}{\linewidth}
        \centering
        \includegraphics[width=\textwidth]{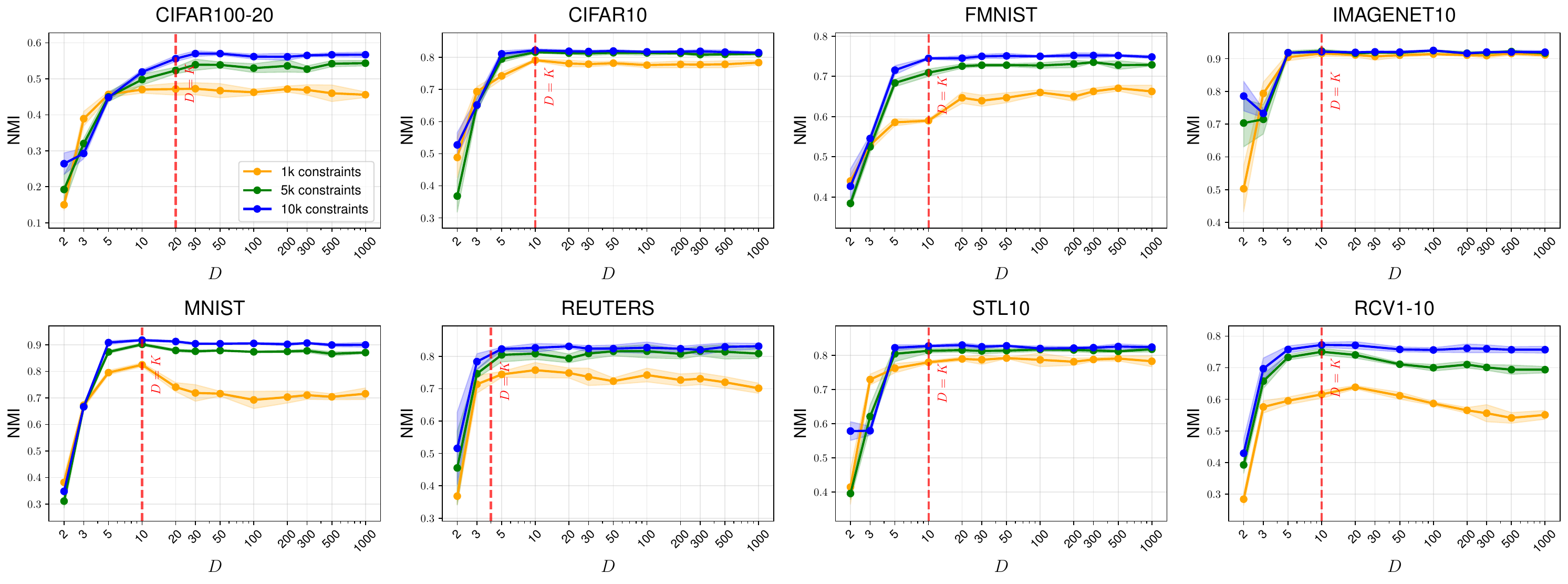}
        \caption{} \label{fig:D_merge:nmi}
    \end{subfigure}
    \vskip 0.05in

    \begin{subfigure}{\linewidth}
        \centering
        \includegraphics[width=\textwidth]{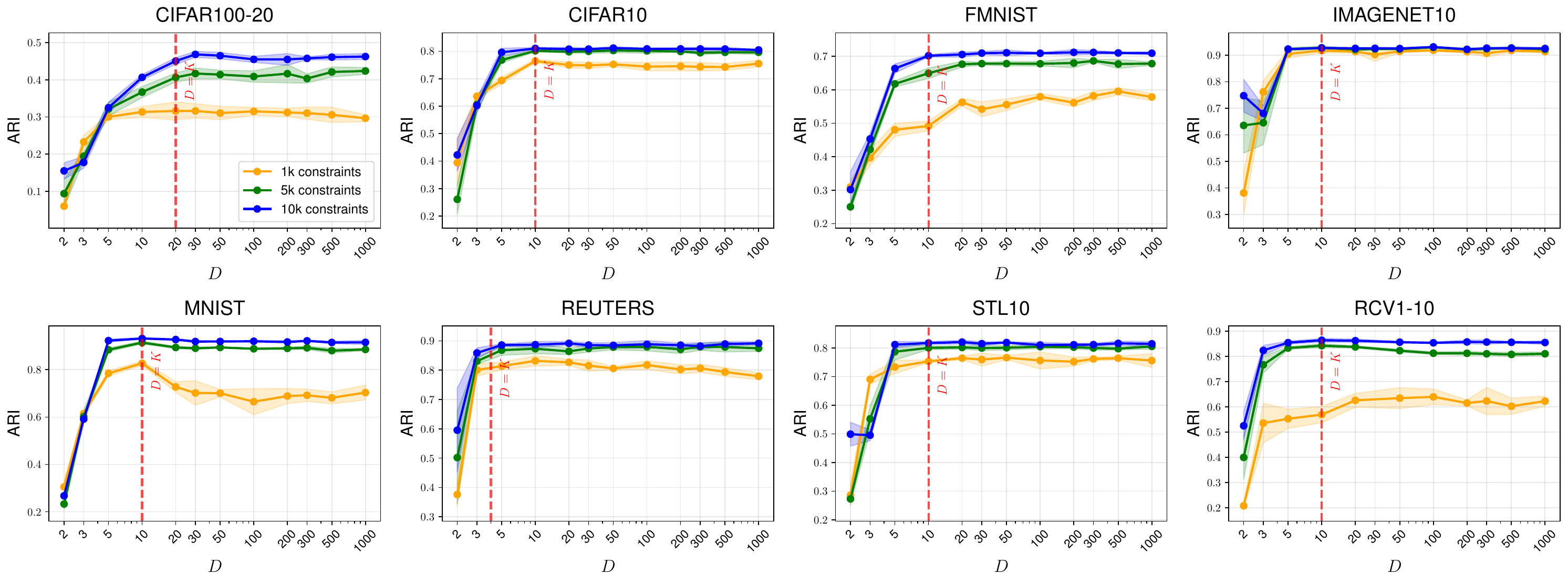}
        \caption{} \label{fig:D_merge:ari}
    \end{subfigure}
    \vskip 0.05in

    \caption{Impact of embedding dimension $D$ on SpherePair performance (mean$\pm$std over 5 runs) across datasets under 1k/5k/10k constraints:
    (a) test ACC, (b) test NMI, and (c) test ARI. The \textcolor{red}{red lines} indicate the theoretical boundary between insufficient and sufficient $D$.}
    \label{fig:D_merge}
\end{figure}

To provide a more comprehensive analysis of the impact of $D$ and support our theoretical findings, we extend our evaluation across varying constraint set sizes (1k, 5k, 10k) and multiple clustering metrics (ACC, NMI, ARI).
\cref{fig:D_merge} display the clustering performance with respect to $D$ for eight datasets under these different settings.

The results consistently demonstrate that ensuring a sufficiently large embedding dimension $D$ achieves near-optimal or fully optimal performance across datasets, metrics, and constraint levels.
Notably, the range of $D$ values yielding optimal performance corresponds to the boundary established by our theoretical analysis in Sect.~\ref{sec:theoretical} ($D \geq K$, where $K$ is the cluster number), and this correspondence becomes increasingly tight as the constraint set size grows.
This alignment underscores the reliability of our theoretical framework for conflict-free constraint embedding in angular space, and provides clear practical guidance for hyperparameter selection.

Furthermore, we observe that settings slightly below the theoretical threshold (i.e., $D \leq K-1$) do not noticeably affect SpherePair’s performance, offering useful flexibility when the number of clusters is unknown. 
This is further supported by \cref{table:CIFAR100}: despite the baselines having no theoretical restriction on $D$, SpherePair still outperforms them on CIFAR-100-20 even at $D=10$ (below the theoretical threshold $K=20$), underscoring the practical flexibility of using slightly sub‑threshold $D$.
Although a sufficiently large $D$ is generally beneficial, we note a minor exception on MNIST under 1,000 constraints: while $D=10$ yields peak results, increasing $D$ beyond 10 leads to a more pronounced drop in clustering quality. This may reflect the broader impact of large embedding dimensions and the resulting overparameterization on deep neural networks, an effect that carries both negative \cite{overparameterization-neg1,overparameterization-neg2,overparameterization-neg3} and positive \cite{overparameterization-pos1,overparameterization-pos2} implications and has been a long-standing subject in deep learning research; with larger constraint sets, however, this adverse effect diminishes, allowing higher-dimensional embeddings to continue enhancing performance.

\begin{table}[htbp]
    \caption{Clustering performance (\%) (ACC, NMI, ARI) on CIFAR-100-20 for models with varying embedding dimensions and 1k/5k/10k constraints. \textcolor{blue}{Blue} and black represent \textcolor{blue}{training} and test performance, respectively. The best results are in \textbf{bold}, and the second-best are \underline{underlined}.}
    \label{table:CIFAR100}
    \begin{center}
    \begin{scriptsize}
    \renewcommand{\arraystretch}{1} 
    \setlength{\tabcolsep}{2.5pt}    
    \setlength{\aboverulesep}{1pt}      
    \setlength{\belowrulesep}{1pt}      
    \setlength{\extrarowheight}{0pt}    

    \begin{tabular}{lccc@{\hskip 5pt}>{\hskip 5pt}ccc<{\hskip 5pt}@{\hskip 5pt}ccc}
    \toprule
      \multicolumn{1}{l}{} & \multicolumn{3}{c}{1k} & \multicolumn{3}{c}{5k} & \multicolumn{3}{c}{10k} \\
    \cline{2-10}\noalign{\vskip 2pt}
      & ACC & NMI & ARI & ACC & NMI & ARI & ACC & NMI & ARI \\
    \midrule

    VanillaDCC          
    & \textcolor{blue}{34.2}, 34.3 & \textcolor{blue}{36.0}, 36.3 & \textcolor{blue}{19.3}, 19.3
    & \textcolor{blue}{47.4}, 47.4 & \textcolor{blue}{46.7}, 47.1 & \textcolor{blue}{32.2}, 32.2
    & \textcolor{blue}{54.6}, 54.5 & \textcolor{blue}{50.2}, 50.3 & \textcolor{blue}{37.9}, 37.6 \\

    VolMaxDCC           
    & \textcolor{blue}{20.1}, 20.3 & \textcolor{blue}{21.4}, 21.6 & \textcolor{blue}{7.1}, 7.2
    & \textcolor{blue}{42.8}, 42.8 & \textcolor{blue}{41.9}, 42.1 & \textcolor{blue}{22.8}, 22.8
    & \textcolor{blue}{51.2}, 51.0 & \textcolor{blue}{48.5}, 48.7 & \textcolor{blue}{33.4}, 33.3 \\

    CIDEC (10D)         
    & \textcolor{blue}{46.2}, 45.4 & \textcolor{blue}{\textbf{47.7}}, 47.8 & \textcolor{blue}{29.7}, 29.1
    & \textcolor{blue}{48.4}, 47.7 & \textcolor{blue}{48.3}, 48.7 & \textcolor{blue}{32.2}, 31.4
    & \textcolor{blue}{49.7}, 48.8 & \textcolor{blue}{49.0}, 49.1 & \textcolor{blue}{33.8}, 32.6 \\

    CIDEC (20D)        
    & \textcolor{blue}{46.6}, 46.2 & \textcolor{blue}{\underline{47.3}}, \underline{47.9} & \textcolor{blue}{30.0}, 29.9
    & \textcolor{blue}{46.7}, 46.1 & \textcolor{blue}{45.4}, 45.7 & \textcolor{blue}{30.3}, 29.6
    & \textcolor{blue}{50.9}, 50.1 & \textcolor{blue}{48.5}, 48.8 & \textcolor{blue}{34.0}, 33.0 \\

    CIDEC (30D)        
    & \textcolor{blue}{45.2}, 44.5 & \textcolor{blue}{46.4}, 46.9 & \textcolor{blue}{30.0}, 29.8 
    & \textcolor{blue}{48.0}, 47.4 & \textcolor{blue}{46.9}, 47.2 & \textcolor{blue}{32.2}, 31.6
    & \textcolor{blue}{49.2}, 48.7 & \textcolor{blue}{48.3}, 48.8 & \textcolor{blue}{33.9}, 33.6\\

    DCGMM (10D)         
    & \textcolor{blue}{44.2}, 43.6 & \textcolor{blue}{45.1}, 45.6 & \textcolor{blue}{28.2}, 28.3
    & \textcolor{blue}{46.6}, 46.5 & \textcolor{blue}{46.0}, 46.4 & \textcolor{blue}{31.0}, 30.8
    & \textcolor{blue}{49.0}, 48.7 & \textcolor{blue}{47.8}, 48.1 & \textcolor{blue}{33.9}, 33.7 \\

    DCGMM (20D)        
    & \textcolor{blue}{44.5}, 44.2 & \textcolor{blue}{44.9}, 45.4 & \textcolor{blue}{28.7}, 28.7
    & \textcolor{blue}{48.1}, 47.9 & \textcolor{blue}{46.7}, 47.1 & \textcolor{blue}{32.2}, 32.2
    & \textcolor{blue}{52.3}, 52.1 & \textcolor{blue}{49.2}, 49.6 & \textcolor{blue}{36.7}, 36.7 \\

    DCGMM (30D)        
    & \textcolor{blue}{45.0}, 44.9 & \textcolor{blue}{46.4}, 46.8 & \textcolor{blue}{29.0}, 29.1 
    & \textcolor{blue}{45.2}, 45.2 & \textcolor{blue}{46.2}, 46.8 & \textcolor{blue}{30.1}, 30.1
    & \textcolor{blue}{46.9}, 46.7 & \textcolor{blue}{47.1}, 47.4 & \textcolor{blue}{31.8}, 31.6\\

    SDEC (10D)          
    & \textcolor{blue}{45.2}, 44.9 & \textcolor{blue}{46.9}, 47.2 & \textcolor{blue}{28.0}, 28.2
    & \textcolor{blue}{45.5}, 45.6 & \textcolor{blue}{47.9}, 48.5 & \textcolor{blue}{29.1}, 29.3
    & \textcolor{blue}{45.9}, 45.6 & \textcolor{blue}{48.2}, 48.8 & \textcolor{blue}{30.0}, 30.1 \\

    SDEC (20D)         
    & \textcolor{blue}{45.7}, 45.4 & \textcolor{blue}{47.0}, 47.5 & \textcolor{blue}{29.0}, 29.2
    & \textcolor{blue}{45.6}, 45.1 & \textcolor{blue}{47.0}, 47.5 & \textcolor{blue}{29.2}, 29.3
    & \textcolor{blue}{45.7}, 45.2 & \textcolor{blue}{47.1}, 47.7 & \textcolor{blue}{29.3}, 29.5 \\

    SDEC (30D)        
    & \textcolor{blue}{45.0}, 44.3 & \textcolor{blue}{45.9}, 46.3 & \textcolor{blue}{28.4}, 28.5 
    & \textcolor{blue}{45.2}, 44.7 & \textcolor{blue}{46.3}, 46.8 & \textcolor{blue}{29.0}, 29.2
    & \textcolor{blue}{45.3}, 44.7 & \textcolor{blue}{46.7}, 47.2 & \textcolor{blue}{29.1}, 29.2\\

    AutoEmbedder (10D)  
    & \textcolor{blue}{29.4}, 29.2 & \textcolor{blue}{31.8}, 32.0 & \textcolor{blue}{12.4}, 12.3
    & \textcolor{blue}{29.0}, 28.9 & \textcolor{blue}{35.0}, 35.3 & \textcolor{blue}{18.1}, 18.1
    & \textcolor{blue}{39.8}, 39.7 & \textcolor{blue}{42.1}, 42.4 & \textcolor{blue}{27.2}, 27.1 \\

    AutoEmbedder (20D) 
    & \textcolor{blue}{21.5}, 21.6 & \textcolor{blue}{23.1}, 23.4 & \textcolor{blue}{7.1}, 7.1
    & \textcolor{blue}{13.8}, 14.2 & \textcolor{blue}{13.5}, 13.8 & \textcolor{blue}{4.7}, 4.7
    & \textcolor{blue}{31.3}, 31.3 & \textcolor{blue}{36.6}, 36.9 & \textcolor{blue}{20.6}, 20.4 \\

    AutoEmbedder (30D)        
    & \textcolor{blue}{33.9}, 34.0 & \textcolor{blue}{34.5}, 35.2 & \textcolor{blue}{17.6}, 17.8
    & \textcolor{blue}{26.4}, 26.2 & \textcolor{blue}{30.9}, 31.3 & \textcolor{blue}{15.2}, 15.1
    & \textcolor{blue}{33.8}, 33.8 & \textcolor{blue}{40.3}, 40.3 & \textcolor{blue}{25.3}, 25.1\\

    SpherePair (Ours) (10D)    
    & \textcolor{blue}{46.9}, 46.5 & \textcolor{blue}{46.0}, 46.3 & \textcolor{blue}{31.0}, 30.9
    & \textcolor{blue}{54.6}, 54.2 & \textcolor{blue}{49.6}, 49.8 & \textcolor{blue}{37.9}, 37.5
    & \textcolor{blue}{57.9}, 57.5 & \textcolor{blue}{51.5}, 51.6 & \textcolor{blue}{41.4}, 41.1 \\

    SpherePair (Ours) (20D)   
    & \textcolor{blue}{\underline{48.3}} , \textbf{48.2} & \textcolor{blue}{\textbf{47.7}} , \textbf{48.0} & \textcolor{blue}{\textbf{32.2}} , \textbf{32.4}
    & \textcolor{blue}{\textbf{59.0}} , \textbf{58.8} & \textcolor{blue}{\underline{52.6}} , \underline{53.0} & \textcolor{blue}{\underline{41.0}} , \underline{40.9}
    & \textcolor{blue}{\underline{62.8}} , \underline{62.6} & \textcolor{blue}{\underline{55.1}} , \underline{55.5} & \textcolor{blue}{\underline{45.3}} , \underline{45.2} \\

    SpherePair (Ours) (30D)        
    & \textcolor{blue}{\textbf{48.4}} , \textbf{48.2} & \textcolor{blue}{46.8} , 47.2 & \textcolor{blue}{\underline{31.4}} , \underline{31.5} 
    & \textcolor{blue}{\underline{58.4}} , \underline{58.4} & \textcolor{blue}{\textbf{53.3}} , \textbf{53.8} & \textcolor{blue}{\textbf{41.7}} , \textbf{41.9}
    & \textcolor{blue}{\textbf{64.4}} , \textbf{64.3} & \textcolor{blue}{\textbf{56.6}} , \textbf{56.9} & \textcolor{blue}{\textbf{46.8}} , \textbf{46.5}\\

    \bottomrule
    \end{tabular}
    \end{scriptsize}
    \end{center}
    \vskip -0in
\end{table}

In summary, these results confirm that respecting the theoretically derived boundary for the embedding dimension $D$ leads to consistently strong clustering performance. In practice, choosing a sufficiently large $D \geq K$ offers a simple yet effective rule, enabling scalable and efficient solutions. This is particularly advantageous in scenarios where the exact number of clusters $K$ is unknown, as the theoretical insights offer robust guidance for parameter selection in diverse real-world applications.

\subsubsection{Regularization strength $\lambda$}
\label{sec:lambda_robustness}

\begin{figure}[t]
    \begin{center}
    \centerline{\includegraphics[width=0.85\textwidth]{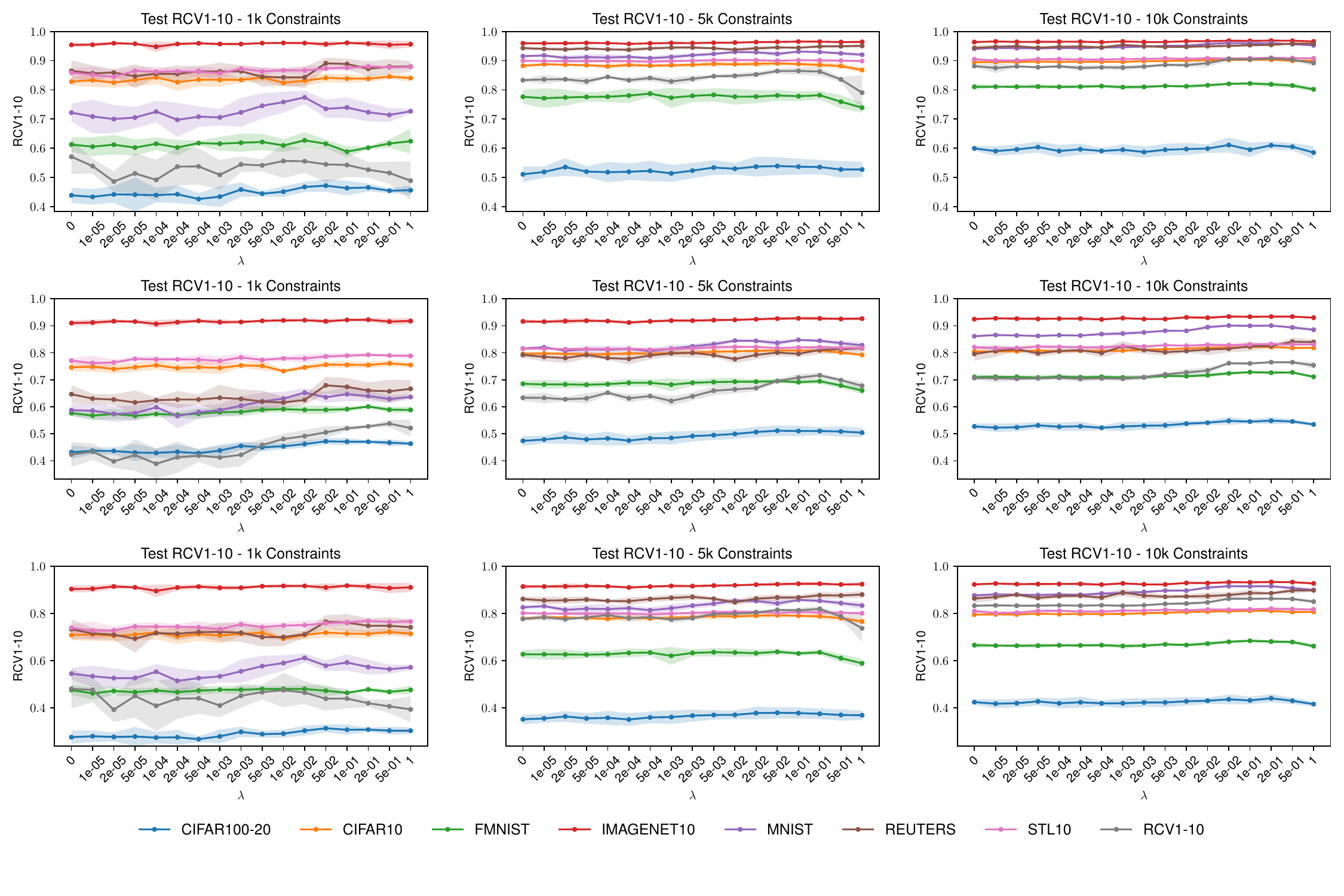}}
    \vskip -0.15in
    \caption{Performance (mean$\pm$std over 5 runs) of SpherePair$^{\dagger}$ (without pretraining) across varying $\lambda$ values (from 0 to 1.0).}
    \label{fig:lambda_noPretrain}
    \end{center}
    \vskip -0.2in
\end{figure}

\begin{figure}[t]
    \begin{center}
    \centerline{\includegraphics[width=0.85\textwidth]{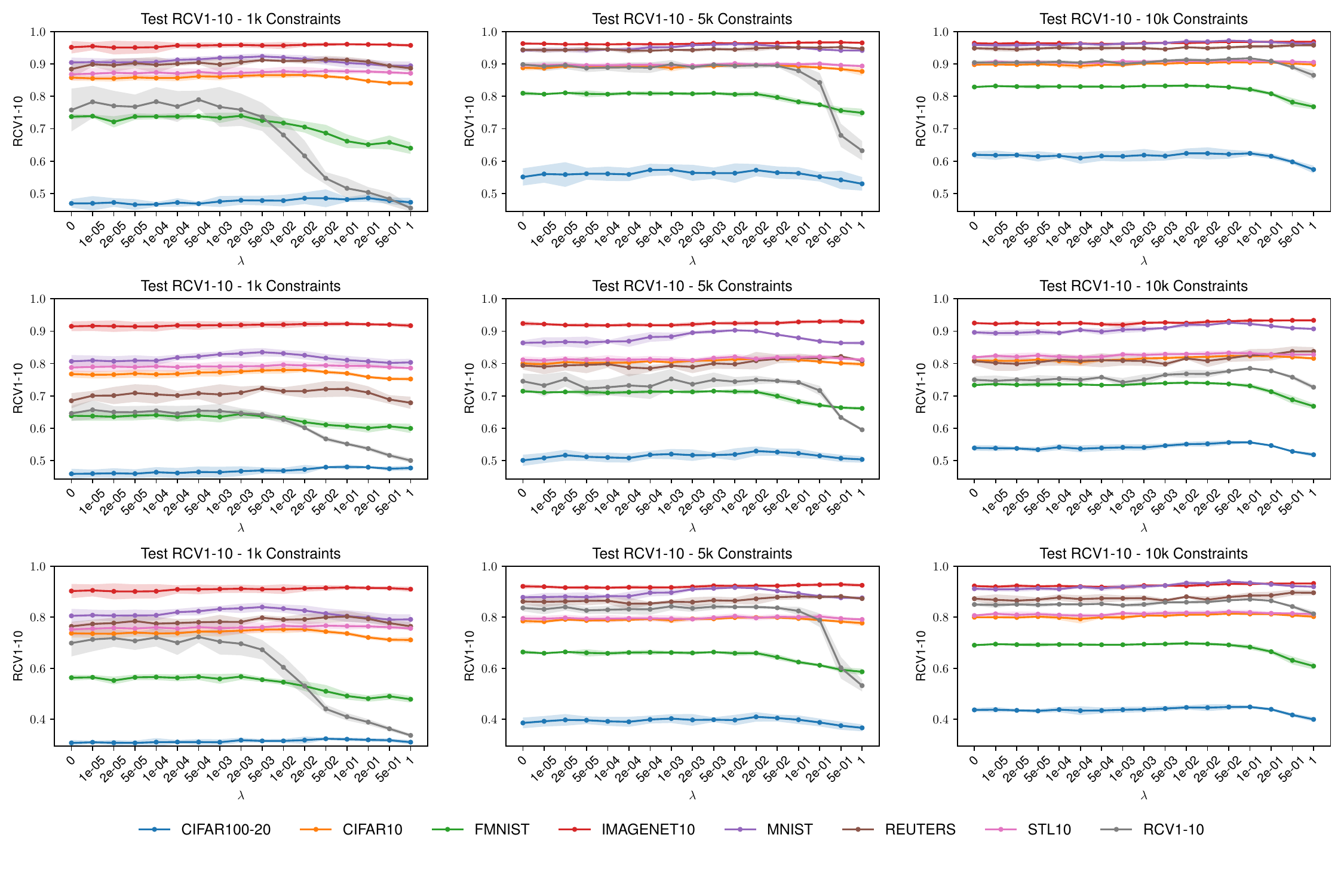}}
    \vskip -0.15in
    \caption{Performance (mean$\pm$std over 5 runs) of SpherePair (with pretraining) across varying $\lambda$ values (from 0 to 1.0).}
    \label{fig:lambda_Pretrain}
    \end{center}
    \vskip -0.3in
\end{figure}

The regularization strength $\lambda$ governs the trade-off between the reconstruction loss and the angular constraint loss in SpherePair’s objective function.
We evaluate SpherePair\(^{\dagger}\)/SpherePair across a wide range of $\lambda$ values under varying constraint levels (1k, 5k, and 10k), reporting test ACC, NMI, and ARI. The detailed results are shown in \cref{fig:lambda_noPretrain,fig:lambda_Pretrain} for both scenarios, without and with pretraining, respectively.

The results demonstrate that SpherePair is generally robust to changes in $\lambda$, with only modest performance variations, except on RCV1-10 where pretraining combined with overly large unsupervised regularization amplifies the negative effect of severe class imbalance. 
Apart from this exception, we observe that the sensitivity to $\lambda$ becomes more pronounced when the number of constraints is small, particularly in scenarios with random initialization (i.e., SpherePair\(^{\dagger}\) without pretraining), and this effect is most noticeable on CIFAR-100-20, MNIST, and Reuters. 
In these cases, selecting an inappropriate $\lambda$ may lead to suboptimal clustering results due to the insufficient supervision provided by the small constraint sets.

Despite this sensitivity, the results suggest using $\lambda=0.02$ as a default setting when validation information is unavailable, as it consistently provides strong performance across most datasets and constraint sizes. 
If prior information on class balance is available, $\lambda$ can be adapted accordingly, with larger values recommended for balanced datasets and smaller values for imbalanced datasets.

\subsubsection{Tail ratio $\rho$}
\label{sec:rho_sensitivity}

\begin{figure}[htbp]
    \centering
    \begin{subfigure}{\linewidth}
        \centering
        \includegraphics[width=\textwidth]{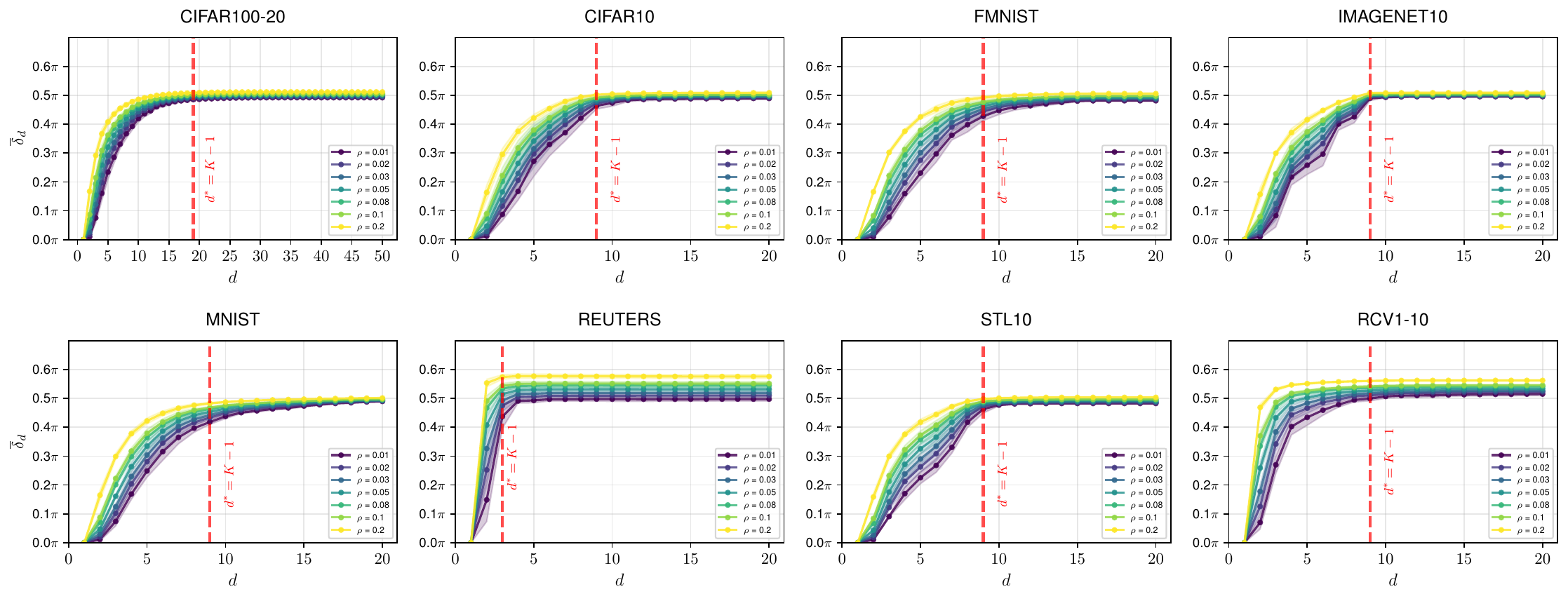}
        \caption{} \label{fig:K_merge_rho_sensitivity:1k}
    \end{subfigure}
    \vskip 0.05in

    \begin{subfigure}{\linewidth}
        \centering
        \includegraphics[width=\textwidth]{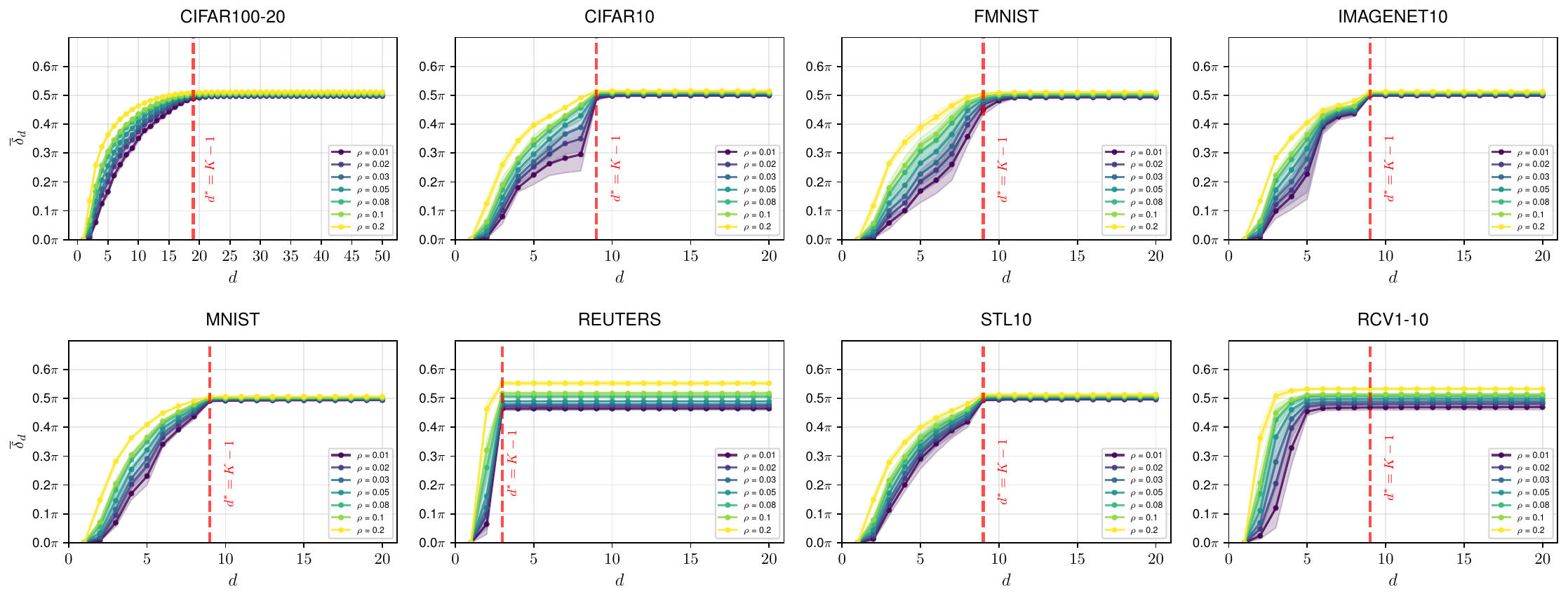}
        \caption{} \label{fig:K_merge_rho_sensitivity:5k}
    \end{subfigure}
    \vskip 0.05in

    \begin{subfigure}{\linewidth}
        \centering
        \includegraphics[width=\textwidth]{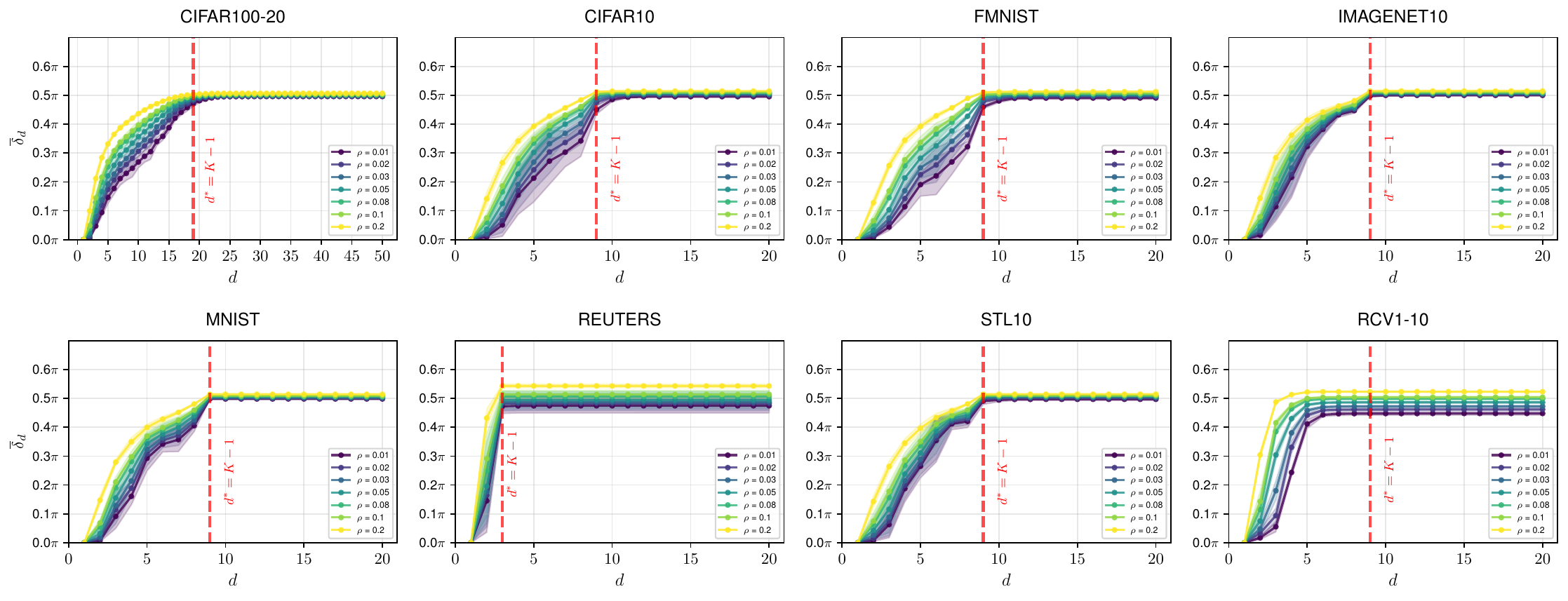}
        \caption{} \label{fig:K_merge_rho_sensitivity:10k}
    \end{subfigure}
    \vskip 0.05in

    \caption{Impact of the tail ratio $\rho$ (from 0.01 to 0.2) on the tail-averaged minimal inter-cluster angle $\overline{\delta}_d$ across PCA subspace dimensions $d$ (mean$\pm$std over five runs), obtained from SpherePair embeddings learned with (a) 1k, (b) 5k, and (c) 10k constraints on 8 datasets. The \textcolor{red}{red vertical dashed lines} indicate the ground-truth intrinsic dimension $d^\ast = K{-}1$.}
    \label{fig:K_merge_rho_sensitivity}
\end{figure}

The tail ratio $\rho$ controls the fraction of negative pairs used to compute the tail-averaged minimal inter-cluster angle in cluster-number inference. 
We evaluate $\rho$ values in the range $[0.01,0.2]$ under different constraint levels (1k, 5k, and 10k), and plot the resulting sequences $\{\overline{\delta}_d\}$ in \cref{fig:K_merge_rho_sensitivity}.

Overall, our method is robust across a broad range of $\rho$ values, although different choices of $\rho$ exhibit characteristic behaviors. 
Specifically, smaller $\rho$ produces sharper rises before $\overline{\delta}_{K-1}$ but results in values slightly below the plateau levels, whereas larger $\rho$ yields more consistent values when $d \ge K{-}1$ but makes the plateau entry less steep. 
Apart from the difficulty of inferring the true cluster number on RCV1-10 due to severe class imbalance, $\rho$ within $[0.03, 0.1]$ provides a good trade-off for highlighting the plateau entry across most datasets. 
We therefore recommend setting $\rho$ in this range for practical use.

\section{Learning efficiency}
\label{sec:efficiency}

We evaluate the learning efficiency of SpherePair and DCC baselines in terms of overall training time, clustering and cluster-number inference overhead, and provide an analysis of computational complexity.

\paragraph{Overall training time.}
Based on our resources and the implementation outlined in \cref{sec:implementation_appendix}, we report the training durations of various DCC methods using 10,000 constraints across multiple datasets. \cref{table:overall_training_time} summarizes the average overall training time for each model, including both the hyperparameter tuning and parameter estimation phases: 
(i) The hyperparameter tuning phase encompasses searching for optimal hyperparameter values and, if necessary, a single pretraining run on the training split (excluding validation samples) prior to the search; (ii) The parameter estimation phase involves training the model with the identified optimal hyperparameters, including any pretraining on the full training split if applicable. It is noteworthy that only VolMaxDCC and AutoEmbedder require hyperparameter tuning, while pretraining is performed for all methods except VanillaDCC and VolMaxDCC.

\paragraph{Clustering overhead.}
Additionally, we report the clustering analysis time for two deep constraint embedding models, AutoEmbedder and SpherePair, using K-means and Agglomerative clustering. 
Unlike other end-to-end baselines that embed clustering into the network training, these models produce clustering-friendly representations, and the time required for subsequent clustering is minimal as shown in \cref{table:clustering_time}.

\paragraph{Cluster-number inference overhead.}
When the number of clusters $K$ is unknown, clustering validation metrics are typically employed to infer the true $K$. In this setting, deep constraint embedding models (e.g., AutoEmbedder and SpherePair) incur only modest overhead, as candidate $K$ values can be swept over pre-learned representations via K-means or with a single agglomeration run. Moreover, our SpherePair further benefits from the proposed PCA-based $K$-inference, achieving even higher efficiency by bypassing post-clustering entirely through a direct PCA solution (see \cref{table:K_inference_time} for the time costs of different $K$-inference methods).
In contrast, end-to-end DCC baselines must be retrained from scratch for each candidate $K$ (see \cref{table:overall_training_time} for single-run training costs), leading to far higher computational expense.

\begin{table}[htbp]
    \caption{Overall training time for different DCC methods on various datasets using 10k constraints. All times are measured on a single Tesla V100 16G GPU. Models marked with \textsuperscript{*} require hyperparameter tuning, and their corresponding times are \underline{underlined}.}
    \label{table:overall_training_time}
    \begin{center}
    \begin{scriptsize}
    \renewcommand{\arraystretch}{1}
    \setlength{\tabcolsep}{3.1pt}    
    \setlength{\aboverulesep}{1pt}
    \setlength{\belowrulesep}{1pt}
    \setlength{\extrarowheight}{0pt}
    
    \begin{tabular}{ccccccccc}
    \toprule
    \multicolumn{1}{l}{}& CIFAR100-20 & CIFAR10 & FMNIST & ImageNet10 & MNIST  & REUTERS & STL10 & RCV1-10 \\
    \midrule
    VanillaDCC          & 6m23s       & 6m38s   & 7m7s   & 5m23s      & 7m32s  & 2m6s    & 5m47s  & 1m48s \\
    VolMaxDCC\textsuperscript{*}  & \underline{32m32s}      & \underline{77m45s}  & \underline{73m57s} & \underline{61m31s}     & \underline{70m6s}  & \underline{17m31s}  & \underline{22m7s}   & \underline{41m58s}  \\
    CIDEC               & 29m13s      & 26m10s  & 41m20s & 5m56s      & 36m9s  & 6m38s   & 6m2s   & 71m44s   \\
    DCGMM               & 21m9s       & 19m21s  & 27m55s & 4m46s      & 21m12s & 5m45s   & 4m47s  & 42m25s  \\
    SDEC                & 25m1s       & 24m33s  & 33m58s & 5m44s      & 32m53s & 6m27s   & 5m50s  & 61m12s  \\
    AutoEmbedder\textsuperscript{*} & \underline{81m32s}      & \underline{77m15s}  & \underline{89m11s} & \underline{30m11s}     & \underline{93m3s}  & \underline{36m13s}  & \underline{34m29s}  & \underline{82m39s} \\
    SpherePair (Ours)   & 25m31s      & 25m21s  & 33m51s & 6m36s      & 34m22s & 7m9s    & 6m21s  & 64m5s  \\
    \bottomrule
    \end{tabular}
    \end{scriptsize}
    \end{center}
    \vskip -0.1in
\end{table}

\begin{table}[htbp]
    \caption{Clustering analysis time for anchor-free deep constraint embedding models using K-means (\textsuperscript{†}) and hierarchical clustering (\textsuperscript{‡}).}
    \label{table:clustering_time}
    \begin{center}
    \begin{scriptsize}
    \renewcommand{\arraystretch}{1} 
    \setlength{\tabcolsep}{3.1pt}    
    \setlength{\aboverulesep}{1pt}
    \setlength{\belowrulesep}{1pt}
    \setlength{\extrarowheight}{0pt}
    
    \begin{tabular}{ccccccccc}
    \toprule
    \multicolumn{1}{l}{}& CIFAR100-20 & CIFAR10 & FMNIST & ImageNet10 & MNIST  & REUTERS & STL10  & RCV1-10 \\
    \midrule
    AutoEmbedder        & 18s\textsuperscript{†}, 45s\textsuperscript{‡} & 4s\textsuperscript{†}, 50s\textsuperscript{‡} & 6s\textsuperscript{†}, 1m25s\textsuperscript{‡} & 1s\textsuperscript{†}, 2s\textsuperscript{‡}     & 5s\textsuperscript{†}, 1m27s\textsuperscript{‡} & 1s\textsuperscript{†}, 2s\textsuperscript{‡}  & 1s\textsuperscript{†}, 2s\textsuperscript{‡}  & 9s\textsuperscript{†}, 25s\textsuperscript{‡}\\
    SpherePair (Ours)   & 10s\textsuperscript{†}, 47s\textsuperscript{‡} & 3s\textsuperscript{†}, 52s\textsuperscript{‡} & 4s\textsuperscript{†}, 1m24s\textsuperscript{‡} & 1s\textsuperscript{†}, 2s\textsuperscript{‡}     & 2s\textsuperscript{†}, 1m23s\textsuperscript{‡} & 1s\textsuperscript{†}, 2s\textsuperscript{‡}  & 1s\textsuperscript{†}, 2s\textsuperscript{‡}  & 5s\textsuperscript{†}, 23s\textsuperscript{‡}\\
    \bottomrule
    \end{tabular}
    \end{scriptsize}
    \end{center}
\end{table}

\begin{table}[htbp]
    \caption{Comparison of time costs for three $K$-inference methods across datasets, based on SpherePair embeddings learned with 10k constraints.}
    \label{table:K_inference_time}
    \begin{center}
    \begin{scriptsize}
    \renewcommand{\arraystretch}{1} 
    \setlength{\tabcolsep}{3.1pt}     
    \setlength{\aboverulesep}{1pt}
    \setlength{\belowrulesep}{1pt}
    \setlength{\extrarowheight}{0pt}
    
    \begin{tabular}{ccccccccc}
    \toprule
    \multicolumn{1}{l}{}& CIFAR100-20 & CIFAR10 & FMNIST & ImageNet10 & MNIST  & REUTERS & STL10 & RCV1-10\\
    \midrule
    K-means + SC                 & 3m14s & 30s & 39s & 18s & 34s & 20s & 18s & 1m7s\\
    Agglomerative + lifetime      & 47s   & 52s & 1m24s & 2s  & 1m23s & 2s  & 2s & 35s  \\
    PCA-based (Ours) & 3s    & 1s  & 1s   & 1s  & 1s   & 1s  & 1s & 2s \\
    \bottomrule
    \end{tabular}
    \end{scriptsize}
    \end{center}
    \vskip -0.1in
\end{table}

\paragraph{Computational complexity analysis.}
Aside from the empirical results, we analyze the computational complexity of SpherePair’s learning, which is theoretically governed by standard DNN operations, as well as our PCA-based $K$-inference, which relies on a closed-form PCA solution. Let $T_{f_{\boldsymbol{\phi}}}$ and $T_{g_{\boldsymbol{\phi}'}}$ denote the forward-pass costs of encoder $f_{\boldsymbol{\phi}}$ and decoder $g_{\boldsymbol{\phi}'}$, respectively, $|\mathcal{C}|$ the number of constraints, and ${|\mathcal{X}|}$ the number of instances. Then the cost of angular pairwise learning (scanning constrained instance pairs) is $\mathcal{O}(|\mathcal{C}| \, T_{f_{\boldsymbol{\phi}}})$, and that of angular reconstruction (scanning instances) is $\mathcal{O}({|\mathcal{X}|}(T_{f_{\boldsymbol{\phi}}}+T_{g_{\boldsymbol{\phi}'}}))$. The additional cost of PCA-based $K$-inference comes from running PCA once on the $D$-dimensional embeddings of instances involved in negative constraints $\mathcal{C}^-$, i.e., $\mathcal{O}(|\mathcal{C}^-|D^2)$. Notably, by avoiding the need to optimize $K$ anchors and clustering assignments—which would incur an additional $\mathcal{O}(KD)$ cost—SpherePair enjoys lower overhead than end-to-end DCC, while its angular reconstruction cost is on par with the standard reconstruction employed in methods such as CIDEC \cite{CIDEC1}, and its $K$-inference overhead is negligible compared to repeated clustering-based validation.

\section{Effect of network structure}
\label{sec:model_selection}

\begin{figure}[h]
    \begin{center}
    \centerline{\includegraphics[width=\textwidth]{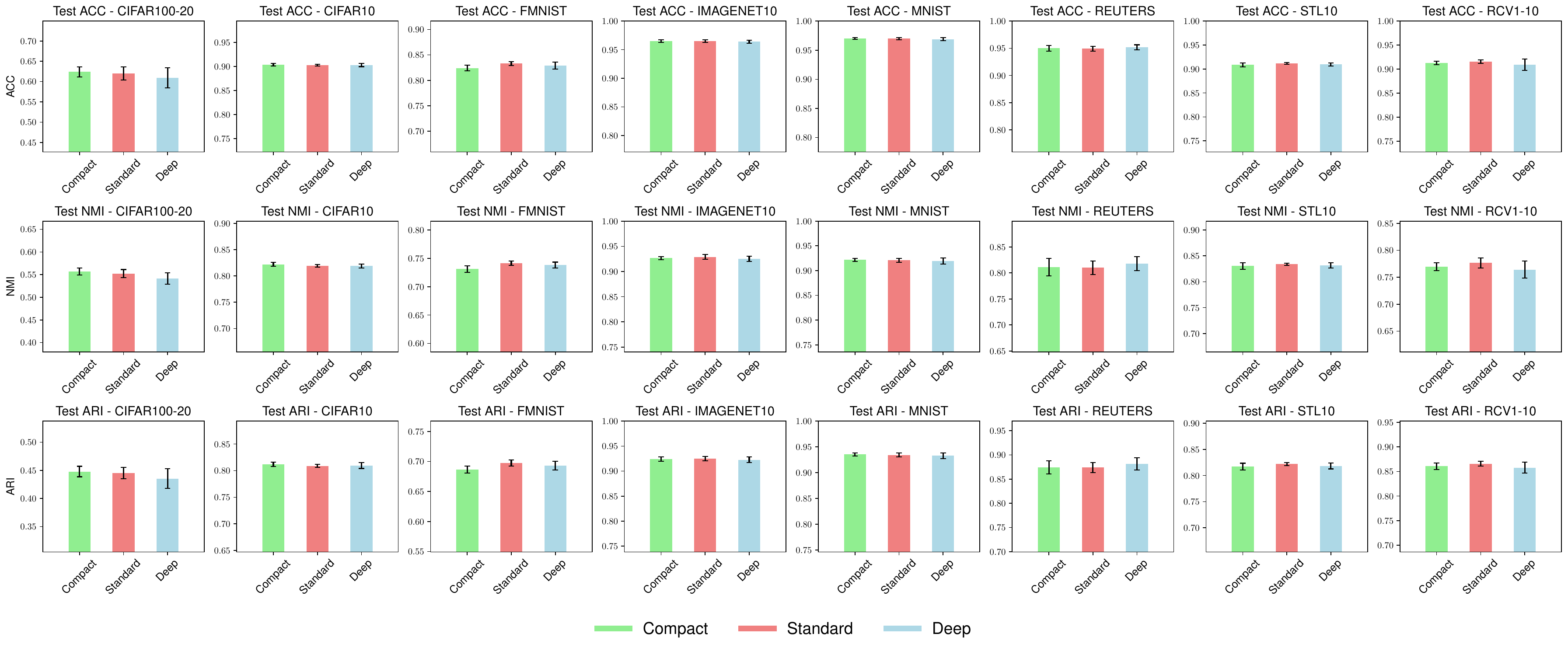}}
    \caption{Performance comparison of SpherePair across three encoder-decoder structures: \emph{Compact}, \emph{Standard}, and \emph{Deep}. Results (mean$\pm$std over 5 runs) are based on 10k balanced constraints, and metrics include test ACC, NMI, and ARI across multiple datasets.}    
    \label{fig:model_selection}
    \end{center}
\end{figure}

To evaluate the impact of network structure on SpherePair’s performance, we test three different encoder structures (paired with symmetric decoders): \emph{Compact} (256--256--512), \emph{Standard} (500--500--2000), and \emph{Deep} (500--500--500--2000). Using 10k balanced constraints, we measure test ACC, NMI, and ARI on all datasets, with results summarized in \cref{fig:model_selection}.

The results indicate that SpherePair’s performance is largely robust to the choice of network structure, with only minor differences observed across datasets. For instance, the \emph{Compact} network performs slightly better on CIFAR-100-20, CIFAR-10, and MNIST, while the \emph{Standard} network achieves the best results on FashionMNIST, ImageNet-10, STL-10, and RCV1-10. The \emph{Deep} network performs marginally better on Reuters. These variations suggest that while specific structures may provide slight advantages for certain datasets, SpherePair maintains high clustering quality across all structures.

Given the observed consistency, we recommend the \emph{Standard} structure (500--500--2000) as a practical default choice due to its balanced performance and moderate complexity. However, for real-world applications, selecting the optimal structure based on the target dataset and computational resources can further enhance performance.

\end{document}